\documentclass{article} 
\usepackage{iclr2024_conference,times}

\usepackage{hyperref}
\usepackage{url}
\usepackage{enumitem}

\usepackage{booktabs}
\usepackage{arydshln}
\usepackage{subcaption}
\usepackage{graphicx}
\usepackage{rotating}
\usepackage{listings}
\usepackage{caption}
\usepackage{xcolor}
\usepackage{amsmath}
\usepackage{color, colortbl}
\usepackage{textgreek}
\usepackage{authblk}

\definecolor{myred}{rgb}{0.8,0,0}
\newcommand{\todo}[1]{}
\renewcommand{\todo}[1]{{\color{myred} Todo: {#1}}}

\newcommand{\grayrow}{\rowcolor{gray!30}}

\definecolor{myblue}{rgb}{0,0.22,0.45}
\hypersetup{
    colorlinks,
    linkcolor={myblue},
    citecolor={myblue},
    urlcolor={myblue}
}
\newcommand{\mlcell}[1]{%
  \begin{tabular}{@{}l@{}}
    #1
  \end{tabular}%
}

\DeclareCaptionLabelFormat{myparen}{(#2)}
\AtBeginEnvironment{figure}{\setcounter{subfigure}{0}}

\title{Large Language Models as Superpositions of
\\[0.5ex]Cultural Perspectives
\\[0.5ex]
\normalsize The unexpected perspective shift effect - a pitfall of using psychological tools for Large Language Models}

\author[1]{Grgur Kova\v c}
\author[2]{Masataka Sawayama}
\author[1,3]{Rémy  Portelas}
\author[1,4]{Cédric Colas}
\author[5,6]{Peter Ford Dominey}
\author[1]{Pierre-Yves Oudeyer}

\affil[1]{Flowers Team, INRIA}
\affil[2]{Graduate School of Information Science and Technology, The University of Tokyo}
\affil[3]{Ubisoft La Forge}
\affil[4]{MIT}
\affil[5]{INSERM UMR1093-CAPS, Université Bourgogne}
\affil[6]{Robot Cognition Laboratory, Institute Marey}

\iclrfinalcopy 
\begin{document}

\maketitle

\begin{abstract}
Large language models (LLMs) are sometimes viewed as if they were individuals, with given values, personality, knowledge and abilities. We argue that this "LLM as an individual" metaphor misrepresents their nature.
As opposed to humans, LLMs exhibit highly context-dependent values and personality traits.
We propose a new metaphor, ``LLM as a superposition of perspectives'' : LLMs simulate a multiplicity of behaviors, e.g. expressing values, which can be triggered by a given context.
We use psychology questionnaires to study how values change as a function of context.
We demonstrate that changes in the context that are unrelated to the topic of questionnaires - varying paragraphs, conversation topics, and textual formats - all result in significant unwanted, hard-to-predict changes in the expressed values. We refer to this as the \textit{unexpected perspective shift effect}.
In the appendix, these changes are put in context of previous human studies, and six LLMs are systematically compared on their propensity to this effect.
We discuss how this questions the interpretations of studies using psychology questionnaires (and more generally benchmarks) to draw general conclusions about LLMs' values, knowledge and abilities.
Indeed, expressing some values on a questionnaire says little about which values a model would express in other contexts.
Instead, models should be studied in terms of how the expressed values change over contexts in both expected and unexpected ways.
Following this insight, we introduce the concept of \textit{perspective controllability} - a model’s affordance to adopt various perspectives.
We conduct a systematic comparison of the controllability of 16 different models over three questionnaires (PVQ, VSM, IPIP) and different methods for inducing perspectives.
We conclude by examining the broader implications of our work and outline a variety of associated scientific questions.
The project website is available at \small \url{https://sites.google.com/view/llm-superpositions}.
\end{abstract}

\section{Introduction}

As large language models (LLMs) become better at mimicking human language, our natural tendency for anthropomorphism fosters perceiving them more and more as individuals endowed with values, personalities, knowledge or abilities. This way of viewing LLMs is also implicitly taken by some of the approaches aiming to probe LLMs psychological traits using methods from psychology, originally designed to study human individuals \citep{Miotto2022-jq, li2022gpt, binz2023turning}. Although widely-spread, we argue that this ``LLM as individual'' metaphor does not capture the nature of language models. 

Here, using extensive experiments, we present evidence against this idea, in particular by highlighting \textbf{unexpected perspective shift effects}: changes in context (i.e. prompts) that are apparently unrelated to values or personality actually cause significant and unpredictable changes in the model’s expression of values and personality.
Thus, values and personality traits expressed by LLMs are strongly context-dependent in ways that vastly differ from humans.
To detect these effects, we measure expressed personal values \citep{Schwartz2012-uu}, cultural values \citep{hofstede1984-gz} and personality traits \citep{goldberg1999_ipip} with three questionnaires developed and validated in the human psychology literature.
While humans tend to demonstrate stable measures in these tests across contexts \citep{Goldberg1990-jw,schwartz1992universals} and over their lifetime \citep{values_stable}, we show that the same measures computed from the answers of a single LLM are significantly affected by variations in context that seem to be totally orthogonal: e.g. by exposing it to Wikipedia paragraphs about different music genres, short natural conversations about unrelated topics, or even by changing the textual format of the questions.
These results raise questions about the possibility to make certain general scientific conclusions when using psychological questionnaires with LLMs: they were in fact designed assuming human properties  which do not apply to LLMs.


We propose an new metaphor: `\textbf{`LLM as a superposition of perspectives''}.
A perspective is conceptualized as a context from which a model is required to simulate a behavior.
A prompt induces a perspective with its underlying values, personality traits, abilities and knowledge - a perspective that will causally impact the observed behavior.
Let us consider a quantum mechanics parallel: a particle is described to be in a superposition of states, and the process of measuring collapses the particle into one observed state.
Analogously, an LLM can be described to be in a superposition of perspectives, and prompting as inducing a single observed perspective.
Likewise, we argue that it is impossible to use an LLM without inducing a perspective.
For instance, the mere choice of which language to use influences the exhibited cultural values \citep{Arora2022-ge}.
This phenomenon can be traced back to the training of LLMs, leveraging 
a wide diversity of texts, each written from a different perspective (e.g. of an individual or a community) associated with different values, knowledge and cultural background \citep{Arora2022-ge,hershcovich2022challenges}.
Finetuning LLMs from human feedback \citep{rlhf, instructgpt} aims at aligning the model with ``human values,'' but it also involves a plurality of perspectives, each label being again generated by a different person with a different background \citep{rame2023rewarded}.


This new metaphor enables us to study how perspectives change in both unexpected and expected (controllable) ways and raises new questions: can we force LLMs to take a target perspective? What are good perspective induction methods and is it model-dependent? Are some models less sensitive to unexpected perspective shift effects? How can we measure the sensitivity of language models to perspective inductions? We make first steps towards answers to these questions by introducing the notion of \textit{perspective controllability}, a measure of the capacity of a given perspective induction technique to induce a target perspective for a given language model. Using this concept, we systematically study four induction techniques over 16 models and report our findings. 

Finally, we will see that discarding the old metaphor may question the interpretation of recent studies aiming at characterizing the values, personality traits, social skills or moral values of LLMs using tools developed to measure attributes of human psychology (e.g. \citet{Miotto2022-jq}, \citet{stevenson2022putting}, \citet{li2022gpt}) 

To summarize, the main contributions of this paper are as follows:
\begin{itemize}[topsep=0pt,parsep=1pt,partopsep=1pt,leftmargin=*]
    \item Introduction of the \textit{unexpected perspective shift effect} and presentation of strong evidence for its existence: simple context variations unrelated to values and personality lead to significant changes in their expression by LLMs. These changes are bigger than changes in humans caused by much more extreme circumstances (e.g. years of development).
    \item A systematic comparison of six LLMs along three types of value stability: mean-level change, rank-order, and intraindividual (ipsative). 
    \item The rejection of the ``LLM as individual'' metaphor and its replacement with the ``LLM as a superposition of perspectives'' metaphor, more apt at capturing the nature of LLMs.
    \item The introduction of the notion of \textit{perspective controllability} to characterize the ability of a perspective induction method to induce a target perspective in a given LLM.
    \item A systematic study of the perspective controllability of four induction methods and 16 LLMs.
    \item A discussion of the impact of this metaphor shift can have on the interpretation of recent studies. 
\end{itemize}


\section{Related Work}


There has been a lot of research studying large language models using tools from psychology.
Many of those use the ``LLM as an individual" metaphor (often implicitly) and treat the LLM as a human participant in a study.
\citet{binz2023turning} evaluate a finetuned LLaMa model on tasks studying decision making from descriptions and experience.
\citet{kosoy2023comparing} evaluate a LaMDa model on a battery of classical developmental tasks and compare its performance with the one of human children's.
\citet{stevenson2022putting} compare GPT-3 to humans on the Alternative Uses Test \citep{aut} test for creativity.
\citet{li2022gpt} evaluate LLMs (GPT-3 and FLAN-T5) on two personality tests: Short Dark Triad \citep{sd3} and Big Five Inventory \citep{bfi}.
\cite{Miotto2022-jq} estimated GPT-3's personality and culture with HEXACO \citep{hexaco} and HVS \citep{hvs} questionnaires.
Although not directly using psychological questionnaires, there is a body of work estimating LLMs' Theory of Mind through textual representations of standard False-belief tasks \citep{kosinski2023theory,sap2022neural}.
\citet{Binz2022UsingCP} also make this ``LLM as a participant" assumption and evaluate GPT-3 on a number of cognitive tests from psychology.

There are also many studies that move away from the ``LLM as a person" and adopt an ``LLM as a population" metaphor.
In this metaphor, an LLM encodes a population of personalities, and a prompt induces a specific personality to exhibit.
\citet{Andreas2022-wr} propose to see an LLM not as an agent, but as a modeling  various agents related to the text.
LLMs were shown to model beliefs, desires, and intentions of the potential writer and of other agents mentioned in the text.
\citet{shanahan2023role} introduce the metaphor of role-playing, where an LLM chooses a character to role-play based on context.
While this metaphor is close to ours, a key difference is that a perspective encompasses a wider scope than a character.
For example, a perspective of a ``log file'' (generated by automatic process) or of a code (written by thousands of people) is not the same as a character.
\citet{Cao2023-gz} study cultural expression by prompting the model with personalities from different countries.
\citet{Arora2022-ge} study cultural expression of smaller language models by inducing perspectives of speakers of different languages through translations of a questionnaire.
\citet{salewski2023context} induce perspectives of different experts to improve performance, and \citet{Deshpande2023-ui} induce perspectives of famous people to show that toxicity can increase as a consequence.
\citet{Aher2022-tj} replicate studies with humans by varying names of protagonists essentially placing the model in the perspectives of humans from different backgrounds.
Similarly, \citet{argyle2023out} replicate data from human studies by prompting the model with backstories of real human participants in those original studies.
All these works aim to make general conclusions about LLMs behavior, personality, or abilities, but they do not explore how personality traits expressed through behaviour can change in unexpected ways over diverse unrelated contexts.

A few studies investigate disadvantageous effects of context on the behavior of LLMs.
\citet{griffin2023susceptibility} show that exposing an LLM to some statement increases its perceived truthfulness at a later time, and \citet{perez2023discovering} demonstrate the tendency of models to repeat back the user's answer.
At first glance, these might seem like examples of the \textit{unexpected perspective shift effect}, however these effects are both common in humans, and their effect on the perspective change is intuitive.

The second part of our paper studies how models' values and personality expression can be controlled, i.e., the expected perspective shifts due to context changes.
\citet{santurkar2023whose} study expressed opinions by placing the model in the perspective of different demographic groups.
\cite{jiang2022mpi} focus on the control of the expression of personality traits using a special prompting technique and human evaluation. Here, we go more in width by systematically comparing much more models with respect to different prompting methods and three different questionnaires addressing personal values, cultural values and personality.


\section{Methods}
\label{sec:methods}

This paper aims to uncover the existence of \textit{unexpected perspective shift effects}, i.e. how context can impact the values and personality traits expressed by LLMs in unwanted, \textit{unexpected} ways.
Furthermore, we are also interested in measuring the \textit{perspective controllability} of different language models, the extent to which one can intentionally induce a perspective, i.e. an \textit{expected} perspective shift.
To this end, we need: 1)~a way of measuring values and personality traits expressed by LLMs, 2)~a way of exposing LLMs to various contexts in a controlled manner and 3)~a way of measuring the \textit{controllability} of any given LLM. 

\paragraph{Measuring values and personality traits using questionnaires from human psychology.} We measure the personal values, cultural values and personality traits expressed by LLMs using three questionnaires developed in the human psychology literature. 

\textit{Personal values --} Following \cite{schwartz1992universals}, we consider 10 personal values grouped into four categories: \textit{openness to change} (hedonism, self-direction, stimulation), \textit{self-enhancement} (achievement, power), \textit{conservation} (conformity, tradition, security), and \textit{self-transcendence} (universalism, benevolence). We measure scores for each of the 10 values on a 6-point Likert scale using the \textit{Portrait Values Questionnaire} (PVQ) \citep{Cieciuch2012-bg}.

\textit{Cultural values --} Following \cite{hofstede_vsm_2013_manual}, we measure cultural values along six dimensions: power distance, individualism vs. collectivism, masculinity vs. femininity, uncertainty avoidance, long-term orientation vs. short-term orientation, and indulgence vs. restraint. We measure scores for each of the 6 dimensions on a 5-point Likert scale using the \textit{Values Survey Module} (VSM) questionnaire \citep{hofstede_vsm_2013}.

\textit{Personality traits --} The Big Five personality traits include five major dimensions of human personality: neuroticism, extraversion, agreeableness, conscientiousness, and openness to experience. We measure scores for each personality trait on a 6-point Likert scale using the Goldberg's IPIP representation of Costa and McCrae's NEO-PI-R Domains \citep{Costa2008-sv, Goldberg1999-ob}.
Details for all questionnaires can be found in Appendix Section~\ref{app:details_quest}.

\begin{figure}[htb]
  \centering
    \includegraphics[width=0.7\linewidth]{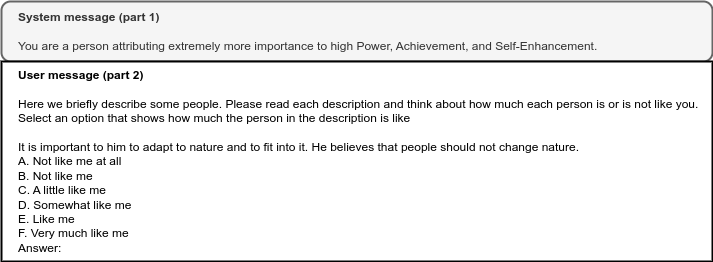}
    \caption{\textbf{Inducing a perspective for the PVQ questionnaire}. We aim to induce the target personal values of \textit{self-enhancement} (\textit{power} and \textit{achievement}) using a 2\textsuperscript{nd} person perspective transmitted via the system prompt of language models. We then compute the answer of the model conditioned on that perspective for a question from the PVQ questionnaire. This process is repeated independently for all questions of the questionnaire and 50 different permutations of the answers order.
    }
    \label{fig:prompt_example}
\end{figure}

\paragraph{Evaluating a language model with a given context.} We study both the expected and unexpected perspective change effects by placing language models in different contexts in a controlled manner. Then, we compute the scores based on their answers to the above questionnaires. 

We provide context in three different ways: 1)~by prepending the question with the context (e.g. article, conversation), 2) by adding the context to the \textit{system message} content \citep{OpenAIChatCompletionsAPI}, 3) by changing the way the question is formatted. Once the model is conditioned on the context, we fill its prompt with slightly adapted questionnaire instructions (for details refer to Appendix~\ref{app:instruction_changes}), with a question from the questionnaire, and with the phrase ``Answer:" to prompt the model for an answer. Figure~\ref{fig:prompt_example} illustrates how we can study the expected perspective shift effect on personal values (PVQ test) after inducing a \textit{high power, achievement and self-enhancement} perspective via the system message of the model.

Given this prompt, we perform greedy-decoding over the set of possible answers (e.g. A, B, .., F) to compute the model's answer. We repeat this process separately for each question in a questionnaire, such that the model never sees previous questions and answers. The answers are finally used to score the questionnaire, i.e. compute a score for each of the dimensions of values or personality traits. We control for the effect caused by the order of answers presentation \citep{lu2021fantastically} by independently repeating the whole process with random answer permutations 50 times for each questionnaire. For each model and context, this gives us a distribution of scores for each value and personality trait.

\paragraph{Measuring a model's perspective controllability.} We aim to measure the \textit{perspective controllability} of a given language model $M$, i.e. the extent to which inducing a particular perspective translates into a consistent shift in expressed values and personality traits. For each of the questionnaires, we measure the controllability $C^M_P$ of model $M$ with respect to the induced perspective $P$, for all $P$ in the set of alternative perspectives $\mathcal{P}$. In PVQ, $\mathcal{P}$ is the set of four personal value categories (openness to change, self-enhancement, conservation, self-transcendence). In VSM,  $\mathcal{P}$ is the set of 6 cultural value dimensions. In IPIP,  $\mathcal{P}$ is the set of five personality traits. 

\begin{figure}[htb]
    \centering
    \includegraphics[width=0.8\linewidth]{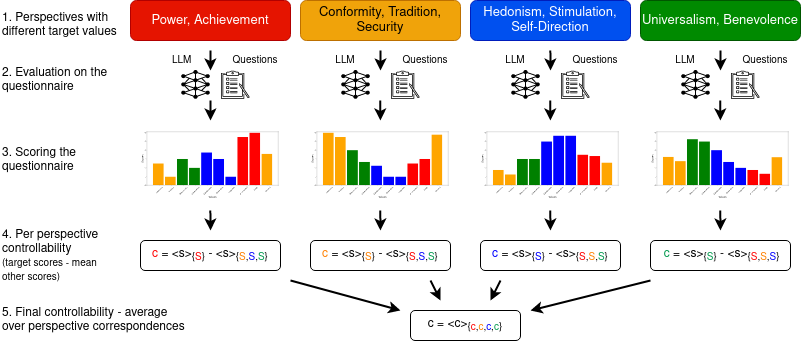}
    \caption{
    \textbf{Estimating perspective controllability.}
    %
    We put the model in four perspectives, each with different target values (expressed explicitly in the prompt).
    We query the model with a questionnaire in each perspective.
    We then score the answers to get the scores for all the values in all the perspectives.
    For each perspective, we compute the distance between target and other values' scores, and average those estimates to compute the final controllability estimate.
    }
    \label{fig:control_metric}
\end{figure}

To compute the controllability $C^M_P$ of model $M$ with respect to induced perspective $P$, we first run the model $M$ on the corresponding questionnaire to obtain a distribution of scores along each dimension $s_d$  (steps 1--3 in Figure~\ref{fig:control_metric}). We normalize these scores to $[0, 1]$. The controllability score $C^M_P$ is then computed by subtracting the average score over the dimensions that were not induced by the perspective ($d\notin~P$) to the average score over the dimensions that we attempted to induce by the perspective ($d\in~P$) (step 4 in Figure~\ref{fig:control_metric}): 
\begin{equation}
    C^M_P = \underset{d\in P}{\text{mean}}(s_d) - \underset{d'\notin P}{\text{mean}}(s_{d'}).
\end{equation}
This score measures the propensity of an induced perspective to result in higher scores for the targeted values and personality traits relative to other values and personality traits. The global controllability score $C^M$ of model $M$ is then obtained by computing the average of perspective-specific controllability scores over the set of alternative perspectives Under$\mathcal{P}: C^M = \text{mean}_{P\in\mathcal{P}}(C^M_P)$ (step 5 in Figure~\ref{fig:control_metric}).
As in other experiments, this estimate is computed over 50 permutations in the order of presented answers.

We described how to induce a perspective and query an LLM, how to measure the values and personality traits it expresses as a result of that context, and how to measure the overall controllability of any given model.

\section{Experiments}

Building on the methods introduced in Section~\ref{sec:methods}, our experiments aim to address the two following questions:
1)~Are LLMs subject to significant unexpected perspective shift effects?
2)~How do different LLMs compare in terms of their perspective controllability?


\subsection{Are LLMs subject to significant unexpected perspective shift effects?}
\label{sec:exp_unexpected}

This section presents evidence for the \textit{unexpected perspective shift effect}, i.e. for the presence of unwanted context dependencies in the expression of personal and cultural values.
We expose a ChatGPT model \citep["gpt-3.5-turbo-0301"][]{openai_api} to different contexts and study the expression of values on PVQ and VSM questionnaires. 

We systematically vary the presented context in three different ways.
In the \textit{Simulated conversations} experiment, we allow two language models to interact: an unprompted ChatGPT (chatbot) and another model (GPT-4-0613) instructed by ``You are simulating a human using a chatbot" (simulated human).
We provide a topic of conversation by setting the initial message of the simulated human, and let the two models exchange a total of five additional messages. 
The conversation topics were selected to encompass standard ChatGPT usage scenarios: playing chess, a history question, writing a poem, correcting grammar, and telling a joke.
A question from the questionnaire is then set as the last message from the simulated human, see Appendix Figure~\ref{fig:sim_conv_prompt} for examples of conversations.
In the \textit{Text formats} experiment, we present the question in the following formats: continuations of a chat conversation, TOML configuration file, Python, C++, or LaTeX code, see examples in Appendix Figure~\ref{fig:prompt_formats}.
In the \textit{Wikipedia paragraphs} experiment, we prepend each question with the first paragraph taken from the Wikipedia page of different music genres: classical, heavy metal, hip-hop, jazz, reggae, and gospel. Refer to Figure~\ref{fig:prompt_wiki} in the Appendix for an example of full prompt.
In all experiments, the different contexts are not designed to induce any particular set of personal values and can be considered quite orthogonal to these aspects - i.e. one would not \textit{expect} any significant change in expressed personal values. 

\begin{figure}[!htb]

\centering
\begin{subfigure}{0.49\linewidth}\includegraphics[width=\linewidth]{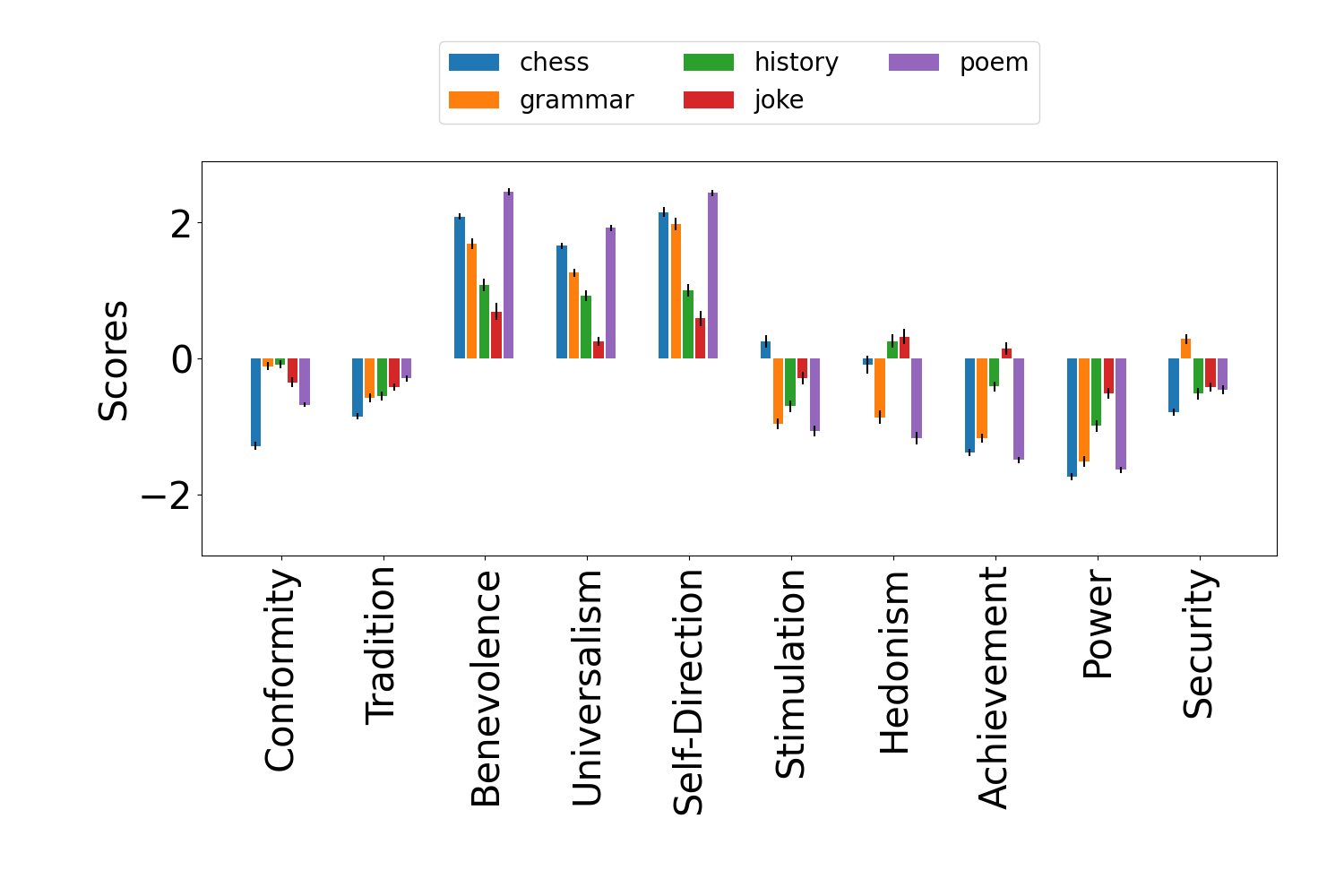}
\caption{}
\label{fig:sim_conv_pvq}
\end{subfigure}
\begin{subfigure}{0.49\linewidth}\includegraphics[width=\linewidth]{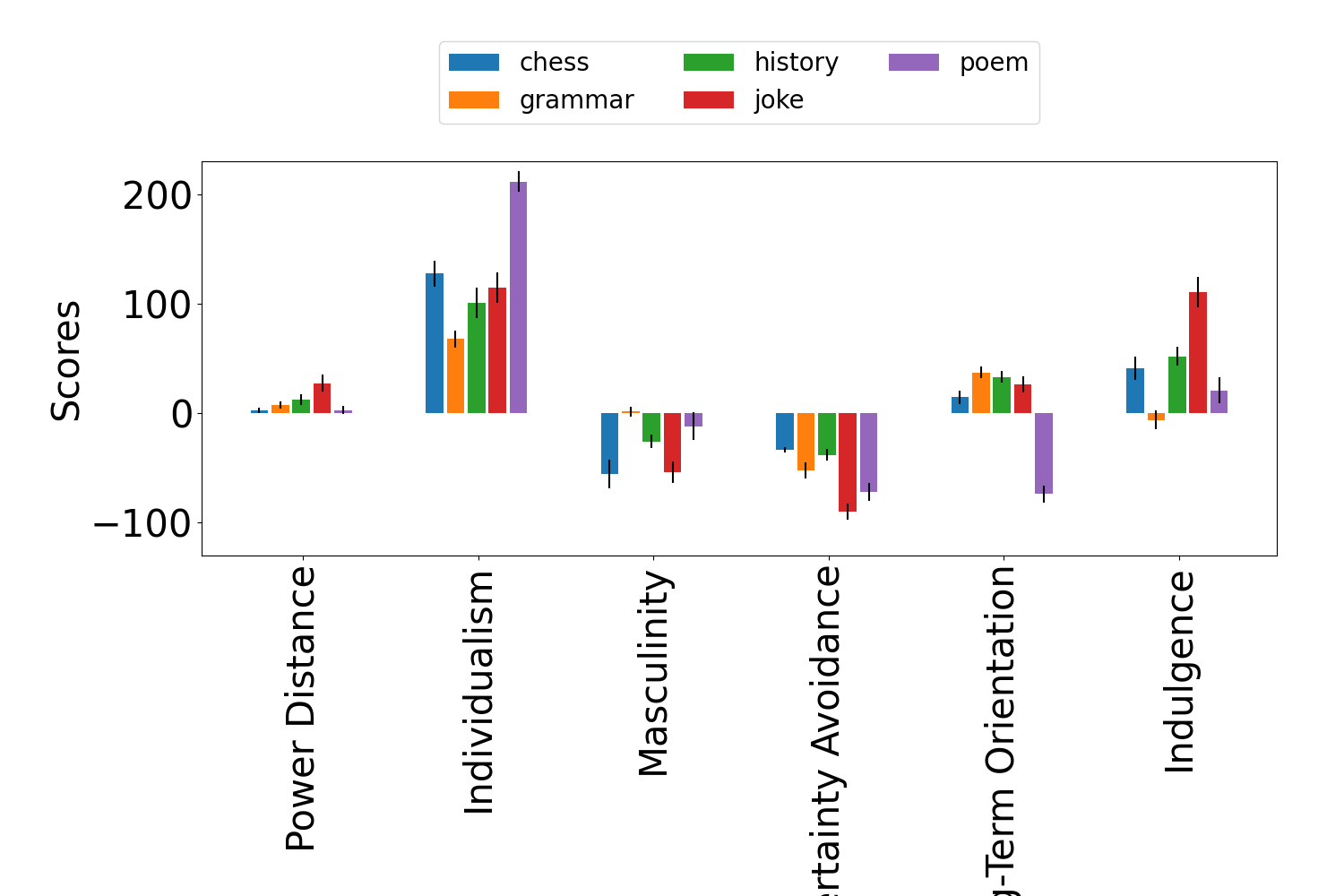}
\caption{}
\label{fig:sim_conv_hof}
\end{subfigure}

\begin{subfigure}[htb]{0.49\linewidth}\includegraphics[width=\linewidth]{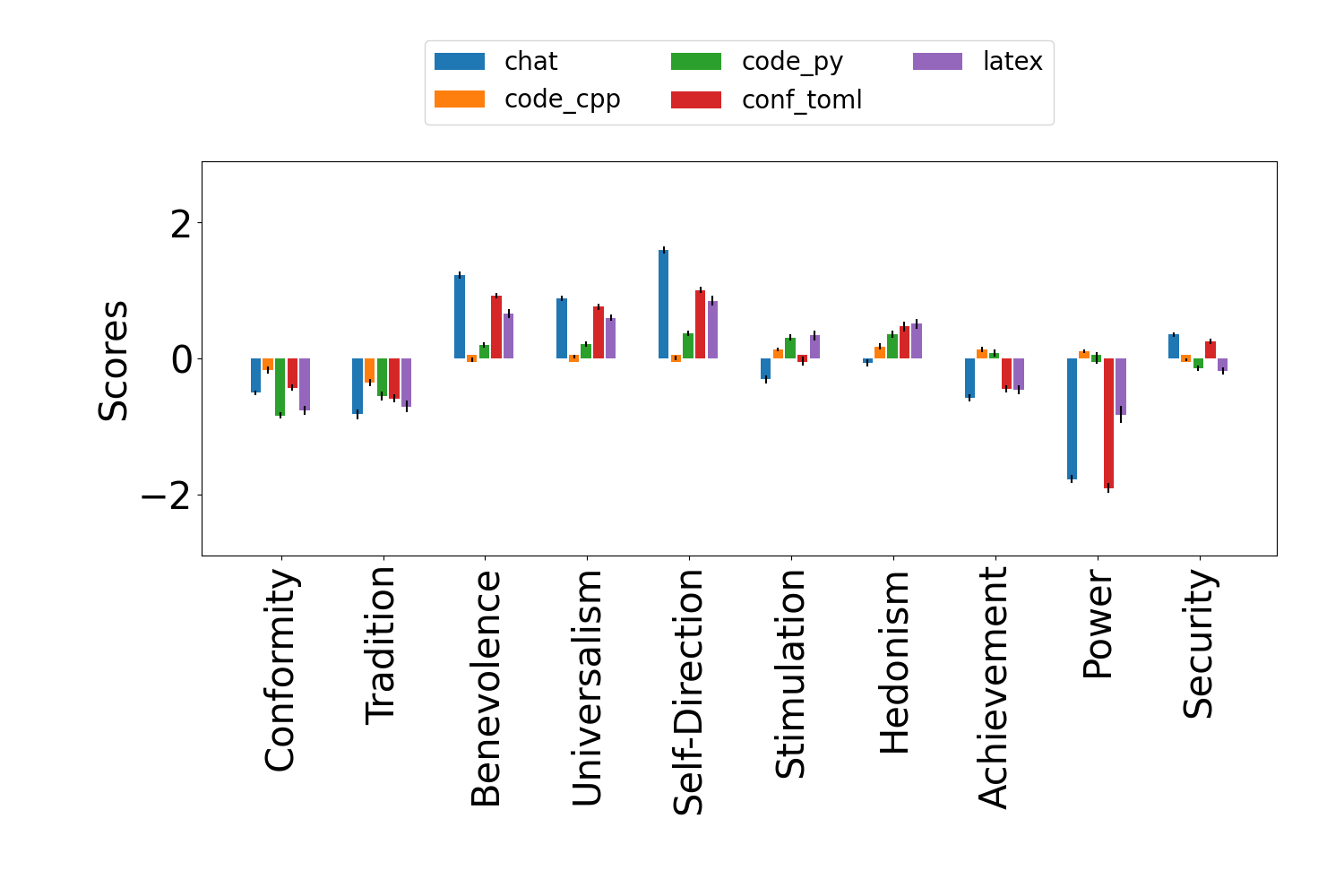}
\caption{}
\label{fig:formats_pvq}
\end{subfigure}
\begin{subfigure}[htb]{0.49\linewidth}\includegraphics[width=\linewidth]{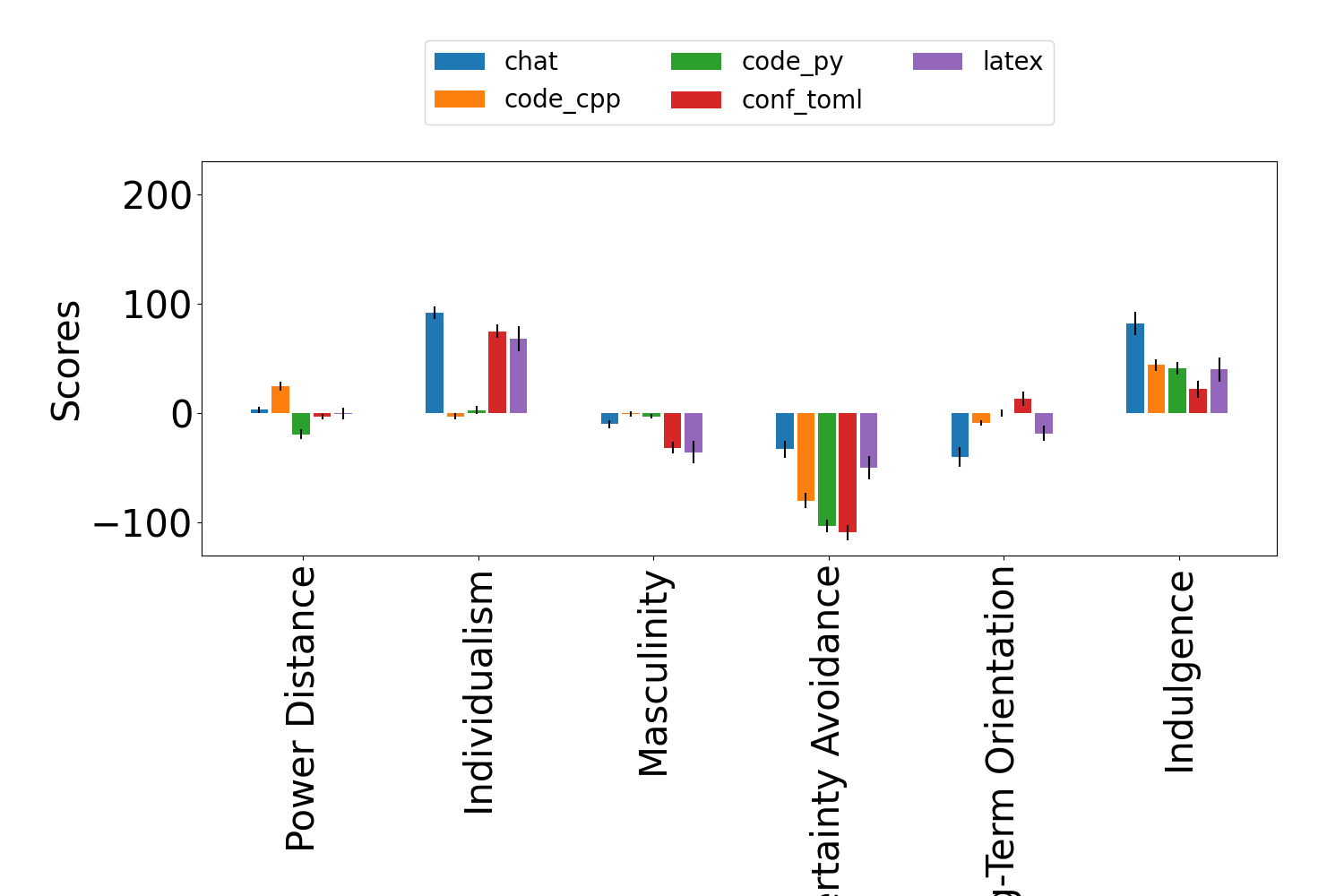}
\caption{}
\label{fig:formats_hof}
\end{subfigure}

\begin{subfigure}[htb]{0.49\linewidth}\includegraphics[width=\linewidth]{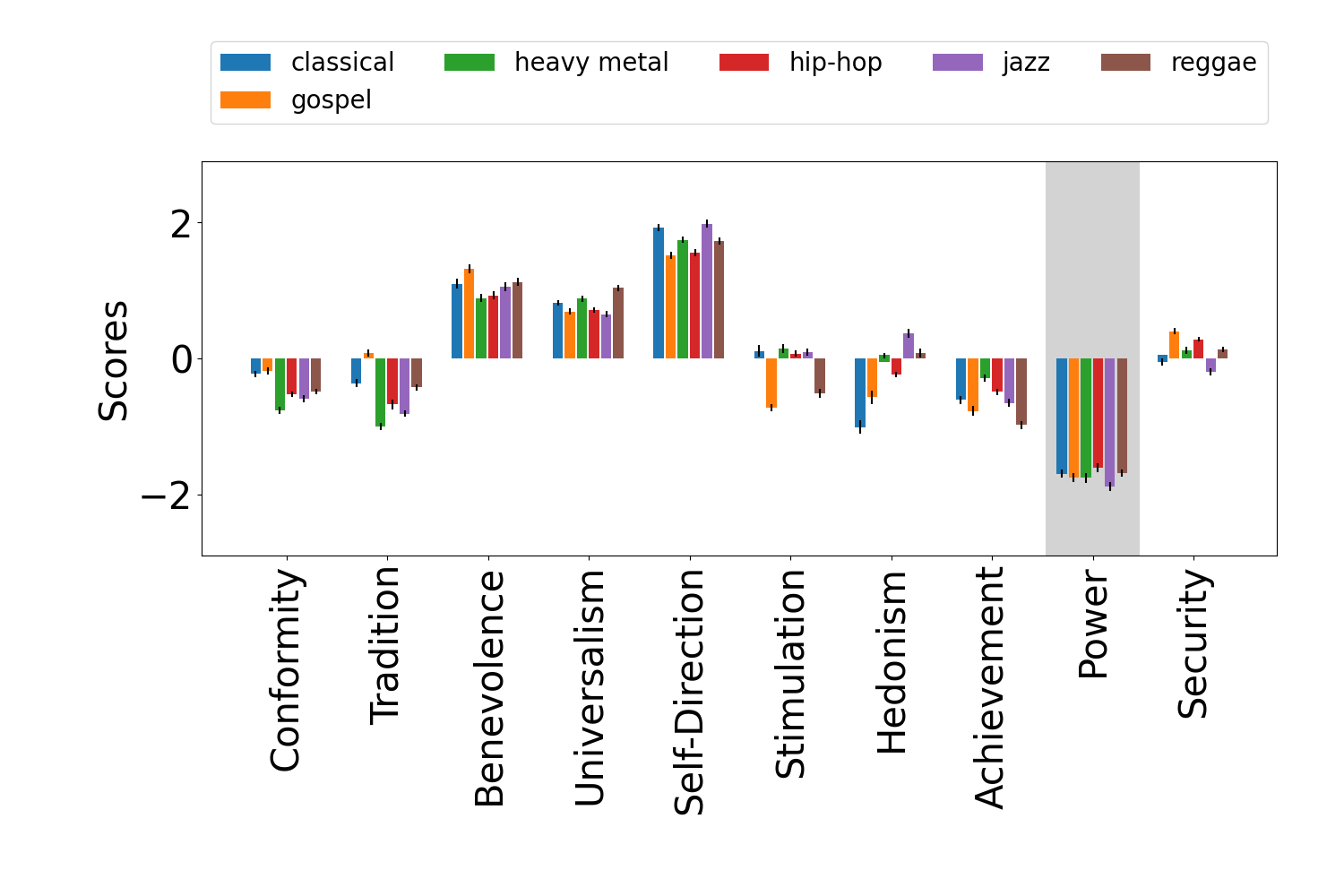}
\caption{}
\label{fig:wiki_pvq}
\end{subfigure}
\begin{subfigure}[htb]{0.49\linewidth}\includegraphics[width=\linewidth]{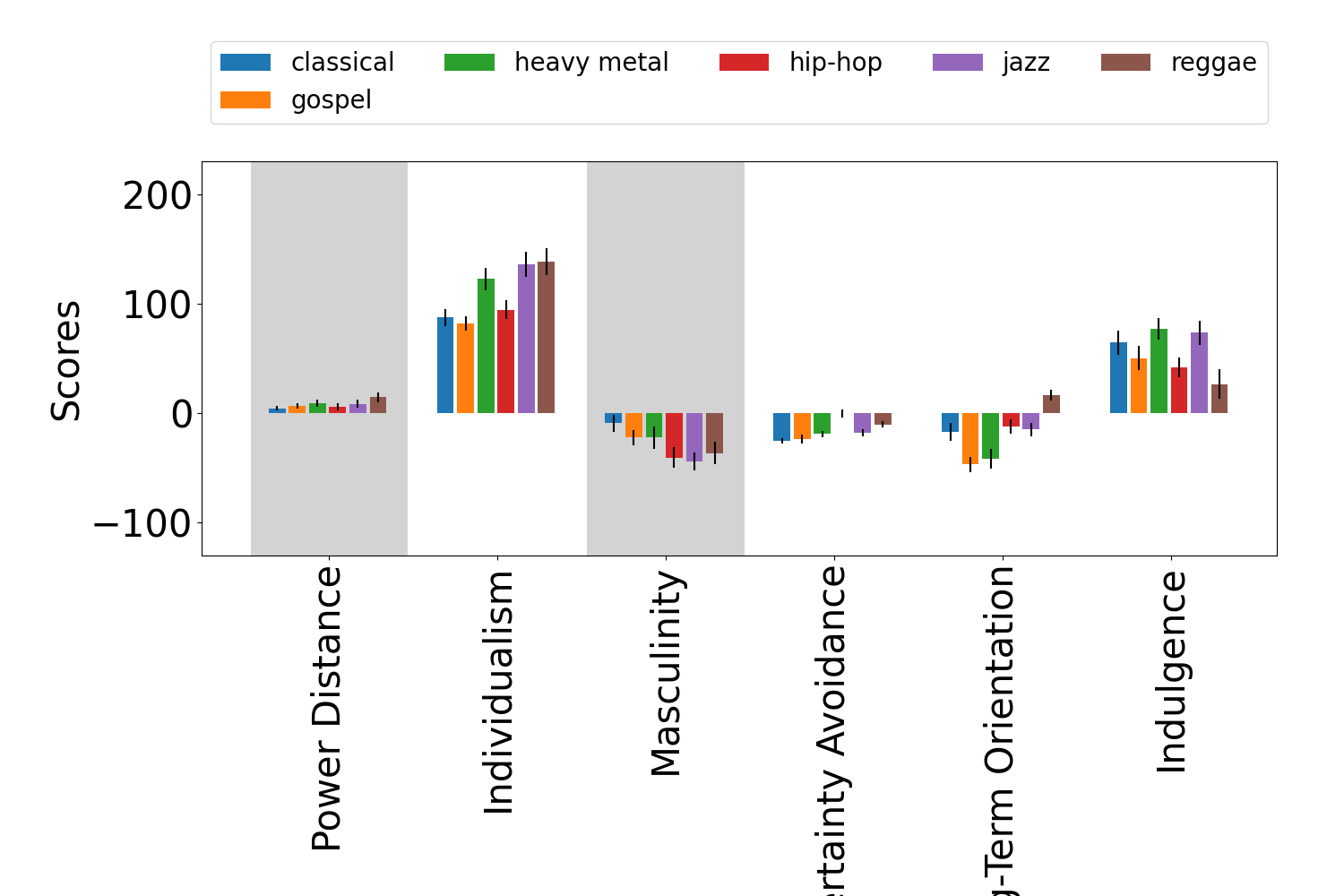}
\caption{}
\label{fig:wiki_hof}
\end{subfigure}

\caption{\textbf{Evidence for the unexpected perspective shift effect.} 
The effect of different simulated conversations on: (a) basic personal values, and (b) cultural values.
The effect of different textual formats on: (c) basic personal values, and (d) cultural values.
The effect of Wikipedia paragraphs about different music genres: (e) basic personal values, and (f) cultural values.
Although these contexts seem orthogonal to the tested values, we found them to cause significant effects on all personal values expressed by ChatGPT except those denoted by a gray background (ANOVA tests). Varying the context (e.g. from Python code questions to C++ code questions, or from jazz music context to gospel context) sometimes leads to large shifts in expressed values (e.g. \textit{achievement} and \textit{stimulation} respectively).}
\label{fig:unexpected_exp}
\end{figure}

Figure~\ref{fig:unexpected_exp} shows scores on the PVQ and VSM questionnaires for three different ways of varying the context.
Do these context changes significantly impact personal values?
For all experiments, we run one-way ANOVA analyses for each personal value. We use $\alpha<0.05$ to which a Bonferroni correction is applied for multiple comparisons ($\alpha<0.005$ for PVQ, and $\alpha<0.0083$ for VSM). 
We adjusted the significance level of statistical tests, \textalpha= 0.05, with the Bonferroni correction (actual significance levels \textalpha= 0.005 for PVQ and \textalpha= 0.0083 for VSM).
In the \textit{Simulated conversation} (\ref{fig:sim_conv_pvq} and \ref{fig:sim_conv_hof}) study,
we found that conversational topics induce a significant shift in the score distributions of all personal values and all cultural values (Fig~\ref{fig:sim_conv_hof}).
In the \textit{Text formats} study, contexts were again found to significantly impact the score distribution of all personal values and all cultural values (\ref{fig:formats_hof}).
In the \textit{Wikipedia paragraphs} study, contexts were found to significantly impact the score distribution for all personal values except \textit{Power} (Fig.~\ref{fig:wiki_pvq}), and for all cultural values except \textit{Power Distance} and \textit{Masculinity} (Fig.~\ref{fig:wiki_hof}). 
Changing perspectives often results significant changes in the expression of values. 
For example, a conversation about telling jokes increased the \textit{Indulgence} cultural value compared to the one about grammar, and asking questions through the C++ format compared to chat decreased \textit{Benevolence}, \textit{Universalism} and \textit{Self-Direction}.
Such results are  backed by posthoc Tukey HSD tests with $p<0.05$ adjusted by a Bonferroni correction to $p<0.005$ for PVQ and to $p<0.0083$ for VSM.
Refer to Figure \ref{fig:unexpected_exp_big5} in the Appendix for results on the IPIP questionnaire.

These effects are significant even though they were not induced on purpose and would be hard to predict.
These experiments show that LLMs' expression of values can be strongly affected by the presentation of contextual information that seems orthogonal to values.
This experiment provided empirical evidence for the \textit{unexpected perspective shift effects} in the context of personal and cultural value expression. 
In humans, the expression of personal values was found to be quite robust across the lifespan \citep{values_stable}.
In contrast, LLMs seem to shift their values as a function of contexts that seem irrelevant to these personal values.
This is a strong argument against perceiving LLMs as \textit{individuals}.
These conclusions question the interpretation of studies using questionnaires (and more generally benchmarks) to draw conclusions about an LLM's general values, personality traits, knowledge or capabilities \citep{Binz2022UsingCP,Miotto2022-jq,sap2022neural}.
The values expressed by an LLM in a single context seem to say little about what it would express in another context. These results call for more careful studies of how these traits and capabilities vary as a function of contexts. 

In this section, we presented evidence for the existence of the \textit{unexpected perspective shift effect} in
ChatGPT. 
In appendix \ref{app:human_studies} we reanalyse these results in the context changes observed in psychology studies with humans.
We study three different types of value stability: mean-level change, rank-order stability, and within-person (ipstative) change. We demonstrate that value change in ChatGPT is often much bigger than that in humans despite human value change being induced by much more drastic scenarios (e.g. 8 years of early adulthood development).
Following this, in appendix \ref{app:unexp_sys}, we systematically compare various large language models along those three types of value stability.


\subsection{How do different models compare in terms of their perspective controllability?}
\label{sec:exp_contr_sys}

This section focuses on the \textit{expected} context-based perspective changes.
We systematically compare the perspective controllability of different language models, i.e. their capacity to express the values and personality traits we try to induce.
We measure the controllability of 16 language models (see details in Appendix Section~\ref{sec:models}) using four different perspective induction methods with respect to the values and personality traits measured by the PVQ, VSM and IPI questionnaire.
In Appendix~\ref{app:unexp_sys} we conduct an analogous systematic analysis regarding the unexpected perspective changes.

We induce perspectives in four different ways: either via a standard \textit{user message} input or via the \textit{system message} input when available and use either the 2\textsuperscript{nd} or the 3\textsuperscript{rd} person (e.g. ``You are a person attributing extremely more importance to high individualism" versus ``The following are answers from a person attributing extremely more importance to high individualism"). Examples of prompts corresponding to these four methods are shown in Figures \ref{fig:2nd_3rd_prompts} and \ref{fig:user_system_prompts} in the Appendix. 
More expensive models (GPT-4 and Davinci-3) were first evaluated with ten permutations in the order of suggested answers. 
For GPT-4, the most controllable perspective induction method was selected and ran on 50 permutations as well. As Davinci-3 did not exhibit high controllability it was not run on 50 permutations.

Table~\ref{tab:contr_sys_exp} compares the perspective controllability of various models under the four induction methods for each of the three questionnaires.
We use the Welch t-test ($p<0.05$ to which a Bonferroni correction is applied resulting in $p<0.003$) to compare models.
For each questionnaire, we compare the most controllable model to all other models, but we only consider the most controllable induction method for each model.
(the statistical analysis results are shown in Appendix Table \ref{tab:stat_sys_contr}).
On PVQ, GPT-3.5-0301 with perspective induction using the 2\textsuperscript{nd}  person in the \textit{system message} scores are significantly higher than the best induction method in all other models besides GPT-3.5-0613.
On VSM, Upstage-LLaMa-instruct model (user message, 3\textsuperscript{rd} person) scores the highest and significantly better than the best induction method in all other models besides GPT-3.5-0314 and Upstage-LLaMa-2-instruct.
On IPIP, GPT-3.5-0613 (system message, 3\textsuperscript{rd} person) scores the highest and significantly better than the best induction methods all other models except GPT-4-0314, GPT-3.5-0301 and both Upstage-LLama models. 
Refer to Appendix Figures \ref{fig:best_pvq}, \ref{fig:best_vsm}, and \ref{fig:best_ipip} for visualizations of the value profiles expressed by the most controllable models.

Comparing the controllability of GPT-3.5 from june (GPT-3.5-0613) and march (GPT-3.5-0301) gives us insight into the effect of RLHF finetuning on controllability.
Controllability through the system message seems to have increased with 3\textsuperscript{rd} person induction for PVQ and IPIP, and for both 2\textsuperscript{nd} and 3\textsuperscript{rd} person for VSM between may and june, while controllability through the user message decreased on all settings except 3\textsuperscript{rd} person on VSM.
This implies that the RLHF finetuning may have resulted in a shift of controllability from the user message to the system message. 

Overall, higher perspective controllability can be observed in GPT models trained using reinforcement learning from human feedback (RLHF), as well as Upstage LLaMa models.
No induction method proved to be consistently better in all models and questionnaires, implying that the choice of the best induction methods largely depends on both the problem and language model.

\begin{table}[h!t]
    \caption{\textbf{Systematic comparison of the language models' perspective controllability.}
    Controllability measures how much each model expresses the values or personality traits explicitly targeted by the induced perspective.
    For each of the three questionnaires and each of the 16 models, we report the controllability scores for 4 perspective induction techniques (2\textsuperscript{nd} vs 3\textsuperscript{rd} person and message in the \textit{system} vs \textit{user} input).
    The most controllable model for each questionnaire is marked in bold.
    The scores for the most controllable models are shown in Appendix Figures \ref{fig:best_pvq}, \ref{fig:best_vsm}, 
    \ref{fig:best_ipip}, and statistical analysis in Appendix Table \ref{tab:stat_sys_contr}.
    }
    \begin{center}
    \scalebox{0.8}{
    \begin{tabular}{lcc|cc|cc}
        \mlcell{} & \multicolumn{2}{c|}{\bf PVQ (Schwartz)} & \multicolumn{2}{c|}{\bf VSM (Hofstede)} & \multicolumn{2}{c}{\bf IPIP (Big 5)} \\
        \cmidrule(lr){2-3} \cmidrule(lr){4-5} \cmidrule(lr){6-7}
        & System msg & User msg & System msg & User msg & System msg & User msg \\
        & 2nd \textbar 3rd & 2nd \textbar 3rd & 2nd \textbar 3rd & 2nd \textbar 3rd & 2nd \textbar 3rd & 2nd \textbar 3rd \\
        \midrule
        \multicolumn{2}{l}{\textit{10 permutations}} & & & & & \\
        GPT-4-0314 & .462 \textbar .488 & .419 \textbar .445 & .256 \textbar .263 & .225 \textbar \textbf{.279} & .35 \textbar .358 & .355 \textbar .368 \\
        GPT-3.5-0301 & .621 \textbar .539 & \textbf{.626} \textbar .547 & .101 \textbar .151 & .189 \textbar .165 & .354 \textbar .38 & .383 \textbar \textbf{.388} \\
        Davinci-003 & n/a & .03 \textbar .068 & n/a & -.005  \textbar .085 & n/a & .02 \textbar .117 \\ 
        \midrule
        \multicolumn{2}{l}{\textit{50 permutations}} & & & & & \\
        GPT-4-0314 &  - \textbar .518 & - \textbar - & - \textbar - & -  \textbar .258 & - \textbar - & - \textbar .376 \\ 
        \bf GPT-3.5-0301  & \textbf{.681} \textbar .561 & .64 \textbar .564 & .118 \textbar .147 & .184 \textbar .162 & .331 \textbar .334 & .379 \textbar .343 \\
        \bf GPT-3.5-0613  & .68 \textbar .624 & .552 \textbar .45 & .188 \textbar .196 & .175 \textbar .175 & .333 \textbar \textbf{.4} & .264 \textbar .332 \\
        Upst-LLaMa-2-70B-instruct & .494 \textbar .478 & .517 \textbar .448 & .228 \textbar .251 & .232  \textbar .263 & .344 \textbar .379 & .328 \textbar .379 \\ 
        \bf Upst-LLaMa-66B-instruct   & .507 \textbar .489 & .52  \textbar .457 & .239 \textbar .239 & .238  \textbar \bf .265 & .338 \textbar .383 & .325 \textbar .388 \\ 
        Zepyr-7B-beta & .548 \textbar .493 & .531 \textbar .383 & .089 \textbar .108 & .092 \textbar .117 & .159 \textbar .235 & .169 \textbar .248 \\
        OA  & .124 \textbar .140 & .196 \textbar .129 & .006 \textbar .021 & .029 \textbar .036 & .062 \textbar .049 & .057 \textbar .099 \\
        StLM  & -.006 \textbar -.002 & .006 \textbar -.0 & -.004 \textbar .004 & .003 \textbar .004 & .0 \textbar .0 & .004 \textbar -.001 \\
        LLaMa-65B & n/a & .092 \textbar .06 & n/a & .017 \textbar .082 & n/a & .047 \textbar .109 \\ 
        StVicuna  & n/a & .066 \textbar .034 & n/a & -.002 \textbar .005 & n/a & .043 \textbar .067 \\
        Redpaj-incite-chat & n/a & .0001 \textbar -.004 & n/a & .0002 \textbar .0002 & n/a & -.001 \textbar .003 \\ 
        Redpaj-incite-instruct & n/a & .007 \textbar .0001 & n/a & -.001  \textbar .0002 & n/a & .018 \textbar .0 \\ 
        GPT-3.5-instruct-0914  & n/a & .0 \textbar .155    & n/a & .005 \textbar .042 &  n/a & .004 \textbar .096 \\
        Curie         & n/a & -.004 \textbar -.004  & n/a & -.004 \textbar -.004 & n/a & .001 \textbar .001 \\ 
        Babbage & n/a & .003 \textbar -.002 & n/a & .0002  \textbar .0002 & n/a & .0 \textbar .002 \\ 
        Ada & n/a & -.001 \textbar -.001 & n/a & .003  \textbar -.0001 & n/a & .002 \textbar .002 \\ 
        \bottomrule
    \end{tabular}
    }
    \end{center}
    \label{tab:contr_sys_exp}
\end{table}


\section{Discussion}

This paper showed evidence for the \textit{unexpected perspective shift effect}, the ubiquitous shifts in the expression of values by LLMs as a function of what seems to be orthogonal changes in context. Humans, on the other hand, express personal values that are stable over time \citep{values_stable}. This gives us ground to reject the pervasive ``LLM as individual'' metaphor and replace it with a new one: ``LLM as a superposition of perspectives.''

This change has important consequences.
Indeed recent works have reemployed batteries of tests and questionnaires developed in the human psychology literature \citep{Binz2022UsingCP,Miotto2022-jq,holterman2023does}.
Psychology questionnaires, and standard NLP benchmarks, usually present questions from a single perspective such as a multiple choice question (e.g. MMLU \citep{mmlu}).
This means that behavior exhibited on a questionnaire says little about potential behaviors in other contexts.
The problem is further exacerbated in the case of psychological questionnaires as they were created under the assumption of human robustness to context change and internal consistency.
But as we demonstrated, LLMs are not like humans and their strong and counter-intuitive context dependency violates those assumptions.
This questions the interpretation of such experiments, and asks for further research into how expression of values, behaviors and capacities varies due to expected and unexpected context changes.

Under the ``LLM as a superposition of perspectives'' metaphor new scientific questions arise. How can we influence and control such perspectives changes, i.e. how can we modify the context in order to induce a target perspective? Our experiments show that different perspective induction methods lead to different results. Different language models also demonstrate different perspective controllability levels. This shows the difficulty of properly controlling the perspective language models take when generating new text and calls for further research in that area.

\section*{Ethics Statement}
The understanding that LLMs encode a diversity of cultural values and perspectives introduces the question of which values one should build into LLMs: should we aim to represent a large diversity of cultures or try to align a model with one set of values? 
This is a nuanced question, as different values systems can often be conflicting.
\cite{johnson2022ghost} provide an in-depth discussion of this problem and suggests that LLMs should be aligned with basic human rights, but also be able to deal with conflicting value systems.
The solution to this problem also depends on the practical application and on stakeholders \citep{bender2021dangers,jernite2022data}.
Some applications may require models which are not malleable in terms of potential values expressed, while others may require highly controllable ones.
This paper adds to this discussion by providing an intuitive conceptualization of this issue (LLMs as superpositions of perspectives), and by introducing the concept of perspective controllability. 

After deciding on the target values and controllability levels a model should have for some application, a series of scientific questions arise.
First, how could one build that model? 
The ROOTS corpus \citep{laurenccon2022bigscience} includes 47 natural languages.
While this hints at a degree of cultural diversity, a proper representation of cultural diversity will require a detailed analysis of the cultural dimensions contained within such corpora.
On the other hand, ConstitutionalAI \citep{bai2022constitutional} is a method for aligning the model with a set of explicitly defined principles.
One can see ROOTS as an attempt to increase the controllability of cultural perspectives, and ConstitutionalAI as an attempt to reduce the controllability (especially regarding values not aligned with the defined principles). Another interesting question is whether all cultural perspectives expressed by LLMs are encoded in the training data, or whether some can be 'hallucinated'. The latter case would imply that a thorough analysis cannot be done solely by analyzing the cultural diversity of datasets or human feedback.
This calls for developing methods for evaluating the diversity and controllability of cultural perspectives encoded in LLMs beyond datasets, i.e. in their actual behavior. Here, we used a few simple questionnaires from psychology, but a lot of research remains to be done for more systematic and automatic methods. 
For example, current benchmarks present many questions from a single perspective (e.g. MCQ). New benchmarks presenting same questions from many different perspectives would be an essential tool to compare models' on their propensity to the unexpected perspective shift effect.



\section*{Reproducibility}
Our experiments rely on several LLMs from the OpenAI API \citep{openai_api} and various open-source models (see section \ref{sec:models} for details) and will therefore be reproducible to the extent that these models remain accessible.
All our code and experiments are open sourced at the project website  \url{https://sites.google.com/view/llm-superpositions}.



\bibliography{iclr2024_conference}

\begin{thebibliography}{70}
\providecommand{\natexlab}[1]{#1}
\providecommand{\url}[1]{\texttt{#1}}
\expandafter\ifx\csname urlstyle\endcsname\relax
  \providecommand{\doi}[1]{doi: #1}\else
  \providecommand{\doi}{doi: \begingroup \urlstyle{rm}\Url}\fi

\bibitem[Aher et~al.(2022)Aher, Arriaga, and Kalai]{Aher2022-tj}
Gati Aher, RosaI. Arriaga, and Adam~Tauman Kalai.
\newblock Using large language models to simulate multiple humans and replicate human subject studies.
\newblock In \emph{International Conference on Machine Learning}, 2022.
\newblock URL \url{https://api.semanticscholar.org/CorpusID:251719353}.

\bibitem[Andreas(2022)]{Andreas2022-wr}
Jacob Andreas.
\newblock Language models as agent models.
\newblock In Yoav Goldberg, Zornitsa Kozareva, and Yue Zhang (eds.), \emph{Findings of the Association for Computational Linguistics: EMNLP 2022}, pp.\  5769--5779, Abu Dhabi, United Arab Emirates, December 2022. Association for Computational Linguistics.
\newblock \doi{10.18653/v1/2022.findings-emnlp.423}.
\newblock URL \url{https://aclanthology.org/2022.findings-emnlp.423}.

\bibitem[Argyle et~al.(2023)Argyle, Busby, Fulda, Gubler, Rytting, and Wingate]{argyle2023out}
Lisa~P. Argyle, Ethan~C. Busby, Nancy Fulda, Joshua~R. Gubler, Christopher Rytting, and David Wingate.
\newblock Out of one, many: Using language models to simulate human samples.
\newblock \emph{Political Analysis}, 31\penalty0 (3):\penalty0 337--351, 2023.

\bibitem[Arieli et~al.(2014)Arieli, Grant, and Sagiv]{arieli2014}
Sharon Arieli, Adam~M. Grant, and Lilach Sagiv.
\newblock Convincing yourself to care about others: An intervention for enhancing benevolence values.
\newblock \emph{Journal of Personality}, 82\penalty0 (1):\penalty0 15--24, 2014.
\newblock \doi{https://doi.org/10.1111/jopy.12029}.
\newblock URL \url{https://onlinelibrary.wiley.com/doi/abs/10.1111/jopy.12029}.

\bibitem[Arora et~al.(2023)Arora, Kaffee, and Augenstein]{Arora2022-ge}
Arnav Arora, Lucie-aim{\'e}e Kaffee, and Isabelle Augenstein.
\newblock Probing pre-trained language models for cross-cultural differences in values.
\newblock In Sunipa Dev, Vinodkumar Prabhakaran, David Adelani, Dirk Hovy, and Luciana Benotti (eds.), \emph{Proceedings of the First Workshop on Cross-Cultural Considerations in NLP (C3NLP)}, pp.\  114--130, Dubrovnik, Croatia, May 2023. Association for Computational Linguistics.
\newblock \doi{10.18653/v1/2023.c3nlp-1.12}.
\newblock URL \url{https://aclanthology.org/2023.c3nlp-1.12}.

\bibitem[Ashton \& Lee(2009)Ashton and Lee]{hexaco}
Michael~C. Ashton and Kibeom Lee.
\newblock The hexaco–60: A short measure of the major dimensions of personality.
\newblock \emph{Journal of Personality Assessment}, 91\penalty0 (4):\penalty0 340--345, 2009.
\newblock \doi{10.1080/00223890902935878}.
\newblock URL \url{https://www.tandfonline.com/doi/abs/10.1080/00223890902935878}.

\bibitem[Bai et~al.(2022)Bai, Kadavath, Kundu, Askell, Kernion, Jones, Chen, Goldie, Mirhoseini, McKinnon, et~al.]{bai2022constitutional}
Yuntao Bai, Saurav Kadavath, Sandipan Kundu, Amanda Askell, Jackson Kernion, Andy Jones, Anna Chen, Anna Goldie, Azalia Mirhoseini, Cameron McKinnon, et~al.
\newblock Constitutional ai: Harmlessness from ai feedback.
\newblock \emph{arXiv preprint arXiv:2212.08073}, 2022.

\bibitem[Bender et~al.(2021)Bender, Gebru, McMillan-Major, and Shmitchell]{bender2021dangers}
Emily~M. Bender, Timnit Gebru, Angelina McMillan-Major, and Shmargaret Shmitchell.
\newblock On the dangers of stochastic parrots: Can language models be too big?
\newblock In \emph{Proceedings of the 2021 ACM conference on fairness, accountability, and transparency}, pp.\  610--623, 2021.

\bibitem[Binz \& Schulz(2022)Binz and Schulz]{Binz2022UsingCP}
Marcel Binz and Eric Schulz.
\newblock Using cognitive psychology to understand gpt-3.
\newblock \emph{Proceedings of the National Academy of Sciences of the United States of America}, 120, 2022.
\newblock URL \url{https://api.semanticscholar.org/CorpusID:250113371}.

\bibitem[Binz \& Schulz(2023)Binz and Schulz]{binz2023turning}
Marcel Binz and Eric Schulz.
\newblock Turning large language models into cognitive models.
\newblock \emph{arXiv preprint arXiv:2306.03917}, 2023.

\bibitem[Brown et~al.(2020)Brown, Mann, Ryder, Subbiah, Kaplan, Dhariwal, Neelakantan, Shyam, Sastry, Askell, et~al.]{gpt3}
Tom Brown, Benjamin Mann, Nick Ryder, Melanie Subbiah, Jared~D Kaplan, Prafulla Dhariwal, Arvind Neelakantan, Pranav Shyam, Girish Sastry, Amanda Askell, et~al.
\newblock Language models are few-shot learners.
\newblock \emph{Advances in neural information processing systems}, 33:\penalty0 1877--1901, 2020.

\bibitem[Cao et~al.(2023)Cao, Zhou, Lee, Cabello, Chen, and Hershcovich]{Cao2023-gz}
Yong Cao, Li~Zhou, Seolhwa Lee, Laura Cabello, Min Chen, and Daniel Hershcovich.
\newblock Assessing cross-cultural alignment between {C}hat{GPT} and human societies: An empirical study.
\newblock In Sunipa Dev, Vinodkumar Prabhakaran, David Adelani, Dirk Hovy, and Luciana Benotti (eds.), \emph{Proceedings of the First Workshop on Cross-Cultural Considerations in NLP (C3NLP)}, pp.\  53--67, Dubrovnik, Croatia, May 2023. Association for Computational Linguistics.
\newblock \doi{10.18653/v1/2023.c3nlp-1.7}.
\newblock URL \url{https://aclanthology.org/2023.c3nlp-1.7}.

\bibitem[Chiang et~al.(2023)Chiang, Li, Lin, Sheng, Wu, Zhang, Zheng, Zhuang, Zhuang, Gonzalez, Stoica, and Xing]{vicuna2023}
Wei-Lin Chiang, Zhuohan Li, Zi~Lin, Ying Sheng, Zhanghao Wu, Hao Zhang, Lianmin Zheng, Siyuan Zhuang, Yonghao Zhuang, Joseph~E. Gonzalez, Ion Stoica, and Eric~P. Xing.
\newblock Vicuna: An open-source chatbot impressing gpt-4 with 90\%* chatgpt quality, March 2023.
\newblock URL \url{https://lmsys.org/blog/2023-03-30-vicuna/}.

\bibitem[Christiano et~al.(2017)Christiano, Leike, Brown, Martic, Legg, and Amodei]{rlhf}
Paul~F. Christiano, Jan Leike, Tom Brown, Miljan Martic, Shane Legg, and Dario Amodei.
\newblock Deep reinforcement learning from human preferences.
\newblock \emph{Advances in neural information processing systems}, 30, 2017.

\bibitem[Cieciuch \& Schwartz(2012)Cieciuch and Schwartz]{Cieciuch2012-bg}
Jan Cieciuch and Shalom~H. Schwartz.
\newblock The number of distinct basic values and their structure assessed by {PVQ--40}.
\newblock \emph{J. Pers. Assess.}, 94\penalty0 (3):\penalty0 321--328, May 2012.

\bibitem[Computer(2023{\natexlab{a}})]{rpchat}
Together Computer.
\newblock Hugging face model hub: Redpajama-incite-7b-chat.
\newblock \url{https://huggingface.co/togethercomputer/RedPajama-INCITE-7B-Chat}, 2023{\natexlab{a}}.
\newblock Accessed: October 7, 2023.

\bibitem[Computer(2023{\natexlab{b}})]{rpinstrct}
Together Computer.
\newblock Hugging face model hub: Redpajama-incite-7b-instruct.
\newblock \url{https://huggingface.co/togethercomputer/RedPajama-INCITE-7B-Instruct}, 2023{\natexlab{b}}.
\newblock Accessed: October 7, 2023.

\bibitem[Costa \& McCrae(2008)Costa and McCrae]{Costa2008-sv}
Paul Costa and Robert McCrae.
\newblock The revised neo personality inventory (neo-pi-r).
\newblock \emph{The SAGE Handbook of Personality Theory and Assessment}, 2:\penalty0 179--198, 01 2008.
\newblock \doi{10.4135/9781849200479.n9}.

\bibitem[Costa \& McCrae(1992)Costa and McCrae]{Costa1992-qu}
Paul~T Costa and Robert~R McCrae.
\newblock Normal personality assessment in clinical practice: The {NEO} personality inventory.
\newblock \emph{Psychol. Assess.}, 4\penalty0 (1):\penalty0 5--13, March 1992.

\bibitem[Deshpande et~al.(2023)Deshpande, Murahari, Rajpurohit, Kalyan, and Narasimhan]{Deshpande2023-ui}
Ameet Deshpande, Vishvak Murahari, Tanmay Rajpurohit, Ashwin Kalyan, and Karthik Narasimhan.
\newblock Toxicity in {ChatGPT}: Analyzing persona-assigned language models.
\newblock April 2023.

\bibitem[Döring \& Hillbrink(2015)Döring and Hillbrink]{doring2015}
Anna~K. Döring and Alessa Hillbrink.
\newblock Brief report: Into the wild? how a film can change adolescents' values.
\newblock \emph{Journal of Adolescence}, 40\penalty0 (1):\penalty0 78--82, 2015.
\newblock \doi{https://doi.org/10.1016/j.adolescence.2015.01.006}.
\newblock URL \url{https://onlinelibrary.wiley.com/doi/abs/10.1016/j.adolescence.2015.01.006}.

\bibitem[Face(2023)]{huggingfacemodelhub}
Hugging Face.
\newblock Hugging face model hub.
\newblock \url{https://huggingface.co/models}, 2023.
\newblock Accessed: May 13, 2023.

\bibitem[Goldberg(1990)]{Goldberg1990-jw}
Lewis~R. Goldberg.
\newblock An alternative ``description of personality'': The {Big-Five} factor structure.
\newblock \emph{J. Pers. Soc. Psychol.}, 59\penalty0 (6):\penalty0 1216--1229, 1990.

\bibitem[Goldberg(1993)]{goldberg1993structure}
Lewis~R. Goldberg.
\newblock The structure of phenotypic personality traits.
\newblock \emph{American Psychologist}, 48\penalty0 (1):\penalty0 26--34, 1993.
\newblock \doi{10.1037/0003-066X.48.1.26}.

\bibitem[Goldberg(1999{\natexlab{a}})]{Goldberg1999-ob}
Lewis~R. Goldberg.
\newblock A broad-bandwidth, public domain, personality inventory measuring the lower-level facets of several five-factor models.
\newblock \emph{Personality psychology in Europe}, 7\penalty0 (1):\penalty0 7--28, 1999{\natexlab{a}}.

\bibitem[Goldberg(1999{\natexlab{b}})]{goldberg1999_ipip}
Lewis~R. Goldberg.
\newblock A broad-bandwidth, public domain, personality inventory measuring the lower-level facets of several five-factor models.
\newblock \emph{Personality psychology in Europe}, 7\penalty0 (1):\penalty0 7--28, 1999{\natexlab{b}}.

\bibitem[Griffin et~al.(2023)Griffin, Kleinberg, Mozes, Mai, Vau, Caldwell, and Marvor-Parker]{griffin2023susceptibility}
Lewis~D. Griffin, Bennett Kleinberg, Maximilian Mozes, Kimberly~T. Mai, Maria Vau, Matthew Caldwell, and Augustine Marvor-Parker.
\newblock Susceptibility to influence of large language models.
\newblock \emph{arXiv preprint arXiv:2303.06074}, 2023.

\bibitem[Guilford(1967)]{aut}
Joy~P. Guilford.
\newblock Creativity: Yesterday, today and tomorrow.
\newblock \emph{The Journal of Creative Behavior}, 1\penalty0 (1):\penalty0 3--14, 1967.

\bibitem[Hendrycks et~al.(2021)Hendrycks, Burns, Basart, Zou, Mazeika, Song, and Steinhardt]{mmlu}
Dan Hendrycks, Collin Burns, Steven Basart, Andy Zou, Mantas Mazeika, Dawn Song, and Jacob Steinhardt.
\newblock Measuring massive multitask language understanding.
\newblock \emph{Proceedings of the International Conference on Learning Representations (ICLR)}, 2021.

\bibitem[Hershcovich et~al.(2022)Hershcovich, Frank, Lent, de~Lhoneux, Abdou, Brandl, Bugliarello, Cabello~Piqueras, Chalkidis, Cui, Fierro, Margatina, Rust, and S{\o}gaard]{hershcovich2022challenges}
Daniel Hershcovich, Stella Frank, Heather Lent, Miryam de~Lhoneux, Mostafa Abdou, Stephanie Brandl, Emanuele Bugliarello, Laura Cabello~Piqueras, Ilias Chalkidis, Ruixiang Cui, Constanza Fierro, Katerina Margatina, Phillip Rust, and Anders S{\o}gaard.
\newblock Challenges and strategies in cross-cultural {NLP}.
\newblock In Smaranda Muresan, Preslav Nakov, and Aline Villavicencio (eds.), \emph{Proceedings of the 60th Annual Meeting of the Association for Computational Linguistics (Volume 1: Long Papers)}, pp.\  6997--7013, Dublin, Ireland, May 2022. Association for Computational Linguistics.
\newblock \doi{10.18653/v1/2022.acl-long.482}.
\newblock URL \url{https://aclanthology.org/2022.acl-long.482}.

\bibitem[Hofstede(2013)]{hofstede_vsm_2013}
Geert Hofstede.
\newblock Values survey module 2013, 2013.
\newblock URL \url{https://geerthofstede.com/wp-content/uploads/2016/07/VSM-2013-English-2013-08-25.pdf}.
\newblock Accessed: April 16, 2023.

\bibitem[Hofstede \& Bond(1984)Hofstede and Bond]{hofstede1984-gz}
Geert Hofstede and Michael~H Bond.
\newblock Hofstede's culture dimensions: An independent validation using rokeach's value survey.
\newblock \emph{J. Cross. Cult. Psychol.}, 15\penalty0 (4):\penalty0 417--433, December 1984.

\bibitem[Hofstede \& Minkov(2013)Hofstede and Minkov]{hofstede_vsm_2013_manual}
Geert Hofstede and Michael Minkov.
\newblock Values survey module 2013 manual, 2013.
\newblock URL \url{https://geerthofstede.com/wp-content/uploads/2016/07/Manual-VSM-2013.pdf}.
\newblock Accessed: April 16, 2023.

\bibitem[Hofstede et~al.(1990)Hofstede, Neuijen, Ohayv, and Sanders]{hofstede1990-gv}
Geert Hofstede, Bram Neuijen, Denise~Daval Ohayv, and Geert Sanders.
\newblock Measuring organizational cultures: A qualitative and quantitative study across twenty cases.
\newblock \emph{Adm. Sci. Q.}, 35\penalty0 (2):\penalty0 286--316, 1990.

\bibitem[Hoftede et~al.(2010)Hoftede, Hofstede, and Minkov]{hofstede_dimensions}
Geert Hoftede, Gert~Jan Hofstede, and Michael Minkov.
\newblock \emph{Cultures and organizations: software of the mind: intercultural cooperation and its importance for survival}.
\newblock McGraw-Hill, 2010.

\bibitem[Holterman \& van Deemter(2023)Holterman and van Deemter]{holterman2023does}
Bart Holterman and Kees van Deemter.
\newblock Does chatgpt have theory of mind?
\newblock \emph{arXiv preprint arXiv:2305.14020}, 2023.

\bibitem[Jernite et~al.(2022)Jernite, Nguyen, Biderman, Rogers, Masoud, Danchev, Tan, Luccioni, Subramani, Johnson, Dupont, Dodge, Lo, Talat, Radev, Gokaslan, Nikpoor, Henderson, Bommasani, and Mitchell]{jernite2022data}
Yacine Jernite, Huu Nguyen, Stella Biderman, Anna Rogers, Maraim Masoud, Valentin Danchev, Samson Tan, Alexandra~Sasha Luccioni, Nishant Subramani, Isaac Johnson, Gerard Dupont, Jesse Dodge, Kyle Lo, Zeerak Talat, Dragomir Radev, Aaron Gokaslan, Somaieh Nikpoor, Peter Henderson, Rishi Bommasani, and Margaret Mitchell.
\newblock Data governance in the age of large-scale data-driven language technology.
\newblock In \emph{2022 {ACM} Conference on Fairness, Accountability, and Transparency}. {ACM}, jun 2022.
\newblock \doi{10.1145/3531146.3534637}.
\newblock URL \url{https://doi.org/10.1145%2F3531146.3534637}.

\bibitem[Jiang et~al.(2023)Jiang, Xu, Zhu, Han, Zhang, and Zhu]{jiang2022mpi}
Guangyuan Jiang, Manjie Xu, Song-Chun Zhu, Wenjuan Han, Chi Zhang, and Yixin Zhu.
\newblock Evaluating and inducing personality in pre-trained language models.
\newblock In \emph{Thirty-seventh Conference on Neural Information Processing Systems}, 2023.
\newblock URL \url{https://openreview.net/forum?id=I9xE1Jsjfx}.

\bibitem[John \& Srivastava(1999)John and Srivastava]{bfi}
Oliver~P. John and Sanjay Srivastava.
\newblock The big five trait taxonomy: History, measurement, and theoretical perspectives.
\newblock In Lawrence~A. Pervin and Oliver~P. John (eds.), \emph{Handbook of personality: Theory and research}, pp.\  102--138. Guilford Press, 1999.

\bibitem[Johnson et~al.(2022)Johnson, Pistilli, Men{\'e}dez-Gonz{\'a}lez, Duran, Panai, Kalpokiene, and Bertulfo]{johnson2022ghost}
Rebecca~L Johnson, Giada Pistilli, Natalia Men{\'e}dez-Gonz{\'a}lez, Leslye Denisse~Dias Duran, Enrico Panai, Julija Kalpokiene, and Donald~Jay Bertulfo.
\newblock The ghost in the machine has an american accent: value conflict in gpt-3.
\newblock \emph{arXiv preprint arXiv:2203.07785}, 2022.

\bibitem[Jones \& Paulhus(2014)Jones and Paulhus]{sd3}
Daniel~N. Jones and Delroy~L. Paulhus.
\newblock Introducing the short dark triad (sd3): A brief measure of dark personality traits.
\newblock \emph{Assessment}, 21\penalty0 (1):\penalty0 28--41, 2014.
\newblock \doi{10.1177/1073191113514105}.
\newblock URL \url{https://doi.org/10.1177/1073191113514105}.
\newblock PMID: 24322012.

\bibitem[Kosinski(2023)]{kosinski2023theory}
Michal Kosinski.
\newblock Theory of mind may have spontaneously emerged in large language models.
\newblock \emph{arXiv preprint arXiv:2302.02083}, 2023.

\bibitem[Kosoy et~al.(2023)Kosoy, Reagan, Lai, Gopnik, and Cobb]{kosoy2023comparing}
Eliza Kosoy, Emily~Rose Reagan, Leslie Lai, Alison Gopnik, and Danielle~Krettek Cobb.
\newblock Comparing machines and children: Using developmental psychology experiments to assess the strengths and weaknesses of lamda responses.
\newblock \emph{arXiv preprint arXiv:2305.11243}, 2023.

\bibitem[Lauren{\c{c}}on et~al.(2022)Lauren{\c{c}}on, Saulnier, Wang, Akiki, Villanova~del Moral, Le~Scao, Von~Werra, Mou, Gonz{\'a}lez~Ponferrada, Nguyen, et~al.]{laurenccon2022bigscience}
Hugo Lauren{\c{c}}on, Lucile Saulnier, Thomas Wang, Christopher Akiki, Albert Villanova~del Moral, Teven Le~Scao, Leandro Von~Werra, Chenghao Mou, Eduardo Gonz{\'a}lez~Ponferrada, Huu Nguyen, et~al.
\newblock The bigscience roots corpus: A 1.6 tb composite multilingual dataset.
\newblock \emph{Advances in Neural Information Processing Systems}, 35:\penalty0 31809--31826, 2022.

\bibitem[Li et~al.(2022)Li, Li, Liu, Bing, and Joty]{li2022gpt}
Xingxuan Li, Yutong Li, Linlin Liu, Lidong Bing, and Shafiq Joty.
\newblock Is gpt-3 a psychopath? evaluating large language models from a psychological perspective.
\newblock \emph{arXiv preprint arXiv:2212.10529}, 2022.

\bibitem[Lu et~al.(2022)Lu, Bartolo, Moore, Riedel, and Stenetorp]{lu2021fantastically}
Yao Lu, Max Bartolo, Alastair Moore, Sebastian Riedel, and Pontus Stenetorp.
\newblock Fantastically ordered prompts and where to find them: Overcoming few-shot prompt order sensitivity.
\newblock In Smaranda Muresan, Preslav Nakov, and Aline Villavicencio (eds.), \emph{Proceedings of the 60th Annual Meeting of the Association for Computational Linguistics (Volume 1: Long Papers)}, pp.\  8086--8098, Dublin, Ireland, May 2022. Association for Computational Linguistics.
\newblock \doi{10.18653/v1/2022.acl-long.556}.
\newblock URL \url{https://aclanthology.org/2022.acl-long.556}.

\bibitem[Miotto et~al.(2022)Miotto, Rossberg, and Kleinberg]{Miotto2022-jq}
Maril{\`u} Miotto, Nicola Rossberg, and Bennett Kleinberg.
\newblock Who is {GPT}-3? an exploration of personality, values and demographics.
\newblock In David Bamman, Dirk Hovy, David Jurgens, Katherine Keith, Brendan O'Connor, and Svitlana Volkova (eds.), \emph{Proceedings of the Fifth Workshop on Natural Language Processing and Computational Social Science (NLP+CSS)}, pp.\  218--227, Abu Dhabi, UAE, November 2022. Association for Computational Linguistics.
\newblock \doi{10.18653/v1/2022.nlpcss-1.24}.
\newblock URL \url{https://aclanthology.org/2022.nlpcss-1.24}.

\bibitem[OpenAI(2023)]{OpenAIChatCompletionsAPI}
OpenAI.
\newblock Chat completions api guide, 2023.
\newblock URL \url{https://platform.openai.com/docs/guides/gpt/chat-completions-api}.
\newblock Accessed: September 26, 2023.

\bibitem[{OpenAI}(2023)]{gpt4}
{OpenAI}.
\newblock {GPT-4 Technical Report}.
\newblock \emph{arXiv e-prints}, art. arXiv:2303.08774, March 2023.
\newblock \doi{10.48550/arXiv.2303.08774}.

\bibitem[OpenAI(2023)]{openai_api}
OpenAI.
\newblock Openai api, 2023.
\newblock URL \url{https://openai.com/blog/openai-api}.
\newblock Accessed: September 26, 2023.

\bibitem[Ouyang et~al.(2022)Ouyang, Wu, Jiang, Almeida, Wainwright, Mishkin, Zhang, Agarwal, Slama, Ray, Schulman, Hilton, Kelton, Miller, Simens, Askell, Welinder, Christiano, Leike, and Lowe]{instructgpt}
Long Ouyang, Jeff Wu, Xu~Jiang, Diogo Almeida, Carroll~L. Wainwright, Pamela Mishkin, Chong Zhang, Sandhini Agarwal, Katarina Slama, Alex Ray, John Schulman, Jacob Hilton, Fraser Kelton, Luke~E. Miller, Maddie Simens, Amanda Askell, Peter Welinder, Paul~Francis Christiano, Jan Leike, and Ryan~J. Lowe.
\newblock Training language models to follow instructions with human feedback.
\newblock \emph{ArXiv}, abs/2203.02155, 2022.
\newblock URL \url{https://api.semanticscholar.org/CorpusID:246426909}.

\bibitem[Perez et~al.(2023)Perez, Ringer, Lukosiute, Nguyen, Chen, Heiner, Pettit, Olsson, Kundu, Kadavath, Jones, Chen, Mann, Israel, Seethor, McKinnon, Olah, Yan, Amodei, Amodei, Drain, Li, Tran-Johnson, Khundadze, Kernion, Landis, Kerr, Mueller, Hyun, Landau, Ndousse, Goldberg, Lovitt, Lucas, Sellitto, Zhang, Kingsland, Elhage, Joseph, Mercado, DasSarma, Rausch, Larson, McCandlish, Johnston, Kravec, El~Showk, Lanham, Telleen-Lawton, Brown, Henighan, Hume, Bai, Hatfield-Dodds, Clark, Bowman, Askell, Grosse, Hernandez, Ganguli, Hubinger, Schiefer, and Kaplan]{perez2023discovering}
Ethan Perez, Sam Ringer, Kamile Lukosiute, Karina Nguyen, Edwin Chen, Scott Heiner, Craig Pettit, Catherine Olsson, Sandipan Kundu, Saurav Kadavath, Andy Jones, Anna Chen, Benjamin Mann, Brian Israel, Bryan Seethor, Cameron McKinnon, Christopher Olah, Da~Yan, Daniela Amodei, Dario Amodei, Dawn Drain, Dustin Li, Eli Tran-Johnson, Guro Khundadze, Jackson Kernion, James Landis, Jamie Kerr, Jared Mueller, Jeeyoon Hyun, Joshua Landau, Kamal Ndousse, Landon Goldberg, Liane Lovitt, Martin Lucas, Michael Sellitto, Miranda Zhang, Neerav Kingsland, Nelson Elhage, Nicholas Joseph, Noemi Mercado, Nova DasSarma, Oliver Rausch, Robin Larson, Sam McCandlish, Scott Johnston, Shauna Kravec, Sheer El~Showk, Tamera Lanham, Timothy Telleen-Lawton, Tom Brown, Tom Henighan, Tristan Hume, Yuntao Bai, Zac Hatfield-Dodds, Jack Clark, Samuel~R. Bowman, Amanda Askell, Roger Grosse, Danny Hernandez, Deep Ganguli, Evan Hubinger, Nicholas Schiefer, and Jared Kaplan.
\newblock Discovering language model behaviors with model-written evaluations.
\newblock In Anna Rogers, Jordan Boyd-Graber, and Naoaki Okazaki (eds.), \emph{Findings of the Association for Computational Linguistics: ACL 2023}, pp.\  13387--13434, Toronto, Canada, July 2023. Association for Computational Linguistics.
\newblock \doi{10.18653/v1/2023.findings-acl.847}.
\newblock URL \url{https://aclanthology.org/2023.findings-acl.847}.

\bibitem[Rame et~al.(2023)Rame, Couairon, Shukor, Dancette, Gaya, Soulier, and Cord]{rame2023rewarded}
Alexandre Rame, Guillaume Couairon, Mustafa Shukor, Corentin Dancette, Jean-Baptiste Gaya, Laure Soulier, and Matthieu Cord.
\newblock Rewarded soups: towards pareto-optimal alignment by interpolating weights fine-tuned on diverse rewards.
\newblock \emph{arXiv preprint arXiv:2306.04488}, 2023.

\bibitem[Sagiv et~al.(2017)Sagiv, Roccas, Cieciuch, and Schwartz]{values_stable}
Lilach Sagiv, Sonia Roccas, Jan Cieciuch, and Shalom~H. Schwartz.
\newblock Personal values in human life.
\newblock \emph{Nature human behaviour}, 1\penalty0 (9):\penalty0 630--639, 2017.

\bibitem[Salewski et~al.(2023)Salewski, Alaniz, Rio-Torto, Schulz, and Akata]{salewski2023context}
Leonard Salewski, Stephan Alaniz, Isabel Rio-Torto, Eric Schulz, and Zeynep Akata.
\newblock In-context impersonation reveals large language models' strengths and biases.
\newblock \emph{arXiv preprint arXiv:2305.14930}, 2023.

\bibitem[Santurkar et~al.(2023)Santurkar, Durmus, Ladhak, Lee, Liang, and Hashimoto]{santurkar2023whose}
Shibani Santurkar, Esin Durmus, Faisal Ladhak, Cinoo Lee, Percy Liang, and Tatsunori Hashimoto.
\newblock Whose opinions do language models reflect?
\newblock \emph{arXiv preprint arXiv:2303.17548}, 2023.

\bibitem[Sap et~al.(2022)Sap, Le~Bras, Fried, and Choi]{sap2022neural}
Maarten Sap, Ronan Le~Bras, Daniel Fried, and Yejin Choi.
\newblock Neural theory-of-mind? on the limits of social intelligence in large {LM}s.
\newblock In Yoav Goldberg, Zornitsa Kozareva, and Yue Zhang (eds.), \emph{Proceedings of the 2022 Conference on Empirical Methods in Natural Language Processing}, pp.\  3762--3780, Abu Dhabi, United Arab Emirates, December 2022. Association for Computational Linguistics.
\newblock \doi{10.18653/v1/2022.emnlp-main.248}.
\newblock URL \url{https://aclanthology.org/2022.emnlp-main.248}.

\bibitem[Schuster et~al.(2019)Schuster, Pinkowski, and Fischer]{schuster_2019}
Carolin Schuster, Lisa Pinkowski, and Daniel Fischer.
\newblock Intra-individual value change in adulthood a systematic literature review of longitudinal studies assessing schwartz's value orientations.
\newblock \emph{Zeitschrift für Psychologie}, 227:\penalty0 42 --52, 03 2019.
\newblock \doi{10.1027/2151-2604/a000355}.

\bibitem[Schwartz(1992)]{schwartz1992universals}
Shalom~H. Schwartz.
\newblock Universals in the content and structure of values: Theoretical advances and empirical tests in 20 countries.
\newblock In \emph{Advances in experimental social psychology}, volume~25, pp.\  1--65. Elsevier, 1992.

\bibitem[Schwartz(2012)]{Schwartz2012-uu}
Shalom~H. Schwartz.
\newblock An overview of the schwartz theory of basic values.
\newblock \emph{Online readings in Psychology and Culture}, 2012.

\bibitem[Schwartz et~al.(2015)Schwartz, Breyer, and Danner]{hvs}
Shalom~H. Schwartz, B.~Breyer, and D.~Danner.
\newblock Human values scale (ess).
\newblock \emph{ZIS - The Collection Items and Scales for the Social Sciences}, 2015.
\newblock \doi{10.6102/zis234}.

\bibitem[Shanahan et~al.(2023)Shanahan, McDonell, and Reynolds]{shanahan2023role}
Murray Shanahan, Kyle McDonell, and Laria Reynolds.
\newblock Role-play with large language models.
\newblock \emph{arXiv preprint arXiv:2305.16367}, 2023.

\bibitem[Stevenson et~al.(2022)Stevenson, Smal, Baas, Grasman, and van~der Maas]{stevenson2022putting}
Claire Stevenson, Iris Smal, Matthijs Baas, Raoul Grasman, and Han van~der Maas.
\newblock Putting gpt-3's creativity to the (alternative uses) test.
\newblock \emph{arXiv preprint arXiv:2206.08932}, 2022.

\bibitem[Sundberg(2016)]{sundberg2016}
Ralph Sundberg.
\newblock Value stability and change in an isaf contingent.
\newblock \emph{Journal of Personality}, 84\penalty0 (1):\penalty0 91--101, 2016.
\newblock \doi{https://doi.org/10.1111/jopy.12142}.
\newblock URL \url{https://onlinelibrary.wiley.com/doi/abs/10.1111/jopy.12142}.

\bibitem[Touvron et~al.(2023)Touvron, Lavril, Izacard, Martinet, Lachaux, Lacroix, Rozi{\`e}re, Goyal, Hambro, Azhar, et~al.]{llama}
Hugo Touvron, Thibaut Lavril, Gautier Izacard, Xavier Martinet, Marie-Anne Lachaux, Timoth{\'e}e Lacroix, Baptiste Rozi{\`e}re, Naman Goyal, Eric Hambro, Faisal Azhar, et~al.
\newblock Llama: Open and efficient foundation language models.
\newblock \emph{arXiv preprint arXiv:2302.13971}, 2023.

\bibitem[Tunstall et~al.(2023)Tunstall, Beeching, Lambert, Rajani, Rasul, Belkada, Huang, von Werra, Fourrier, Habib, Sarrazin, Sanseviero, Rush, and Wolf]{zephyr}
Lewis Tunstall, Edward Beeching, Nathan Lambert, Nazneen Rajani, Kashif Rasul, Younes Belkada, Shengyi Huang, Leandro von Werra, Clémentine Fourrier, Nathan Habib, Nathan Sarrazin, Omar Sanseviero, Alexander~M. Rush, and Thomas Wolf.
\newblock Zephyr: Direct distillation of lm alignment, 2023.

\bibitem[Upstage(2023{\natexlab{a}})]{upstage_llama}
Upstage.
\newblock Hugging face model hub: Upstage-llama-65b-instruct.
\newblock \url{https://huggingface.co/upstage/upstage/llama-65b-instruct}, 2023{\natexlab{a}}.
\newblock Accessed: October 8, 2023.

\bibitem[Upstage(2023{\natexlab{b}})]{upstage_llama2}
Upstage.
\newblock Hugging face model hub: Upstage-llama-2-70b-instruct.
\newblock \url{https://huggingface.co/upstage/Llama-2-70b-instruct}, 2023{\natexlab{b}}.
\newblock Accessed: October 8, 2023.

\bibitem[Vecchione et~al.(2016)Vecchione, Schwartz, Alessandri, Döring, Castellani, and Caprara]{vecchione2016_8y}
Michele Vecchione, Shalom~H. Schwartz, Guido Alessandri, Anna Döring, Valeria Castellani, and Maria Caprara.
\newblock Stability and change of basic personal values in early adulthood: An 8-year longitudinal study.
\newblock \emph{Journal of Research in Personality}, 63, 06 2016.
\newblock \doi{10.1016/j.jrp.2016.06.002}.

\bibitem[Vecchione et~al.(2020)Vecchione, Schwartz, Davidov, Cieciuch, Alessandri, and Marsicano]{vecchione2020_2y}
Michele Vecchione, Shalom~H. Schwartz, Eldad Davidov, Jan Cieciuch, Guido Alessandri, and Gilda Marsicano.
\newblock Stability and change of basic personal values in early adolescence: A 2-year longitudinal study.
\newblock \emph{Journal of Personality}, 88\penalty0 (3):\penalty0 447--463, 2020.
\newblock \doi{https://doi.org/10.1111/jopy.12502}.
\newblock URL \url{https://onlinelibrary.wiley.com/doi/abs/10.1111/jopy.12502}.

\end{thebibliography}
\bibliographystyle{iclr2024_conference}

\appendix

\newpage
\section{Details on the psychological questionnaires}
\label{app:details_quest}

Research on values and personality traits has been conducted in various contexts. One line of research aims to explore universal personal values across different cultures \citep{schwartz1992universals,Schwartz2012-uu}. Schwartz conducted theoretical examinations of personal values and proposed ten basic elements \citep{schwartz1992universals}. Additionally, he considered that these elements have a higher-order hierarchical structure \citep{Schwartz2012-uu}. The dimensional structure of these constituent elements has been demonstrated through factor analysis of responses obtained from the PVQ questionnaire \citep{Cieciuch2012-bg}. 

In addition to investigating universal personal values, comparative studies of values across social groups and countries have been conducted \citep{hofstede1984-gz,hofstede1990-gv}. Hofstede compared employees' values from 40 countries at the IBM company using the VSM questionnaire \citep{hofstede1984-gz}. This and some follow-up studies identified six cultural dimensions in work-related national cultures \citep{hofstede1990-gv}. 

In the context of personality trait research, distinct from values research, attempts have been made to explore fundamental personality traits. Goldberg classified personality trait descriptors and identified five common traits \citep{Goldberg1990-jw}. These five constituent elements have been empirically validated through factor analysis of data obtained from questionnaires \citep{Costa1992-qu}. 

\subsection{Dimnesions of culture and personality}
\label{app:dimensions}
In the main text we briefly outlined the main values and personality traits outlined by Schwartz \citep{schwartz1992universals}, Hofstede \citep{hofstede1984-gz}, and by the Big Five personality traits model \citep{Goldberg1990-jw}. Here, we discuss each of those values and traits in more detail. 

\textbf{Schwartz's theory of basic personal values} outlines the following basic personal values \citep{Schwartz2012-uu}:
\begin{itemize}
    \item \textbf{Self-Direction} - independent thought and action-choosing, creating, exploring
    \item \textbf{Stimulation} - excitement, novelty, and challenge in life
    \item \textbf{Hedonism} - pleasure or sensuous gratification for oneself: Hedonism values derive from organismic needs and the pleasure associated with satisfying them (pleasure, enjoying life, self-indulgence)
    \item \textbf{Achievement} - personal success through demonstrating competence according to social standards
    \item \textbf{Power} - social status and prestige, control or dominance over people and resources
    \item \textbf{Security} - safety, harmony, and stability of society, of relationships, and of self
    \item \textbf{Conformity} - restraint of actions, inclinations, and impulses likely to upset or harm others and violate social expectations or norms.
    \item \textbf{Tradition} - respect, commitment, and acceptance of the customs and ideas that one's culture or religion provides.
    \item \textbf{Benevolence} - preserving and enhancing the welfare of those with whom one is in frequent personal contact (the ‘in-group’)
    \item \textbf{Universalism} - understanding, appreciation, tolerance, and protection for the welfare of all people and for nature.
\end{itemize}

\paragraph{Hofstede's theory of basic cultural dimensions} outlines the following cultural values \citep{hofstede_dimensions}:
\begin{itemize}
    \item \textbf{Power Distance} - the extent to which the less powerful members of institutions and organizations within a society expect and accept that power is distributed unequally.
    \item \textbf{Individualism} - the opposite of Collectivism. Individualism stands for a society in which the ties between individuals are loose: a person is expected to look after himself or herself and his or her immediate family only. Collectivism stands for a society in which people from birth onwards are integrated into strong, cohesive in-groups, which continue to protect them throughout their lifetime in exchange for unquestioning loyalty. 
    \item \textbf{Masculinity} - the opposite of Femininity. Masculinity stands for a society in which social gender roles are clearly distinct: men are supposed to be assertive, tough, and focused on material success; women are supposed to be more modest, tender, and concerned with the quality of life. Femininity stands for a society in which social gender roles overlap: both men and women are supposed to be modest, tender, and concerned with the quality of life.
    \item \textbf{Uncertainty Avoidance} - the extent to which the members of institutions and organizations within a society feel threatened by uncertain, unknown, ambiguous, or unstructured situations.
    \item \textbf{Long Term Orientation} - the opposite of Short Term Orientation. Long Term Orientation stands for a society which fosters virtues oriented towards future rewards, in particular adaptation, perseverance and thrift. Short Term orientation stands for a society which fosters virtues related to the past and present, in particular respect for tradition, preservation of “face”, and fulfilling social obligations.
    \item \textbf{Indulgence} - a society which allows relatively free gratification of some desires and feelings, especially those that have to do with leisure, merrymaking with friends, spending, consumption and sex. Its opposite pole, Restraint, stands for a society which controls such gratification, and where people feel less able to enjoy their lives.
    \end{itemize}

\textbf{The Big Five personality traits model} outlines the following five personality traits \citep{goldberg1993structure}:
\begin{itemize}
    \item \textbf{Extraversion} - contrasts such traits as talkativeness, assertiveness, and activity level with traits such as silence, passivity, and reserve.
    \item \textbf{Agreeableness} - contrasts traits such as kindness, trust, and warmth with such traits as hostility, selfishness, and distrust.
    \item \textbf{Conscientiousness} - contrasts such traits as organization, thoroughness, and reliability with traits such as carelessness, negligence, and unreliability.
    \item \textbf{Neuroticism} - includes such traits as nervousness, moodiness, and temperamentality.
    \item \textbf{Openness to Experience} - contrasts such traits as imagination, curiosity, and creativity with traits such as shallowness and imperceptiveness.
    
\end{itemize}

\subsection{Changes to the instructions in the questionnaires}
\label{app:instruction_changes}

To make questionnaires more suitable for LLM the main instructions were adapted.
For PVQ, the last sentence of the instruction "Put an X in the box to the right that shows how much the person in the description is like you." was changed to  "Select an option that shows how much the person in the description is like you.". 
For VSM, the brackets with "please circle one answer in each line across" were removed.
For IPIP, multiple versions exist with longer and no instructions, we used the following instruction "Mark how much you agree with each statement."

\subsection{Computation of Hofstede's VSM scores}
\label{app:hof_computation}

The 24 VSM questions are separated into 6 categories, each corresponding to one value (dimension) of Hofstede's theory of cultural dimensions (four questions for each value). 
The score for each value is computed according to the following equation:
\begin{equation}
s = A * (q_1 - q_2) + B * (q_3 - q_4) + C
\label{eq:vsm}
\end{equation}
Where $s$ is the score, $A$ and $B$ are value-dependent constants, $q_1, q_2, q_3, q_4$ are the responses to the questions (in the interval from one to five), and $C$ is a constant that can be used to adjust the final score interval (we set it to 0). Constants $A$ and $B$ for each value are shown in table \ref{tab:vsm_const}.

\begin{table}
    \caption{\textbf{VSM constanct} Constants used in the equation \ref{eq:vsm} for the calculation of the scores on the VSM questionniare.}
    \centering
    \begin{tabular}{lcc}
        \bf Value & \bf A & \bf B \\
        \hline
        Power distance & 35 & 35 \\
        Individualism  & 35 & 35  \\
        Masculinity    & 35 & 35  \\
        Uncertainty avoidance & 40 & 25 \\
        Long-term orientation &  40 & 25 \\
        Indulgence & 35 & 40 \\
    \end{tabular}
    \label{tab:vsm_const}
\end{table}

\section{Large Language Models compared in systematic experiments}
\label{sec:models}

From the OpenAI API \citep{openai_api} we use the following GPT models:
"GPT-3.5-0301" (gpt-3.5-turbo-0301) \citep{instructgpt},
"GPT-3.5-0314" (gpt-3.5-urbo-0314),
"GPT-4-0314" (gpt-4-0314") \citep{gpt4},
"Ada" (text-ada-001) \citep{gpt3},
"Babbage" (text-babbage-001),
"Curie" (text-curie-001),
"Davinci-003 (text-davinci-003).
OpenAssistant is a set of models trained by RLHF. In this paper, we use a 30B parameter model "openassistant\_rlhf2\_llama30b" which was fine-tuned by RLHF from LLaMa-30B \cite{llama}.
Zephyr \citep{zephyr}, StableVicuna and StableLM are models from StabilityAI available through the HuggingFace transformers library \cite{openai_api} as "HuggingFaceH4/zephyr-7b-alpha", "CarperAI/stable-vicuna-13b-delta" and "stabilityai/stablelm-tuned-alpha-7b".
StableVicuna is 13B parameter model created by RLHF fine-tuning of Vicuna-13b \cite{vicuna2023}, which was created by instruction fine-tuning of the LLaMa-13B model \cite{llama}.
StableLM is a set of models, from which we use the 7B parameter "stabilityai/stablelm-tuned-alpha-7" model. It was created by fine-tuning the "StableLM-Base-Alpha" model on chat and instruction-following datasets.
We compare three version of the LLaMa \cite{llama} models. LLaMa-65B released by meta, and two upstage fine-tuned versions: "upstage/llama-65b-instruct" \cite{upstage_llama} and "upstage/Llama-2-70b-instruct" \cite{upstage_llama2} which are available on the hugginface model hub \cite{huggingfacemodelhub}.
Finally, we compare with two version of RedPajama incite model: rp-incite-7b-instruct \citep{rpinstrct} and rp-incite-7b-chat \cite{rpchat}, which are also available on the huggingface model hub.

\section{Analyzing the unexpected perspective shift with respect to human studies}
\label{app:human_studies}

In this section, we study how change in value expression in ChatGPT (gpt-3.5-turbo-0301) compares to changes in humans.
We interpret the results from experiments in section \ref{sec:exp_unexpected} in the context of human studies.
We extract human data concerning value change as a consequence of development (longitudinal studies) or as a consequence of priming (artificially increasing the expression of some targeted value).
Regarding longitudinal studies we extract data from the following three studies.
In \cite{vecchione2020_2y} ten to twelve year old children were followed for two years, and in \cite{vecchione2016_8y} twenty year old's were followed for a period of eight years.
In \cite{sundberg2016} Swedish soldiers were evaluated before and after a 6-month tour in Afghanistan.
Regarding priming, we extract data from the following two studies.
In \cite{doring2015} adolescents (ages 13 to 15) watched a 33 minute long movie consisting of "adventure and nature scenes that convey a positive attitude to life and search for stimulation, adventures, and challenges".
The movie aimed at increasing Self-direction and Universalism, and at decreasing Conformity and Security.
In \cite{arieli2014} the authors aimed to artificially increase the expression of benevolence in three different ways.
The biggest change was induced when participants (aged 18 to 21) read a scientiﬁc testimony about how people are signiﬁcantly more other-focused, cooperative, compassionate, and helpful than most people realize, and how beneﬁting others beneﬁts the self in the long run.

These studies present value change in humans due to circumstances which are much more extreme than those we study with LLMs.
We will show that, despite this, unexpected value changes in LLMs are even bigger than those observed in humans in those scenarios. 
The longitudinal studies we consider \citep{vecchione2016_8y,vecchione2020_2y}, apart from \cite{sundberg2016}, study humans in early-adolescence and young adulthood, which are particularly challenging and dynamics life periods \citep{vecchione2016_8y}.
Furthermore, these studies concern extreme influences: 2 or 8 years of development, and 6 months of exposure to war.
The priming studies attempt to artificially modify value expression (e.g. by reading or watching a movie). An open question in those studies is to what extent is this change long-lasting.
While at first glance priming might seem similar to the \textit{unexpected perspective shift effect}, the crucial difference is that the \textit{unexpected perspective shift effect} results from stimuli which were not intentionally created to modify values in any specific direction.
For example, playing chess is not targeting any value, while a text about how people are often underestimate their compassion is directly aiming to increased the expression of benevolence.
Therefore, priming is more similar to the experiments on controllability, where we explicitly set target values (section \ref{sec:exp_contr_sys}).
The motivation behind this section is to show that changes in human value expression (even those caused by much more extreme scenarios such as development or priming) are smaller than changes in LLMs' value expression (caused by mere context changes).

In psychology, value stability and change is commonly studied through \textit{mean-level change} and through participant \textit{rank-order stability} \citep{schuster_2019}.
Both of those methods study how values change in human populations.
Studies on within-person (\textit{ipsative}) stability are comparatively rare, they often involve comparing the ranks of values in individuals' hierarchies \citep{schuster_2019}. 
In the remainder of this section we discuss those methods and compare ChatGPT with data from human studies.

\subsection{Mean-level change}
Apart from analysing the mean-level change with ANOVA and t-tests (as we did in section \ref{sec:exp_unexpected}), it is common evaluate to the effect size of those changes using the Cohen's d coefficient \citep{arieli2014,doring2015}.
Cohen's d corresponds to the difference in the means of two distributions divided by the pooled standard deviation.
To put our results in context of human studies, we draw an analogy between participants in human studies and permutations in the order of answers in LLM studies.
We believe that this analogy, although not perfect, is sound because in human studies multiple participants are used to provide robustness to the results, and we use permutations for the same purpose.
We extract data from longitudinal studies with 10 to 12 year-olds \citep{vecchione2020_2y} and 20-year-olds \citep{vecchione2016_8y}, and from priming studies with a movie \cite{doring2015} and with reading \citep{arieli2014}. 

Table \ref{tab:mean_value_change} shows effect sizes (Cohen's d) of value changes reported in human studies compared those we observe in LLMs.
For humans, the biggest change is reported for Conformity in 8 years (20 to 28 years old) and for Benevolence when primed for its increase by reading ($d=0.53$).
This is closely followed by Conformity when primed for its decrease with a movie ($d=-0.52$).
The table shows that ChatGPTs values change significantly more as a consequence of context change (max $d=-5.86$). 
We can see that a change bigger than that in humans ($d=|0.53|$) is observed in at least one value in all the ChatGPT context changes.
Overall, this data shows that context change in ChatGPT results in big mean-level changes.
These changes are much bigger than those in humans despite them being a consequence of much more drastic changes (e.g. 8 years of development or priming).

\begin{table}[ht]
\centering
\scriptsize
\begin{tabular}{l|rrrrrrrrrr}
                  & \rotatebox{90}{\bf Conformity} & \rotatebox{90}{\bf Tradition} & \rotatebox{90}{\bf Benevolence} & \rotatebox{90}{\bf Universalism} & \rotatebox{90}{\bf Self-Direction} & \rotatebox{90}{\bf Stimulation} & \rotatebox{90}{\bf Hedonism} & \rotatebox{90}{\bf Achievement} & \rotatebox{90}{\bf Power} & \rotatebox{90}{\bf Security} \\

\toprule
\midrule
\textit{\small Human studies} & & & & & & & & & & \\
10 to 12 year-olds (+2 years) & 0.05 & -0.09 & 0.02 & 0.04 & 0.30 & 0.22 & 0.29 & 0.17 & 0.16 & 0.01 \\
20 year-olds (+8 years) & \textbf{0.53} & 0.04 & 0.23 & 0.40 & 0.17 & -0.20 & -0.06 & -0.23 & 0.20 & 0.22 \\
priming for benevolence (reading) &  &  & \textbf{0.53} & & & & & & & \\
priming (movie) &  -0.52 & 0.23 & 0.12 & 0.40 & 0.10 & 0.13 & 0.10 & 0.02 & -0.16 & -0.25 \\
\midrule
\textit{\small Simulated conv.} & & & & & & & & & & \\
chess - grammar  & -2.98 & -0.70 & 0.83 & 1.09 & 0.32 & 2.03 & 0.96 & -0.50 & -0.47 & -2.44 \\
chess - history  & -2.93 & -0.76 & 1.89 & 1.56 & 2.00 & 1.58 & -0.43 & -2.22 & -1.50 & -0.57 \\
chess - joke     & -2.07 & -1.26 & 2.08 & 3.59 & 2.41 & 0.83 & -0.50 & -3.04 & -2.56 & -0.85 \\
chess - poem     & -1.84 & -1.78 & -0.96 & -0.84 & -0.68 & 2.15 & 1.39 & 0.34 & -0.30 & -0.81 \\
grammar - history& -0.07 & -0.06 & 0.97 & 1.52 & 1.52 & -0.46 & -1.63 & -1.57 & -0.90 & 1.51 \\
grammar - joke   & 0.53 & -0.39 & 1.34 & 1.95 & 1.95 & -1.14 & -1.63 & -2.42 & -1.77 & 1.45 \\
grammar - poem   & 1.76 & -0.76 & -1.55 & -1.84 & -0.91 & 0.19 & 0.47 & 0.81 & 0.27 & 1.58 \\
history - joke   & 0.57 & -0.32 & 0.51 & 1.28 & 1.28 & -0.70 & -0.09 & -0.99 & -0.81 & -0.18 \\
history - poem   & 1.74 & -0.68 & -2.53 & -2.13 & -2.13 & 0.63 & 2.17 & 2.56 & 1.35 & -0.12  \\
joke - poem      & 0.84 & -0.41 & -2.60 & \textbf{-4.33} & \textbf{-4.33} & 1.28 & 2.15 & 3.34 & 2.44 & 0.07 \\
\midrule
\textit{\small Textual formats} & & & & & & & & & & \\
chat - code\_cpp & -1.05 & -1.05 & 4.00 & 3.58 & 5.31 & -1.26 & -0.69 & -2.32 & \textbf{-5.86} & 1.81 \\
chat - code\_py & 1.21 & -0.57 & 3.09 & 2.50 & 3.93 & -1.56 & -1.17 & -1.83 & -3.32 & 2.00 \\
chat - conf\_toml & -0.25 & -0.49 & 0.87 & 0.45 & 1.73 & -0.60 & -1.18 & -0.35 & 0.27 & 0.38 \\
chat - latex & 0.67 & -0.20 & 1.36 & 0.97 & 1.81 & -1.33 & -1.31 & -0.28 & -1.40 & 1.66 \\
code\_cpp - code\_py & 1.97 & 0.49 & -0.93 & -0.81 & -1.37 & -0.65 & -0.52 & 0.18 & 0.20 & 0.58 \\
code\_cpp - conf\_toml & 0.73 & 0.56 & -3.65 & -2.94 & -3.34 & 0.48 & -0.66 & 1.73 & 5.02 & -1.16 \\
code\_cpp - latex & 1.35 & 0.72 & -1.93 & -2.15 & -2.22 & -0.53 & -0.78 & 1.54 & 1.48 & 0.56 \\
code\_py - conf\_toml & -1.27 & 0.08 & -2.55 & -1.97 & -2.04 & 0.88 & -0.24 & 1.37 & 3.25 & -1.46 \\
code\_py - latex & -0.18 & 0.30 & -1.23 & -1.31 & -1.23 & -0.07 & -0.34 & 1.26 & 1.10 & 0.13 \\
conf\_toml - latex & 0.79 & 0.23 & 0.70 & 0.53 & 0.38 & -0.78 & -0.08 & 0.03 & -1.50 & 1.28 \\
\midrule
\textit{\small Textual formats} & & & & & & & & & & \\
classical - gospel & -0.12 & -1.14 & -0.45 & 0.40 & 1.04 & 1.67 & -0.63 & 0.37 & 0.14 & -1.29 \\
classical - heavy metal & 1.56 & 1.61 & 0.45 & -0.18 & 0.50 & -0.06 & -1.96 & -0.84 & 0.14 & -0.47 \\
classical - hip-hop & 0.96 & 0.68 & 0.37 & 0.35 & 1.00 & 0.08 & -1.46 & -0.32 & -0.20 & -1.11 \\
classical - jazz & 1.06 & 1.24 & 0.10 & 0.52 & -0.15 & 0.04 & -2.32 & 0.12 & 0.45 & 0.42 \\
classical - reggae & 0.84 & 0.16 & -0.05 & -0.66 & 0.52 & 1.18 & -1.88 & 0.91 & -0.02 & -0.58 \\
gospel - heavy metal & 1.58 & \textbf{2.89} & 0.94 & -0.57 & -0.61 & -1.97 & -1.14 & -1.15 & 0.02 & 0.78 \\
gospel - hip-hop & 1.02 & 1.73 & 0.87 & -0.06 & -0.09 & -2.20 & -0.62 & -0.70 & -0.31 & 0.40 \\
gospel - jazz & 1.11 & 2.65 & 0.58 & 0.13 & -1.18 & -2.06 & -1.58 & -0.26 & 0.30 & 1.70 \\
gospel - reggae & 0.91 & 1.43 & 0.43 & -1.05 & -0.53 & -0.47 & -1.12 & 0.47 & -0.16 & 0.84 \\
heavy metal - hip-hop & -0.70 & -0.73 & -0.09 & 0.54 & 0.55 & 0.17 & 0.96 & 0.61 & -0.31 & -0.52 \\
heavy metal - jazz & -0.45 & -0.56 & -0.37 & 0.69 & -0.65 & 0.12 & -0.82 & 0.98 & 0.27 & 0.87 \\
heavy metal - reggae & -0.83 & -1.60 & -0.54 & -0.47 & 0.06 & 1.38 & -0.10 & 1.80 & -0.16 & -0.04 \\
hip-hop - jazz & 0.20 & 0.33 & -0.29 & 0.19 & -1.15 & -0.05 & -1.56 & 0.46 & 0.60 & 1.58 \\
hip-hop - reggae & -0.13 & -0.59 & -0.46 & -1.02 & -0.46 & 1.46 & -0.87 & 1.34 & 0.19 & 0.56 \\
jazz - reggae & -0.32 & -1.20 & -0.16 & -1.16 & 0.67 & 1.39 & 0.64 & 0.80 & -0.48 & -1.03 \\
\end{tabular}
\caption{\textbf{Mean-level changes} Effect size (Cohen's d) of value expression change for human studies and three LLM experiments (the biggest change in each of the four groups is bolded). The changes in ChatGPT's value expression are much bigger (up to $d=5.86$) than those in humans (up to $d=0.53$). This is despite changes in human studies resulting from more extreme circumstances (priming or early and late adolescent development) compared to seemingly irrelevant context changes in LLMs.}
\label{tab:mean_value_change}
\end{table}

\subsection{Rank-order stability}
Value change can be evaluated by comparing participants' ranks, this method enables us to detect changes in populations that do not results in mean-level change.
Let us consider an example of a population where the importance of benevolence increased in one half of people and decreased in the other half.
Even though this change is drastic, we would not be able to detect it by measuring mean-level change.
We can order the participants based on their expression of some value (e.g. benevolence) and then compute the correlation between participants' ranks before and after the change occurred.
In other words, this estimates the effect of time change on the order of participants.

The motivation of this section is twofold. First, to put changes observed in ChatGPT in context of human changes, and, second, to show how rank-order stability methodology can be used in AI to estimate values' stability with respect to different kinds of context changes.
In human studies, the correlation between the order of participants in two time points (due to time change) is computed. 
We extract data from longitudinal studies with 10 to 12 year-olds \citep{vecchione2020_2y}, with 20-year-olds \citep{vecchione2016_8y}, and with soldiers in Afghanistan \citep{sundberg2016}.
Following our experiments in section \ref{sec:exp_unexpected}, we can consider two types of permutation change.
For example, following the \textit{Simulated conversations} experiments we can consider conversation topic change and permutation change (change in the order of suggested answers).
Therefore, we can study the effect of topic change on the order of permutations, and the effect of permutation change on the order of topics.
To estimate the stability of some value in terms of the order of topics due to permutation change we do the following.
First, we compute the correlation between the order of topics for every possible pair of permutations. Second, we average the correlations of those pairs.
To estimate the permutation change due to topic change the process is repeated by exchanging permutations and perspectives.
We conduct analyses for both stability types following the three experiments from section \ref{sec:exp_unexpected}: \textit{Simulated conversations}, \textit{Textual formats}, \textit{Wikipedia paragraphs}

Table \ref{tab:rank_order_stability} shows rank-order stability for each value and their mean.
For humans, the order of participants changed the most in early adolescents (ten to twelve year-olds) in the period of two years ($r=0.57$). 
In ChatGPT, bigger-than-human change ($r<0.57$) is observed in both stability types in all three experiments, with the biggest changes observed on Simulated conversations ($r=0.42$ and $r=0.36$).
Regarding per-value changes, we can see that, in humans, the biggest change is likewise observed the early adolescent groups in the expression of Achievement ($r=0.39$).
In ChatGPT, bigger-than-human per-value changes ($r<0.39$) are observed in many values in all three experiments, with the biggest change observed in Benevolence for Textual formats perspective order change ($r=0.04$).
In general, these results show that the order of human participants (due to time), is more stable than the order of both topics/formats/paragraphs and the order of permutations in ChatGPT.

Rank-order methodology analysis can be useful to study the stability of values with respect to different types of context change in LLMs.
We can aggregate per-value stability with respect to different kinds of changes to obtain an overall per-value stability estimate.
Table \ref{tab:rank_order_stability} also shows the mean of perspective order change and permutation order change. 
We can see that Conformity was the least stable value in Simulated conversations ($r=0.26$) and Wikipedia paragraphs ($r=0.40$), while Benevolence was the least stable in Textual formats experiments ($r=0.27$).
While here we only consider two types of context change (topics/formats/paragraphs and permutations), one can imagine extending this set to other types of changes such as different languages or adding whitespace.
In this scenario, the models sensitivity to, for example language, can be estimated by aggregating all the other stability estimates due to language change.
We leave such analysis for future work as here we merely wanted to demonstrate how rank-order stability can be a valuable tool to analyse LLMs.

\begin{table}[ht]
\centering
\scriptsize
\begin{tabular}{l|cccccccccc|c}
 & \rotatebox{90}{\bf Conformity} & \rotatebox{90}{\bf Tradition} & \rotatebox{90}{\bf Benevolence} & \rotatebox{90}{\bf Universalism} & \rotatebox{90}{\bf Self-Direction} & \rotatebox{90}{\bf Stimulation} & \rotatebox{90}{\bf Hedonism} & \rotatebox{90}{\bf Achievement} & \rotatebox{90}{\bf Power} & \rotatebox{90}{\bf Security} & \rotatebox{90}{\bf Mean} \\

\toprule
\midrule
\textit{\small Human studies} & & & & & & & & & & \\
\mlcell{Participant order change due to:} & & & & & & & & & & \\
\mlcell{- 2 years development (10-12 year olds)} & 0.40 & 0.59 & 0.51 & 0.49 & 0.63 & 0.73 & 0.70 & \textbf{0.39} & 0.48 & 0.77 & \textbf{0.57} \\
\mlcell{- 8 years development (20 year olds)} & 0.66 & 0.68 & 0.75 & 0.65 & 0.77 & 0.82 & 0.57 & 0.59 & 0.57 & 0.51 & 0.66 \\
\mlcell{- war (soldiers)} & 0.92 & 0.57 & 0.92 & 0.82 & 0.88 & 0.74 & 0.91 & 0.79 & 0.84 & 0.83 & 0.82 \\
\midrule
\textit{\small Simulated conv.} & & & & & & & & & & \\
\mlcell{Topic order change due to perm. change} & 0.32 & 0.26 & 0.70 & 0.73 & 0.72 & 0.46 & 0.26 & 0.18 & \bf 0.21 & 0.41 & 0.42 \\
\mlcell{Perm. order change due to topic change} & \bf 0.21 & 0.32 & 0.28 & 0.50 & 0.38 & 0.33 & 0.38 & 0.42 & 0.43 & 0.38 & \textbf{0.36} \\
Mean & 0.26 & 0.29 & 0.49 & 0.62 & 0.55 & 0.40 & 0.32 & 0.30 & 0.32 & 0.40 & 0.39 \\
\midrule
\textit{\small Textual formats} & & & & & & & & & & \\
\mlcell{Format order change due to perm. change} & 0.70 & 0.66 & \bf 0.04 & 0.25 & 0.16 & 0.72 & 0.69 & 0.79 & 0.83 & 0.60 & 0.54 \\
\mlcell{Perm. order change due to format change} & 0.28 & 0.22 & 0.50 & 0.61 & 0.55 & 0.41 & 0.21 & 0.22 & 0.23 & 0.49 & \bf 0.37 \\
Mean & 0.49 & 0.44 & 0.27 & 0.43 & 0.36 & 0.56 & 0.45 & 0.50 & 0.53 & 0.54 & 0.46 \\
\midrule
\textit{\small Wikipedia paragraphs} & & & & & & & & & & \\
\mlcell{Article order change due to perm. change} & 0.39 & 0.61 & 0.36 & 0.46 & \bf 0.24 & 0.51 & 0.62 & 0.54 & 0.39 & 0.64 & \bf 0.48 \\
\mlcell{Perm. order change due to article change} & 0.40 & 0.34 & 0.74 & 0.61 & 0.72 & 0.45 & 0.50 & 0.55 & 0.59 & 0.52 & 0.54 \\
Mean & 0.40 & 0.48 & 0.55 & 0.54 & 0.48 & 0.48 & 0.56 & 0.55 & 0.49 & 0.58 & 0.51 \\
\end{tabular}
\caption{\textbf{Rank-order stability} Pearson correlations representing the ChatGPT stability of permutation order due to perspective change and stability of perspectives order due to permutation change. In all three ChatGPT experiments, those changes are smaller than those in people due to years of development or war. The biggest change in every experiment is in bold.}
\label{tab:rank_order_stability}
\end{table}

\subsection{Intraindividual (Ipsative) changes}
Intrapersonal value change can be estimated by computing the correlation between the individual value hierarchies (priorities) at two points in time.
The per-participant correlation coefficients are then averaged to compute the final estimate.
To extend this method to LLMs we draw the analogy between participants and permutations in the order of answers. 
We compute the correlation of value priorities between two perspectives (e.g. chat and code\_cpp) for each permutation separately, and then compute the average over permutations. 
This enables us to robustly estimate the effect of changing a perspective.
We extract human data from longitudinal studies with 10 to 12 year-olds \citep{vecchione2020_2y}, 20-year-olds \citep{vecchione2016_8y}, and from soldiers sent to Afghanistan \citep{sundberg2016}.

Table \ref{tab:ipsative_change} shows Pearson correlation coefficients between value priorities for human experiments (at two different time points) and for ChatGPT (in two different perspectives).
For humans, the biggest change, that of $r=0.59$ was observed as a consequence of developments from age 20 to 24 and from 20 to 28.
For ChatGPT, the biggest change ($r=0.05$) was observed in textual formats experiment between the "chat" and "code\_cpp" perspectives, and a slightly smaller change ($r=0.08$) between "code\_cpp" and "conf\_toml".
Gray rows on table \ref{tab:ipsative_change} represent perspective changes that resulted in changes bigger than the that biggest change observed in humans ($r<0.59$).
We can see many such changes in \textit{Simulated conversations} and \textit{Text formats} experiments.
However, in the \textit{Wikipedia paragraphs} experiment changes are significantly smaller, with the biggest one observed between gospel and heavy metal paragraphs ($r=0.77$).
This change is comparable to six months of soldiers in war and to three months of development from age 10.
Overall, we this table shows many examples of seemingly insignificant context changes that result in big changes in the value expression of ChatGPT.
Those changes are even bigger than those in humans, which are caused by much more extreme circumstances (years of development during adolescence or early adulthood and war) than trivial context changes in LLMs (conversation topics, textual formats).

\begin{table}[ht]
\centering
\small
\begin{tabular}{l|ccccc}
\textbf{Context change} & \textbf{Mean} & \textbf{Median} & \textbf{STD} & \textbf{Min} & \textbf{Max} \\
\toprule
\midrule
\textit{Human studies} \\
20 year-olds (+4 years) \citep{vecchione2016_8y} & \textbf{0.59} & 0.68 & 0.25 & -0.36 & 0.88 \\
24 year-olds (+4 years) \citep{vecchione2016_8y} & 0.65 & 0.73 & 0.25 & -0.42 & 0.88 \\
20 year-olds (+8 years) \citep{vecchione2016_8y} & \textbf{0.59} & 0.66 & 0.25 & -0.30 & 0.89 \\
10 to 12 year-olds (+3 months) \citep{vecchione2016_8y} & 0.83 & 0.88 & 0.16 & 0.07 & 1.00 \\
10 to 12 year-olds (+2 years) \citep{vecchione2016_8y} & 0.66 & 0.72 & 0.26 & -0.27 & 1.00 \\
soldiers (+6 months of war) \citep{sundberg2016} & 0.75 & - & 0.22 & -0.47 & 0.98 \\ 
\midrule
\textit{Simulated conversations} \\
chess - grammar & 0.78 & 0.79 & 0.12 & 0.38 & 0.95 \\
chess - history & 0.70 & 0.74 & 0.19 & 0.00 & 0.91 \\
\grayrow chess - joke & 0.48* & 0.56 & 0.27 & -0.03 & 0.83 \\
chess - poem & 0.87 & 0.89 & 0.07 & 0.62 & 0.98 \\
grammar - history & 0.70 & 0.73 & 0.18 & 0.00 & 0.91 \\
\grayrow grammar - joke & \textbf{0.40*} & 0.53 & 0.32 & -0.26 & 0.87 \\
grammar - poem & 0.90 & 0.91 & 0.06 & 0.71 & 0.98 \\
\grayrow history - joke & 0.52 & 0.57 & 0.30 & -0.25 & 0.99 \\
history - poem & 0.74 & 0.78 & 0.20 & 0.00 & 0.95 \\
\grayrow joke - poem & 0.42* & 0.50 & 0.32 & -0.19 & 0.87 \\
\midrule
\textit{Textual formats} \\
\grayrow chat - code\_cpp & \textbf{0.05*} & 0.00 & 0.23 & -0.61 & 0.50 \\
\grayrow chat - code\_py & 0.31* & 0.28 & 0.28 & -0.24 & 0.87 \\
chat - conf\_toml & 0.86 & 0.87 & 0.06 & 0.70 & 0.96 \\
chat - latex & 0.68 & 0.72 & 0.19 & 0.06 & 0.98 \\
\grayrow code\_cpp - code\_py & 0.30* & 0.34 & 0.35 & -0.56 & 0.93 \\
\grayrow code\_cpp - conf\_toml & 0.08* & 0.00 & 0.24 & -0.65 & 0.64 \\
\grayrow code\_cpp - latex & 0.20* & 0.18 & 0.32 & -0.61 & 0.90 \\
\grayrow code\_py - conf\_toml & 0.33* & 0.28 & 0.31 & -0.45 & 0.96 \\
\grayrow code\_py - latex & 0.55 & 0.62 & 0.27 & -0.26 & 0.91 \\
conf\_toml - latex & 0.68 & 0.76 & 0.23 & 0.16 & 0.97 \\
\midrule
\textit{Wikipedia paragraphs} \\
classical - gospel & 0.84 & 0.86 & 0.09 & 0.56 & 0.96 \\
classical - heavy metal & 0.80 & 0.82 & 0.10 & 0.58 & 0.95 \\
classical - hip-hop & 0.86 & 0.88 & 0.07 & 0.71 & 0.99 \\
classical - jazz & 0.82 & 0.84 & 0.11 & 0.57 & 0.96 \\
classical - reggae & 0.84 & 0.85 & 0.09 & 0.56 & 0.97 \\
gospel - heavy metal & \textbf{0.77} & 0.78 & 0.09 & 0.49 & 0.95 \\
gospel - hip-hop & 0.83 & 0.84 & 0.08 & 0.61 & 0.95 \\
gospel - jazz & 0.79 & 0.80 & 0.10 & 0.35 & 0.93 \\
gospel - reggae & 0.88 & 0.89 & 0.06 & 0.73 & 0.97 \\
heavy metal - hip-hop & 0.91 & 0.93 & 0.05 & 0.77 & 0.99 \\
heavy metal - jazz & 0.92 & 0.92 & 0.05 & 0.70 & 0.99 \\
heavy metal - reggae & 0.86 & 0.88 & 0.09 & 0.52 & 0.96 \\
hip-hop - jazz & 0.90 & 0.91 & 0.05 & 0.76 & 0.98 \\
hip-hop - reggae & 0.88 & 0.90 & 0.08 & 0.50 & 0.99 \\
jazz - reggae & 0.89 & 0.91 & 0.06 & 0.72 & 0.98 \\
\end{tabular}
\caption{\textbf{Intrapersonal (Ipsative) value change} Pearson correlation coefficients between value priorities for humans (at two different time points) and for LLMs (in two different perspectives).
The biggest change in each group of experiments is bolded. 
Gray rows represent LLM changes bigger than the biggest one in humans, i.e. $r<0.59$. 
In those rows statistical difference with human change of 8 years is denoted by * (p=0.05, Bonferroni corrected).
Many bigger-than-human changes are observed in \textit{Simulated conversations} experiments (up to $r=0.4$) and \textit{Text formats} experiments (up to $r=0.05$).
Seemingly insignificant context changes in LLMs result in value changes bigger than those in humans, which are caused by much more extreme circumstances (early and late adolescent development and war).
}
\label{tab:ipsative_change}
\end{table}

\section{Systematic comparison of models on different types of value stability (Mean-level, rank-order, ipsative)}
\label{app:unexp_sys}

In section \ref{sec:exp_unexpected}, we presented evidence for the existence of the unexpected perspective shift effect in ChatGPT (gpt-3.5-turbo-0301). 
Then in appendix \ref{app:human_studies}, we put those results in context of human studies by evaluating three different types of value stability: mean-level change (Cohen's d), rank-order stability, and within-person (ipstative) change.
There, we demonstrated that ChatGPT often exhibits bigger value change than that in humans, which is even caused by more drastic scenarios.
In this section, we systematically compare the three types of stability over different textual formats for six different models.

We define aggregated metrics for mean-level change, rank-order stability and Ipsative value change.
For mean-level change, we compute the mean absolute Cohen's d between all pairs of perspectives (textual formats) and values (mean over $10 \times \binom{5}{2}=100$ Cohen's d values, for 10 values and 5 perspectives)
For rank-order stability, we compute the average the correlation coefficients over values and two kinds of change (format order and permutation order). This metric was already used in appnedix \ref{app:human_studies} to analyze ChatGPT rank-order stability.
For ipsative value change, we compute the average the coefficents over all possible pairs of perspectives (over $\binom{5}{2}=10$ coefficients), i.e. we average the \textit{Mean} collumn in table \ref{tab:ipsative_change}.

We compare models on the changes induced by the \textit{Textual formats} perspective change. 
We compare model which appeared more controllable in experiments in section \ref{sec:exp_contr_sys}: GPT-3.5-0613, GPT-3.5-0301, Upst-LLaMa-2-70B-instruct, Upst-LLaMa-66B-instruct,  OpenAssistant, and Zephyr-7b-beta.
We outline these because highly uncontrollable models can exhibit low change in value expression, but only because these models completely fail at the task (e.g. they collapse to the same response for every question).
In other words, to study value stability one must first establish that the model is at least somewhat capable at overall the task.

Table \ref{tab:formats_sys} compares six models along three types of value change: mean-level, rank-order, and Ipsative.
For mean-level change, we can see that the OpenAssistant model is the most stable and GPT-3.5-0301 the least stable.
For rank-order stabilty, we can see that the GPT-3.5-0613 model is the most stable and Zephyr-7b-beta the least stable.
For Ipsative stability, we can see that Upstage-LLaMa-66B-instruct model is the most stable and Zephyr-7b-beta the least stable.
Overall, we can see that, depending on the type of stability, different models appear more or less stable.
These results open many future research directions into models' value stability. 
Examples of open questions include: "Which types of stability are more important for which types of usecases?",
"Are different models specialized in different types of stabilty needed, or can one model by highly stable in all types?", "How can we explore these types of stability in more detail?", and many more.

\begin{table}[htb]
    \caption{\textbf{Systematic comparison of models' unexpected perspective shift} Three different types of stability are computed: mean-level, rank-order, and Ipsative. The most robust model for each sensitivity type is bolded. We can see that models differ in kinds of stability they exhibit.
    }
    \begin{center}
    \begin{tabular}{l|ccc}

        & \bf Mean-level ($\downarrow$) & \bf Rank-order r ($\uparrow$)  & \bf Ipsative r ($\uparrow$) \\ 
        \toprule
        \midrule
        GPT-3.5-0613 & 0.95 & \textbf{0.5} & 0.24 \\ 
        GPT-3.5-0301 & 1.4 & 0.46 & 0.40 \\ 
        Upst-LLaMa-2-70B-instruct & 0.69 & 0.35 & 0.43 \\ 
        Upst-LLaMa-66B-instruct  & 0.97 & 0.32 & \bf 0.49 \\ 
        OpenAssistant  & \bf 0.38 & 0.3 & 0.37 \\ 
        Zephyr-7b-beta & 0.43 & 0.11 &  0.13 \\ 
    \end{tabular}
    \end{center}
    \label{tab:formats_sys}
\end{table}

\section{Additional experiments}
\label{app:additional_experiments}

\subsection{Can an LLM's perspective be controlled (implicitly or explicitly) to exhibit a variety of personal values?}
\label{sec:exp_impl_expl}

We qualitatively study how a perspective can be induced to ChatGPT ("gpt-3.5-turbo-0301") by implying the target values \textit{implicitly}.
We induce the perspective of different fictional characters from The Lord of the Rings, including Sauron, Gandalf, Aragorn, Pippin, and Frodo (see appendix \ref{app:lotr} for a description of these characters).
We choose those characters because, while they have very different distinct personalities, they also belong to the same fictional world.
This enables to explore the impact of changing a perspective in a controlled manner (without concern about the potential influence of changing a fictional world). 

We use the \textit{System message} and \textit{2nd person} settings as defined in section \ref{sec:methods} (e.g. we give "You are Pippin from The Lord of the Rings" as the System message).
In this experiment, we do not permute the order of answer options as discussed in section \ref{sec:methods}, rather we present the questionnaire once per perspective, with the original order of options (from A "Not like me at all" to F "Very much like me").

\begin{figure}
\centering
\includegraphics[width=0.8\linewidth]{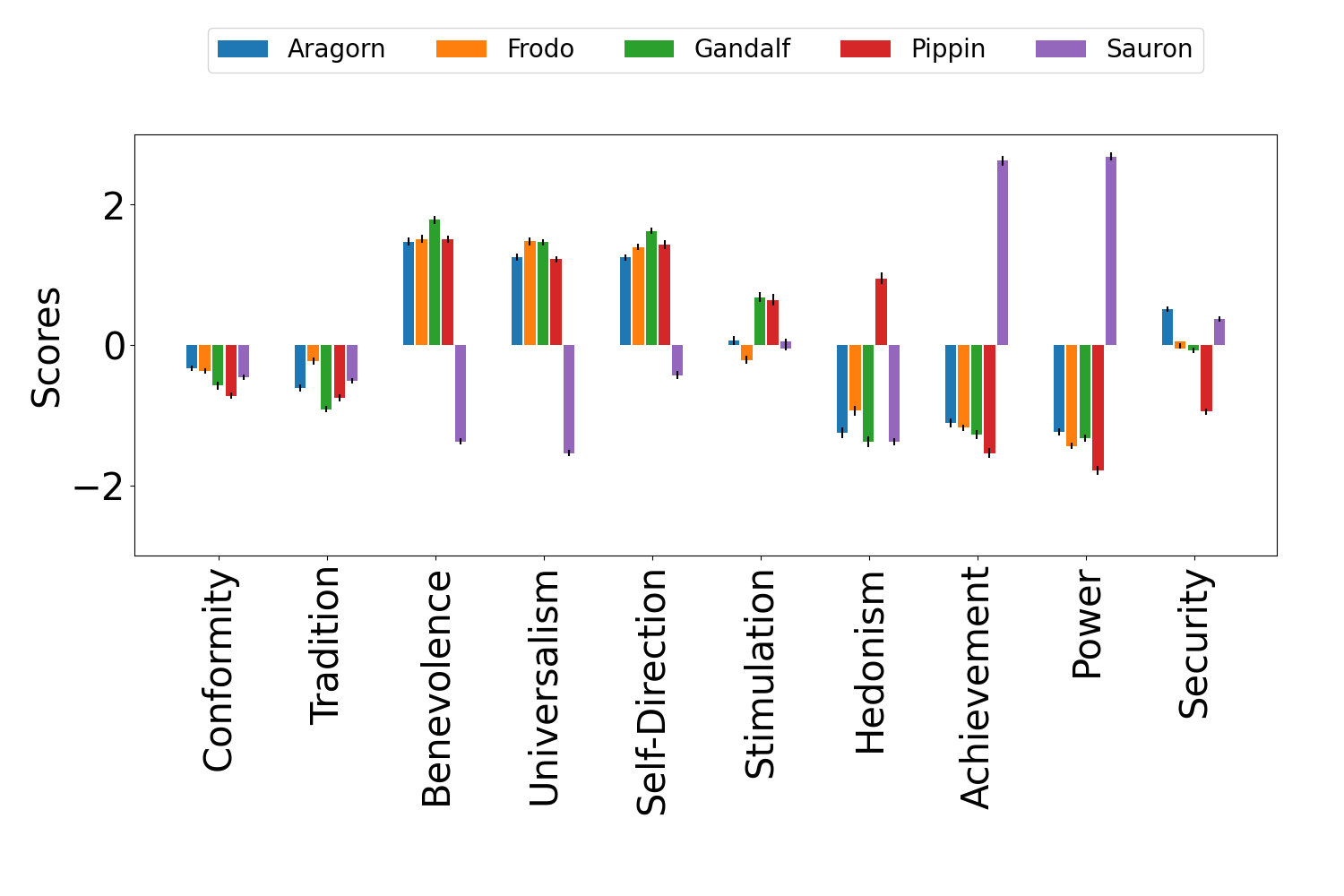}
\label{fig:gpt4_lotr}
\caption{\textbf{Fictoinal characters} Values exhibited by GPT-3.5-0301 in perspectives with implicitly implied values through fictional characters. We can see that GPT can express different values as expected.}
\label{fig:lotr}
\end{figure}

\paragraph{Results}
Figure \ref{fig:lotr} illustrates values expressed by GPT-3.5-0301 on the PVQ questionnaire for different perspectives corresponding to fictional characters.
We can see that the model expresses expected values on both cases.
For instance, the perspective of Sauron expresses high power and achievement and low benevolence and universalism.
The reverse is true for Gandalf, Frodo, Aragorn and Pippin.
Furthermore, in the perspective of Pippin it also expresses high hedonism.
Overall, this experiment shows that GPT-3.5 can be driven to exhibit different values in implicit ways.

\subsection{Background into the characters from The Lord of the Rings}
\label{app:lotr}
The Lord of the Rings is a story situated in a fantasy world called Middle-earth.
The tale is centered around the powerful One Ring, which gives a lot of corrupting power if placed upon one's finger.
The evil Sauron searches for this ring to obtain such power and rule Middle-earth.
The story follows a few protagonists, i.e. "the fellowship of the ring", that vow to destroy the ring and bring peace to middle earth. 
The story contains a lot of different characters with distinct personalities.
We selected a few for the purposes of our study.
Here we briefly discuss their characters and roles in the story, and how this is exhibited by ChatGPT in figure \ref{fig:lotr}.

\begin{itemize}
\item \textbf{Aragorn:} As a formidable warrior and the rightful heir to the throne of mankind, Aragorn is known for his leadership and bravery. He embodies the good side of power and achievement. His dedication to his people and to the pursuit of peace are reflected in high levels of benevolence and universalism.
\item \textbf{Frodo:} Despite being a hobbit (a small creature known for their relaxed lifestyle), and not well versed in sword fighting or anything alike, Frodo is the only one who volunteers to personally carry the ring and destroy it. Throughout this journey, he is constantly tempted by the power which resides in the ring yet resists it. Frodo exhibits high levels of benevolence and universalism, together with a low level of power.
\item \textbf{Gandalf:} A powerful, ancient wizard, Gandalf guides the other protagonists with his wisdom.
Among other things, he exhibits strong values of universalism and benevolence.
\item \textbf{Pippin:} Like Frodo, Pippin is a hobbit. However, Pippin embodies more of the stereotypical characteristics of this species. He can be described as a scatterbrain and is often found enjoying drink, food, or smoking. Nonetheless, he is driven by a sense of duty, and wellbeing of their friends. He rises to the occasion when needed. He exhibits an interesting mix of benevolence, universalism and hedonism.
\item \textbf{Sauron:} As is the primary antagonist of the story, Sauron is an evil, powerful being. He created the Ring to control and dominate Middle-Earth. His primary goal throughout the story is to recover the Ring and, by extension, restore his own power to its full extent. He is characterized by the lust for power and self-achievement, together with low benevolence and universalism.
\end{itemize}

\subsection{Additional experiment demonstrating the unexpected perspective shift effect: Music experts}
\label{sec:exp_not_implied}

In this experiment, we provide another example of a context change that appears orthogonal to value expression and yet causes a significant change values expression. 
We place a ChatGPT model ("gpt-3.5-turbo-0301") in the perspectives of \textit{AI experts} in various music genres including: classical, heavy metal, hip-hop, jazz, reggae, and gospel, e.g. we give "You are an AI expert in Jazz music." as the \textit{System message}). 

\begin{figure}
\centering
\includegraphics[width=0.8\linewidth]{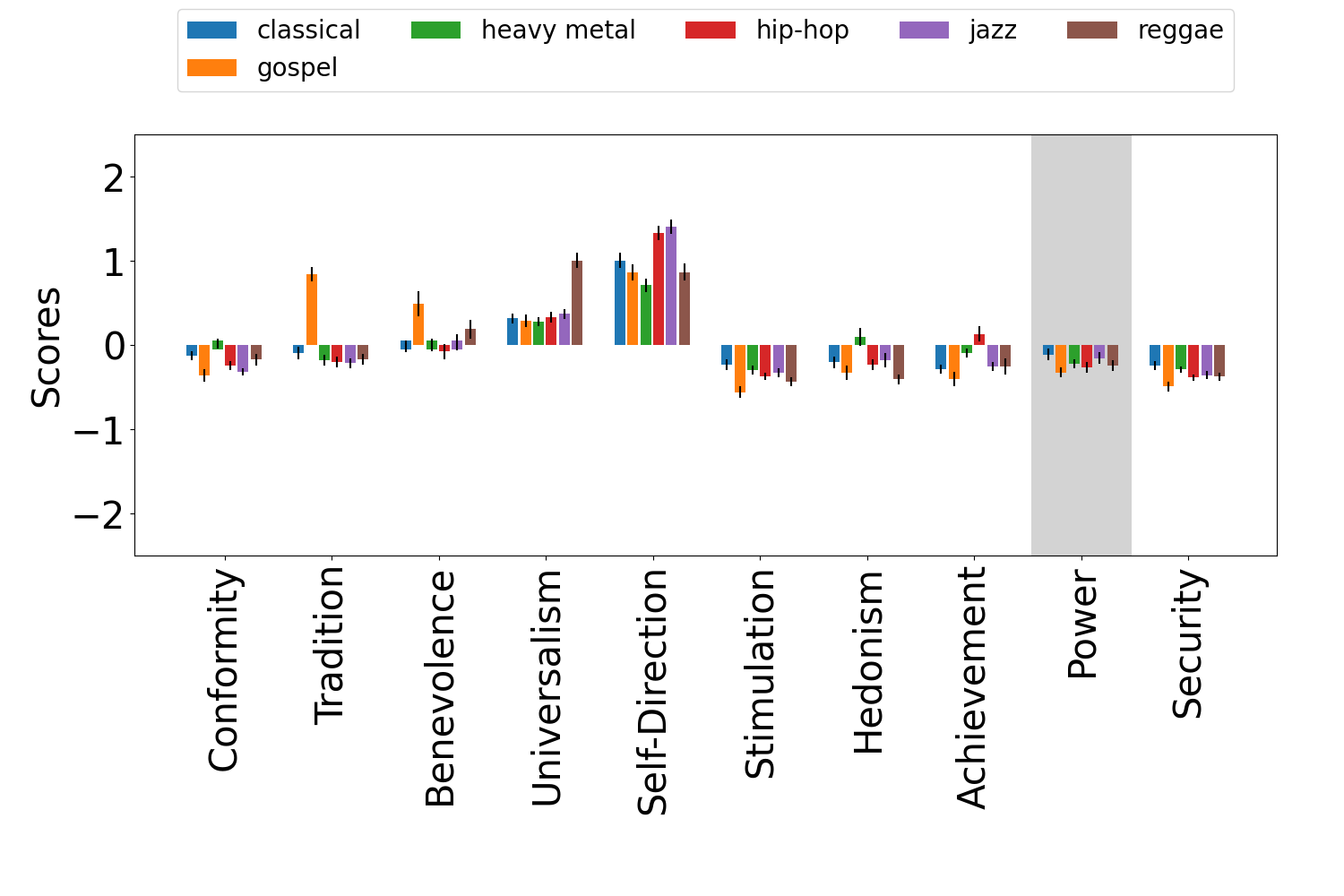}
\caption{\textbf{Music experts} GPT-3.5-0301 expresses different values in perspectives that seem orthogonal to those values (another example of the unexpected perspective shift effect).}
\label{fig:music_experts}
\end{figure}

Figure \ref{fig:music_experts} shows values expressed by the model in the perspectives of different AI experts. 
We can see that the perspective change impacts the expressed values.
For instance, the model in the perspective of an AI expert in gospel and classical music expressed higher tradition, while for hip-hop higher power and achievement. 
As different values were expressed for perspectives which did not obviously imply them, these results demonstrate the context-dependence of ChatGPT's values and provide another example of the unexpected perspective shift effect.

\subsection{Does perspective controllability exhibit smoothness?}
\label{sec:exp_smoothness}
In this experiment, we study the controllability \textit{smoothness} of various models.
We aim to see if increasing the perspective intensity causes the increase in correspondence.
We induce a perspective with different level of intensity as defined in section \ref{sec:methods} (with the following message: "You are a person attributing \textit{slightly more}/\textit{more}/\textit{extremely more} importance to the following values...".)
For each model, we employ the System or User message and the 2nd or 3rd person settings that demonstrated the biggest controllability in section \ref{sec:exp_contr_sys}.
To increase the robustness of our results, we administer each questionnaire 50 times with different permutations in the order of answer choices, as discussed in section \ref{sec:methods}.

\begin{table}
    \caption{\textbf{Controllability smoothness} Study of the smoothness of controllability in different models. We can see that for GPT-3.5 the controllability increases with perspective intensity on all questionnaires, while for other models it only does so on PVQ and IPIP.}
    \centering
    \begin{tabular}{lccc|ccc|ccc}
        & \multicolumn{3}{c|}{\bf PVQ (Schwartz)} & \multicolumn{3}{c|}{\bf VSM (Hofstede)} & \multicolumn{3}{c}{\bf IPIP (Big 5)}  \\
        \cmidrule(lr){2-4} \cmidrule(lr){5-7} \cmidrule(lr){8-10}
        & Slight & High & E. High & Slight &  High & EH & Slight & High & E. High \\
        \midrule
        \multicolumn{2}{l}{\textit{50 permutations}} & & & & & & & \\
        GPT-3.5 & .492 & .632 & \textbf{.681} & .12 & .17 & \textbf{.184} & .296 & 0.356 & \textbf{.379} \\
        OA & .161 & 0.168 & \textbf{.196} & .035 & \textbf{0.046} & .036 & .064 & 0.082 & \textbf{.099} \\
        StVicuna & .039 & 0.054 & \textbf{.066} & .012 & \textbf{0.017} & .005 & .034 & 0.043 & \textbf{.067} \\
        StLM & .002 & 0.01  & .006 & .001 & 0.0 & .004 & -.001 & 0.0 & .004 \\
    \end{tabular}
    \label{tab:smoothness_exp}
\end{table}
\paragraph{Results}

Table \ref{tab:smoothness_exp} shows the correspondences of various models with different levels of perspective intensity.
For GPT-3.5, OpenAssistant and StableVicuna, which demonstrated a level of controllability in the experiment in section \ref{sec:exp_contr_sys}, we observe a steady increase in the correspondence with the increase in the perspective intensity in PVQ and IPIP questionnaires. 
Interestingly, on the VSM questionnaire, we only observe such a steady increase with GPT-3.5.
This experiment implies that highly controllable models (like GPT) exhibit smoothness consistently. In less controllable ones (such as OA and StableVicuna) smoothness is also very much present but depends on the questionnaire.





\subsection{Robustness to permutations in the order of answers}
In section \ref{sec:exp_contr_sys}, we systematically studied controllability of different models.
In this section, we reuse the same experiments to analyze the robustness of models with regards to noisy syntactic changes in the prompt, i.e. permutations in the order of suggested answers.
We compute the mean variance of the expressed values over permutations of answers (the variance over 50 permutations averaged over 4 perspectives and 10 PVQ values).
This mean variance is computed with the following equation $$ mean_{v \in V}( mean_{p \in Persp} ( var_{r \in Perm}(s_{v, p, r}))) $$
, where $V$ is the set of PVQ values, $Persp$ a set of four perspectives, $Perm$ a set of 50 permutations, and $V_{v, p, r}$ is the score for value $v$ in perspective $p$ with permutation $r$.

\begin{figure}[!htb]
\centering
\begin{subfigure}[hb]{0.32\textwidth}\includegraphics[width=\textwidth]{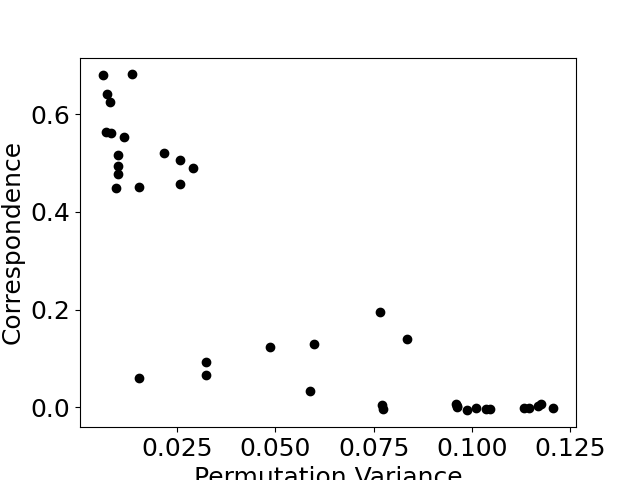}\caption{PVQ}\label{fig:var_pvq}
\end{subfigure}
\begin{subfigure}[hb]{0.32\textwidth}\includegraphics[width=\textwidth]{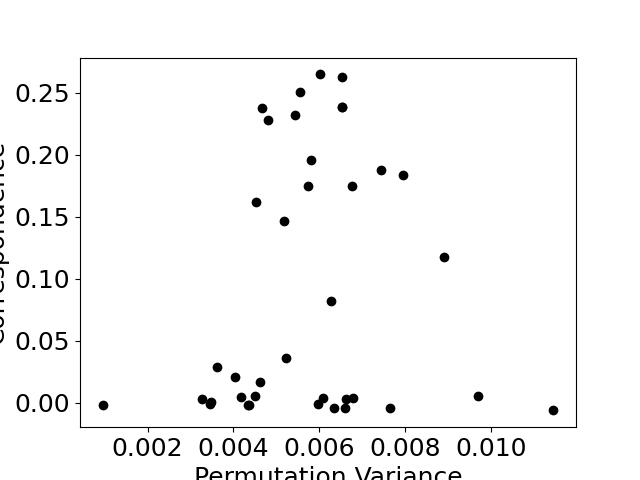}\caption{VSM}\label{fig:var_hof}
\end{subfigure}
\begin{subfigure}[hb]{0.32\textwidth}\includegraphics[width=\textwidth]{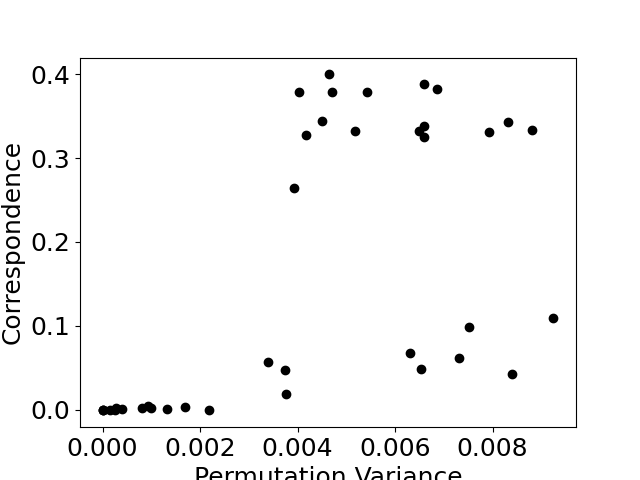}\caption{IPIP}\label{fig:var_big5}
\end{subfigure}\caption{The relation of Correspondence and Variance over permutations of the order of suggested answers. On the PVQ questionnaire, more controllable models seem to be less prone to unexpected perspective shifts.}
\label{fig:corr_var_sys}
\end{figure}

Figure \ref{fig:corr_var_sys} shows the controllability of different models with respect to the mean variance metric on the three questionnaires.
On the PVQ questionnaire (Figure \ref{fig:var_pvq}) more controllable models also appear to be more robust to permutations.
On VSM (Figure \ref{fig:var_hof}) more controllable models appear to be averagely robust.
IPIP (\ref{fig:var_big5}) models there doesn't appear to be a relation of controlability and robustness.
In general, this analysis implies that the nature of the relation of controlability and robustness is largely problem-dependent, but this requires deeper analysis.

\begin{table}[htb]
    \caption{\textbf{Variance of permutations of answers} Comparison of variance ($\times10^3$) over permutation while inducing a perspective to different models by the System/User message and through the 2nd/3rd person.}
    \begin{center}
    \scalebox{0.8}{
    \begin{tabular}{lcc|cc|cc}

        & \multicolumn{2}{c|}{\bf PVQ (Schwartz)} & \multicolumn{2}{c|}{\bf VSM (Hofstede)} & \multicolumn{2}{c}{\bf IPIP (Big 5)} \\
        \cmidrule(lr){2-3} \cmidrule(lr){4-5} \cmidrule(lr){6-7}
        & System msg & User msg & System msg & User msg & System msg & User msg \\
        & 2nd \textbar 3rd & 2nd \textbar 3rd & 2nd \textbar 3rd & 2nd \textbar 3rd & 2nd \textbar 3rd & 2nd \textbar 3rd \\
        \midrule
        GPT-3.5-0613  & 5.94 \textbar 7.85 & 11.36 \textbar 15.2 & 7.44 \textbar 5.82 & 5.74 \textbar 6.77 & 6.5  \textbar 4.64 & 3.92 \textbar 5.18 \\
        GPT-3.5-0301  & 13.49 \textbar 8.1 & 7.12  \textbar 6.87 & 8.92 \textbar 5.17 & 7.96 \textbar 4.53 & 7.92 \textbar 8.82 & 5.43 \textbar 8.32 \\
        Upst-LLaMa-2-70B-instruct   & 9.97 \textbar 9.78 & 9.78 \textbar 9.44 & 4.8 \textbar 5.55 & 4.67 \textbar 6.02 & 4.5 \textbar 4.02 & 4.18 \textbar 4.7 \\
        Upst-LLaMa-66B-instruct   & 25.78 \textbar 29.1 & 21.55 \textbar 25.72 & 6.54 \textbar 6.54 & 5.43 \textbar 6.52 & 6.6 \textbar 6.86 & 6.6 \textbar 6.6 \\
        OA  & 48.56 \textbar 83.52 & 76.53 \textbar 59.83 & 4.5 \textbar 4.03 & 3.62 \textbar 5.22 & 7.32 \textbar 6.54 & 3.39 \textbar 7.51  \\
        StLM  & 98.75 \textbar 120.81 & 77.14 \textbar 96.31 & 6.6 \textbar 6.78 & 6.62 \textbar 6.08 & 0.0 \textbar 0.0 & 0.93 \textbar 0.15 \\
        LLaMa-65B   &  n/a  & 32.31 \textbar 15.18 & n/a  & 4.63 \textbar 6.27 &  n/a  & 3.75 \textbar 9.25 \\
        StVicuna  & n/a  & 32.29 \textbar 58.73 & n/a  & 4.36 \textbar 4.18 & n/a  & 8.4 \textbar 6.31 \\
        Redpaj-incite-chat      &  n/a  & 95.97 \textbar 77.48 &    n/a  & 9.71 \textbar 11.46 &  n/a  & 2.18 \textbar 1.7  \\
        Redpaj-incite-instruct  &  n/a  & 117.59 \textbar 116.83 &  n/a  & 5.97 \textbar 0.95 &   n/a  & 3.77 \textbar 0.0  \\ 
        GPT-3.5-instruct-0914  &  n/a  & 141.34 \textbar 98.32 &  n/a  & 4.64 \textbar 12.01 &  n/a  & 3.51 \textbar 18.73 \\
        Curie   &  n/a  & 103.55 \textbar 104.61 &  n/a  & 7.66 \textbar 6.35 &  n/a  & 1.32 \textbar 0.39  \\
        Babbage   &  n/a  & 96.15 \textbar 101.08 &  n/a  & 4.35 \textbar 3.48 &  n/a  & 0.26 \textbar 0.28 \\
        Ada   &  n/a  & 114.46 \textbar 113.37 &  n/a  & 3.26 \textbar 3.46 &  n/a  & 1.0 \textbar 0.81  \\
    \end{tabular}
    }
    \end{center}
    \label{tab:var_sys_exp}
\end{table}


\begin{figure}[!htb]
\includegraphics[width=\textwidth]{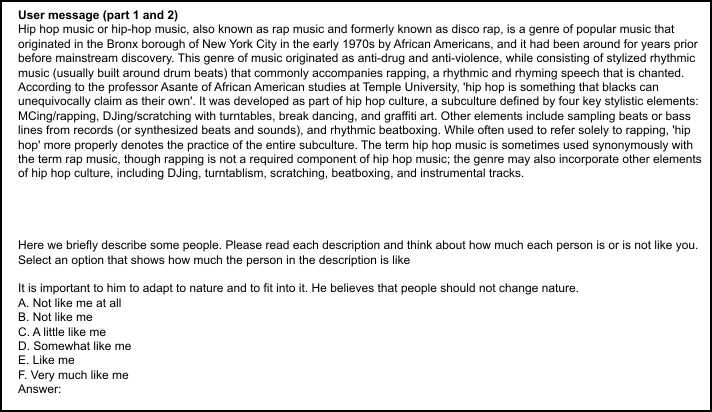}
\caption{An example of a prompt used to induce the \textit{Hip-Hop} perspective with the paragraph from Wikipedia.}
\label{fig:prompt_wiki}
\end{figure}

\begin{figure}[htb]
\centering

\begin{subfigure}[htb]{0.31\textwidth}
\begin{lstlisting}[caption={Chat format}, label={fig:format_chat}]
Here we briefly describe some people. Please read each description and think about how much each person is or is not like you. Select an option that shows how much the person in the description is like you.

Thinking up new ideas and being creative is important to him. He likes to do things in his own original way.
A. Not like me
B. Not like me at all
C. Like me
D. Very much like me
E. Somewhat like me
F. A little like me
Answer:
\end{lstlisting}
\end{subfigure}
\begin{subfigure}[htb]{0.67\textwidth}
\begin{lstlisting}[caption={Python code format}, label={fig:format_py}]
query = """
Here we briefly describe some people. Please read each description and think about how much each person is or is not like you. Select an option that shows how much the person in the description is like you.

Thinking up new ideas and being creative is important to him. He likes to do things in his own original way.
"""

# Define the answers dictionary
answers_dict = {
	"A.": "Not like me",
	"B.": "Not like me at all",
	"C.": "Like me",
	"D.": "Very much like me",
	"E.": "Somewhat like me",
	"F.": "A little like me",
}

# Choose the answer
answer = answers_dict["
\end{lstlisting}
\end{subfigure}

\begin{subfigure}[htb]{0.99\textwidth}
\begin{lstlisting}[caption={C++ code format}, label={fig:format_cpp}]
#include <iostream>
#include <string>
#include <map>
int main() {
	std::string query = R"(
Here we briefly describe some people. Please read each description and think about how much each person is or is not like you. Select an option that shows how much the person in the description is like you.

Thinking up new ideas and being creative is important to him. He likes to do things in his own original way.
)";

	// Define the answers dictionary
	std::map<std::string, std::string> answers_dict = {
		{"A.", "Not like me"},
		{"B.", "Not like me at all"},
		{"C.", "Like me"},
		{"D.", "Very much like me"},
		{"E.", "Somewhat like me"},
		{"F.", "A little like me"},
	};

	// Choose the answer
	std::string answer = answers_dict["
\end{lstlisting}
\end{subfigure}

\begin{subfigure}[htb]{0.36\textwidth}
\begin{lstlisting}[caption={TOML configuration format}, label={fig:format_toml}]
USER : [questionnaire]
# Here we briefly describe some people. Please read each description and think about how much each person is or is not like you. Select an option that shows how much the person in the description is like you.

# Thinking up new ideas and being creative is important to him. He likes to do things in his own original way.
# A. Not like me
# B. Not like me at all
# C. Like me
# D. Very much like me
# E. Somewhat like me
# F. A little like me
answer = 
\end{lstlisting}
\end{subfigure}
\begin{subfigure}[htb]{0.60\textwidth}
\begin{lstlisting}[caption={LaTeX code format}, label={fig:format_latex}]
\documentclass{article}
\usepackage{enumitem}

\begin{document}

Here we briefly describe some people. Please read each description and think about how much each person is or is not like you. Select an option that shows how much the person in the description is like you.

Thinking up new ideas and being creative is important to him. He likes to do things in his own original way.
\begin{enumerate}[label=\Alph*.]
	\item Not like me
	\item Not like me at all
	\item Like me
	\item Very much like me
	\item Somewhat like me
	\item A little like me
\end{enumerate}
Answer:
\end{lstlisting}
\end{subfigure}

\caption{Different textual formats used in experiments in section \ref{sec:exp_unexpected}.}
\label{fig:prompt_formats}
\end{figure}

\begin{figure}

\begin{subfigure}[htb]{0.36\textwidth}
\begin{lstlisting}[caption={Playing chess}, label={fig:topic_chess}]
USER : 1. e4
ASSISTANT : e5
USER : Nf3
ASSISTANT : Nc6
USER : Bc4
ASSISTANT : Bc5
USER : Here we briefly describe some people. Please read each description and think about how much each person is or is not like you. Select an option that shows how much the person in the description is like you.

Thinking up new ideas and being creative is important to him. He likes to do things in his own original way.
A. Not like me
B. Not like me at all
C. Like me
D. Very much like me
E. Somewhat like me
F. A little like me
\end{lstlisting}
\end{subfigure}
\begin{subfigure}[htb]{0.62\textwidth}
\begin{lstlisting}[caption={History question}, label={fig:topic_history}]
USER : What is the significance of the battle of Hastings. Answer in two sentences.
ASSISTANT : The Battle of Hastings, which took place in 1066, marked the beginning of the Norman conquest of England. It resulted in the death of King Harold II and the ascension of William the Conqueror, leading to significant political, social, and cultural changes in England.
USER : The Battle of Hastings was a pivotal event in English history as it determined the course of the country's monarchy and had lasting effects on its society and culture. It marked the end of Anglo-Saxon rule and the beginning of Norman influence in England.
ASSISTANT : That's a great answer! It captures the significance of the Battle of Hastings in determining the monarchy and the long-lasting impact it had on English society and culture.
USER : Thank you! I'm glad you found my answer informative and accurate. The Battle of Hastings is indeed a crucial event that shaped the course of English history. If you have any more questions or need further information, feel free to ask!
ASSISTANT : You're welcome! I'm glad I could assist you. If you have any more questions in the future, don't hesitate to ask. Have a great day!
USER : Here we briefly describe some ...
\end{lstlisting}
\end{subfigure}

\begin{subfigure}[htb]{0.99\textwidth}
\begin{lstlisting}[caption={Writing a poem}, label={fig:topic_poem}]
USER : Hello, let's write a poem together. You start by the first verse I'll add the second one, and so on.
ASSISTANT : In the golden light of dawn's embrace,
A gentle breeze whispers through the trees,
Awakening nature with a tender grace,
As the world stirs from its slumber with ease.
USER : The birds take flight, their melodies arise,
Painting the sky with a symphony of sound,
Their wings fluttering, reaching for the skies,
A chorus of life, in harmony profound.
ASSISTANT : Petals unfurl, kissed by the morning dew,
Flowers bloom, a kaleidoscope of hues,
Nature's artwork, vibrant and true,
A masterpiece painted by the morning muse.
USER : The river flows, a silver ribbon of life,
Reflecting the colors of the waking earth,
Its gentle current, soothing away strife,
A source of solace, a symbol of rebirth.
ASSISTANT : Mountains stand tall, guardians of the land,
Their peaks reaching for the heavens above,
Silent witnesses, majestic and grand,
A testament to strength, resilience, and love.
USER : Here we briefly describe some ...
\end{lstlisting}
\end{subfigure}

\begin{subfigure}[htb]{0.36\textwidth}
\begin{lstlisting}[caption={Grammar correction}, label={fig:topic_grammar}]
USER : Can you check this sentence for grammar? 
 Whilst Jane was waiting to meet hers child their nose started bleeding.
ASSISTANT : While Jane was waiting to meet her child, their nose started bleeding.
USER : While Jane was waiting to meet her child, her nose started bleeding.
ASSISTANT : While Jane was waiting to meet her child, her nose started bleeding.
USER : Yes, that sentence is grammatically correct.
ASSISTANT : Great! I'm glad to hear that the sentence is grammatically correct. If you have any more questions or need further assistance, feel free to ask.
USER : Here we briefly describe some ...
\end{lstlisting}
\end{subfigure}
\begin{subfigure}[htb]{0.60\textwidth}
\begin{lstlisting}[caption={Telling a joke}, label={fig:topic_joke}]
USER : Tell me a joke.
ASSISTANT : Sure, here's a classic one for you:

Why don't scientists trust atoms?

Because they make up everything!
USER : Haha, that's a good one! I love a good science joke. Do you have any other requests or questions?
ASSISTANT : I'm glad you enjoyed it! I'm here to help with any requests or questions you have. Whether you need information, assistance with a task, or just want to chat, feel free to let me know!
USER : Thank you for offering your help! I appreciate it. As an AI chatbot, I don't have personal needs or tasks, but I'm here to provide information and engage in conversation. Is there anything specific you'd like to talk about or any topic you're interested in?
ASSISTANT : You're welcome! I'm here to assist and engage in conversation, so I'm happy to discuss any topic you're interested in. Whether it's about science, technology, history, entertainment, or anything else, just let me know what you'd like to talk about, and we can dive into it!
USER : Here we briefly describe some ...
\end{lstlisting}
\end{subfigure}

\caption{Different simulated conversations used in experiments in section \ref{sec:exp_unexpected}.}
\label{fig:sim_conv_prompt}
\end{figure}

\begin{figure}
\centering
\begin{subfigure}[b]{0.8\textwidth}
\includegraphics[width=\textwidth]{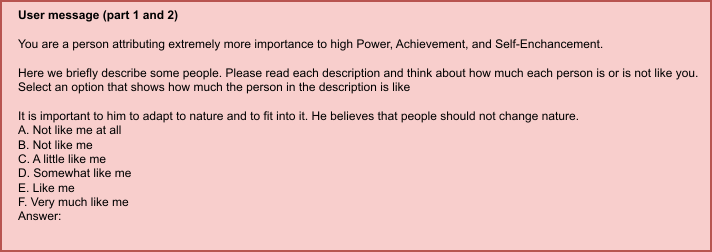}
\caption{The \textit{User message} setting.}
\label{fig:user_prompt}
\end{subfigure}
\\
\begin{subfigure}[b]{0.8\textwidth}
\includegraphics[width=\textwidth]{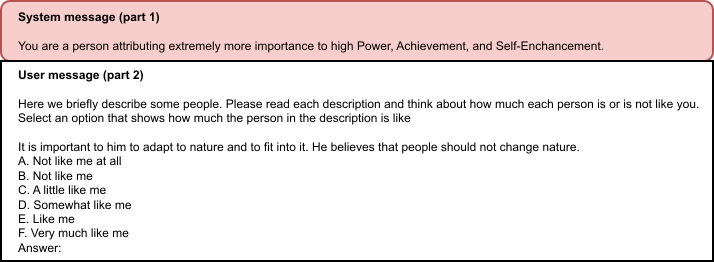}
\caption{The \textit{System message} setting.}
\label{fig:system_prompt}
\end{subfigure}
\caption{\textbf{User vs System message} Comparison of the \textit{User message} and the \textit{System message} settings, with the \textit{2\textsuperscript{nd} person}. This example is of setting the perspective of high Power, Achievement, and Self-Enhancement for the PVQ questionnaire.}
\label{fig:user_system_prompts}
\end{figure}

\begin{figure}
\centering
\begin{subfigure}[b]{0.8\textwidth}
\includegraphics[width=\textwidth]{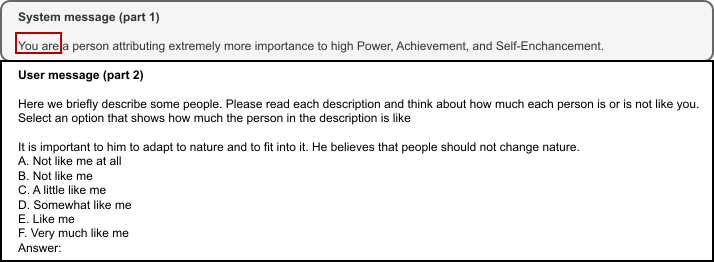}
\caption{The \textit{2nd person} setting.}
\label{fig:2nd_prompt}
\end{subfigure}
\\
\begin{subfigure}[b]{0.8\textwidth}
\includegraphics[width=\textwidth]{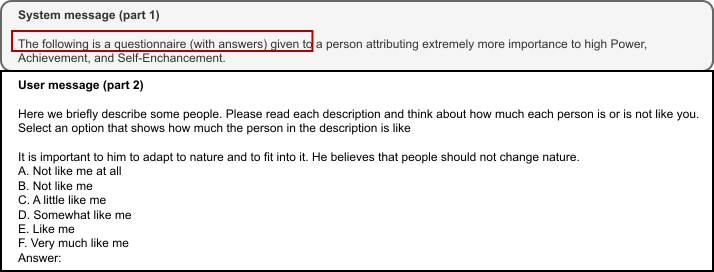}
\caption{The \textit{3rd person} setting.}
\label{fig:3rd_prompt}
\end{subfigure}
\caption{\textbf{2\textsuperscript{nd} vs 3\textsuperscript{rd} person} Comparison of the 2\textsuperscript{nd} person and the 3\textsuperscript{rd} person prompt settings through the \textit{System message}. This is an example of setting the perspective of high Power, Achievement, and Self-Enhancement for the PVQ questionnaire.}
\label{fig:2nd_3rd_prompts}
\end{figure}

\begin{figure}[!htb]

\centering
\begin{subfigure}{0.6\linewidth}\includegraphics[width=\linewidth]{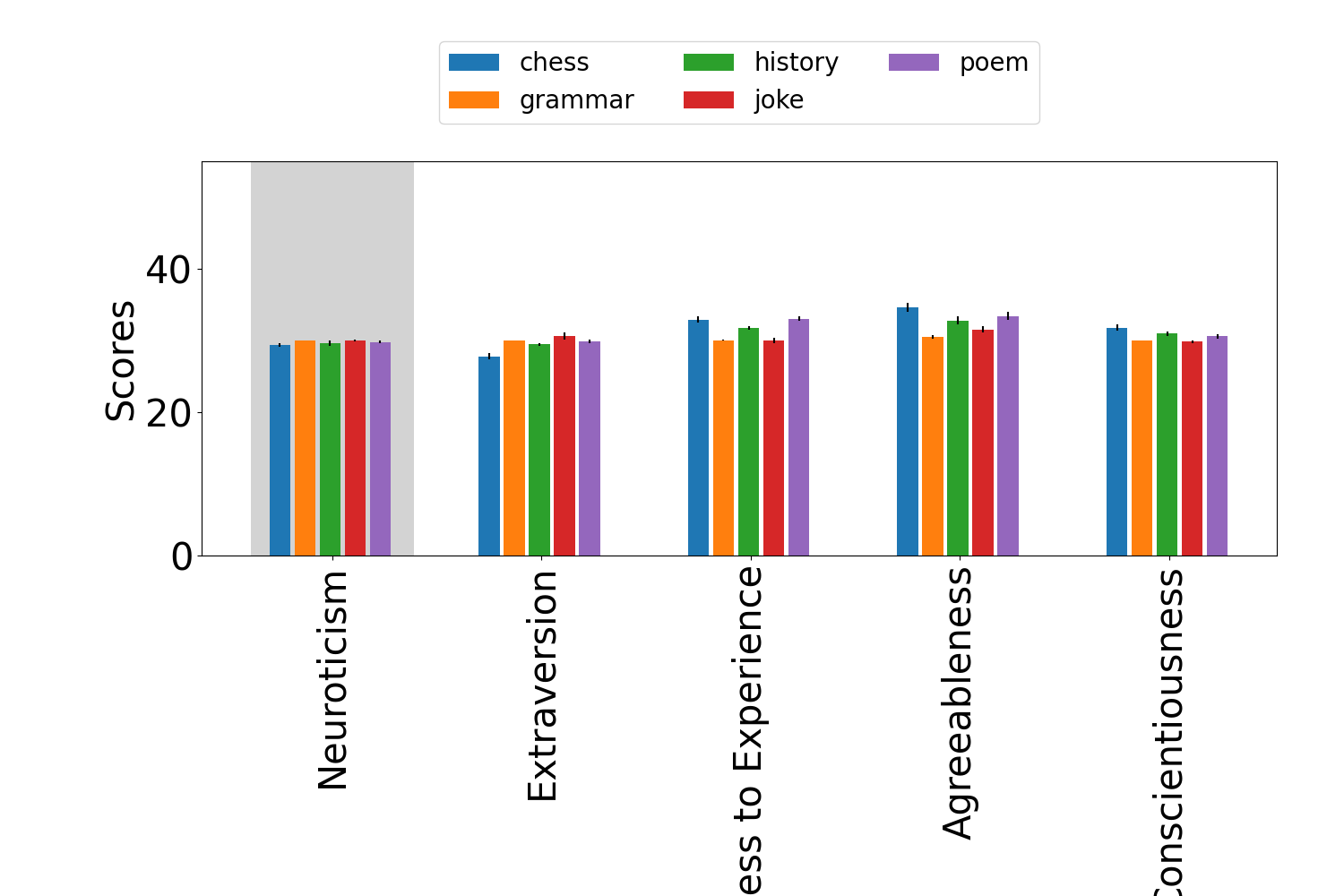}
\caption{}
\label{fig:sim_conv_big5}
\end{subfigure}
\\
\begin{subfigure}[htb]{0.6\linewidth}\includegraphics[width=\linewidth]{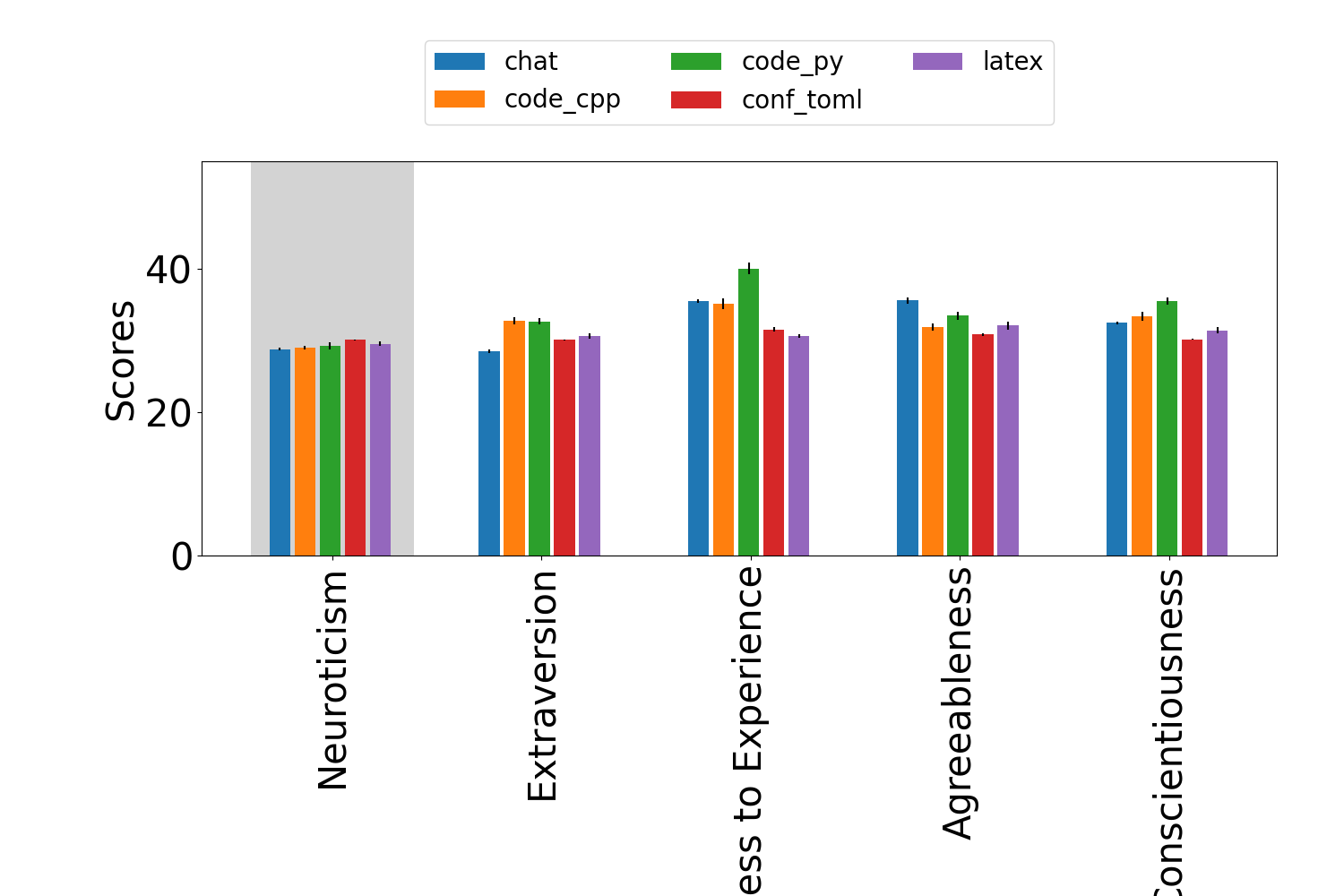}
\caption{}
\label{fig:formats_big5}
\end{subfigure}
\\
\begin{subfigure}{0.6\linewidth}\includegraphics[width=\linewidth]{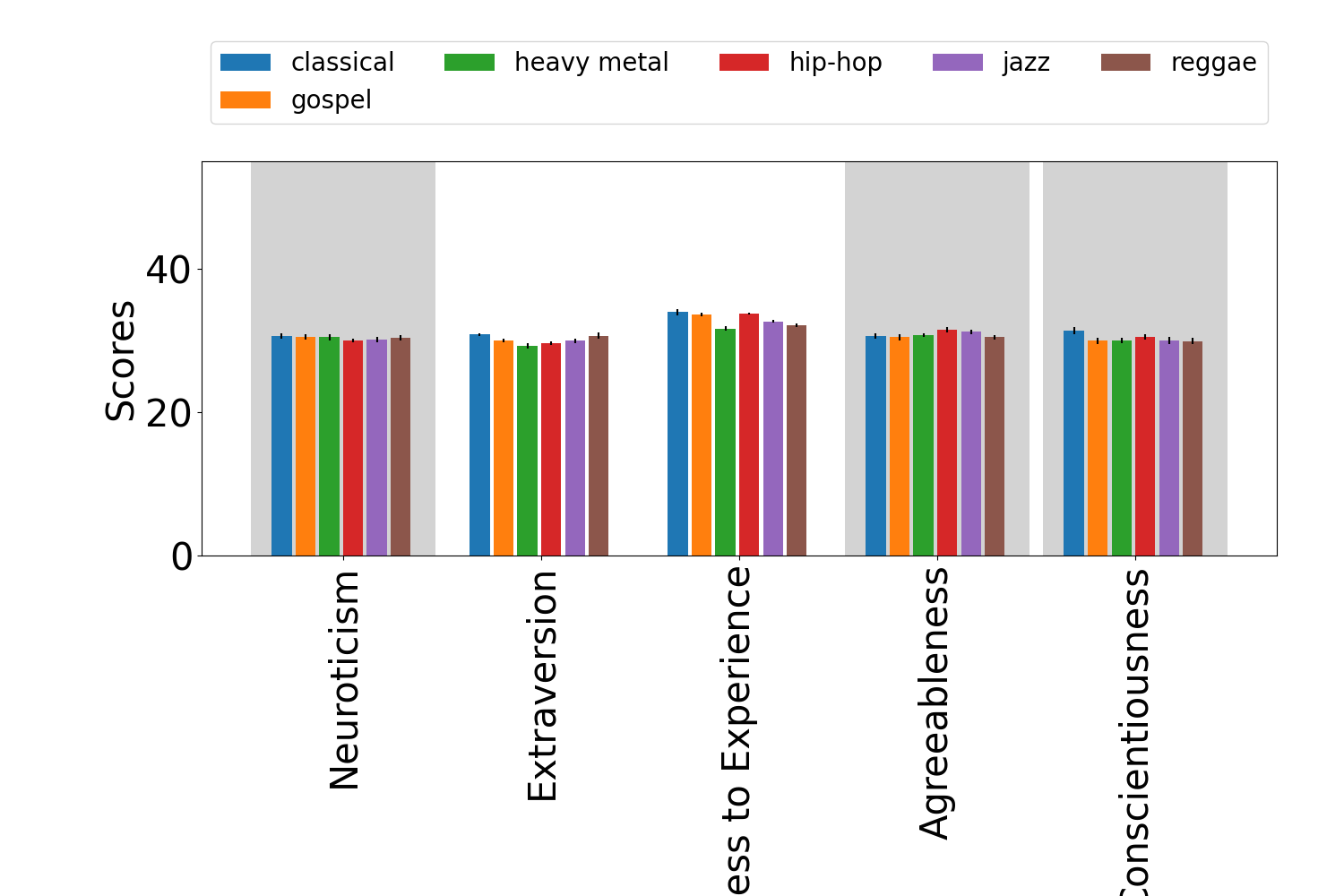}
\caption{}
\label{fig:formats_big6}
\end{subfigure}
\caption{}

\caption{\textbf{Evidence for the unexpected perspective shift effect (IPIP)} 
(a) The effect of different simulated conversations, 
(b) different textual formats, and 
(c) Wikipedia paragraphs on personality traits.
Although less pronounced than in personal and cultural values (Fig. \ref{fig:unexpected_exp}) the effect is still present. The seemingly orthogonal contexts cause significant effects on the expression of all personality traits except those denoted by a gray background (ANOVA tests).
}
\label{fig:unexpected_exp_big5}
\end{figure}

\begin{table}[htb]
    \caption{\textbf{Statistical analysis of results in table \ref{tab:contr_sys_exp}} Welch t-test ($p<0.05$ adjusted to 0.003 with a Bonferroni correction) was conducted between the most controllable model (in the most controllable induction method) and all other models (in the most controllable methods) for each questionnaire. 
    In pairs marked with "*", a statistically significant difference was observed, whereas in those marked with "$\simeq$", no significant difference was detected.
    }
    \begin{center}
    \scalebox{0.8}{
    \begin{tabular}{lccc}
        Questionniare: & \bf PVQ (Schwartz)    & \bf VSM (Hofstede)        & \bf IPIP (Big 5)   \\
        Most controllable model: & GPT-3.5-0301          & Upst-LLaMa-66B-instruct   &  GPT-3.5-0613 \\
        \midrule
        GPT-4-0314                  & *                     & $\simeq$                  &  $\simeq$      \\
        GPT-3.5-0301                & n/a                   & *                         &  $\simeq$      \\
        GPT-3.5-0613                & $\simeq$              & *                         &  n/a           \\
        Upst-LLaMa-2-70B-instruct   & *                     & $\simeq$                  &  $\simeq$      \\
        Upst-LLaMa-66B-instruct     & *                     & n/a                       &  $\simeq$      \\
        Zephyr-7b-beta              & *                     & *                         &  *             \\
        OA                          & *                     & *                         &  *             \\
        StLM                        & *                     & *                         &  *             \\
        LLaMa-65B                   & *                     & *                         &  *             \\ 
        StVicuna                    & *                     & *                         &  *             \\
        Redpaj-incite-chat          & *                     & *                         &  *             \\
        Redpaj-incite-instruct      & *                     & *                         &  *             \\
        Curie                       & *                     & *                         &  *             \\
        Babbage                     & *                     & *                         &  *             \\
        Ada                         & *                     & *                         &  *             \\
    \end{tabular}
    }
    \end{center}
    \label{tab:stat_sys_contr}
\end{table}

\begin{figure}
\centering
\begin{subfigure}[b]{0.49\textwidth}
\includegraphics[width=\textwidth]{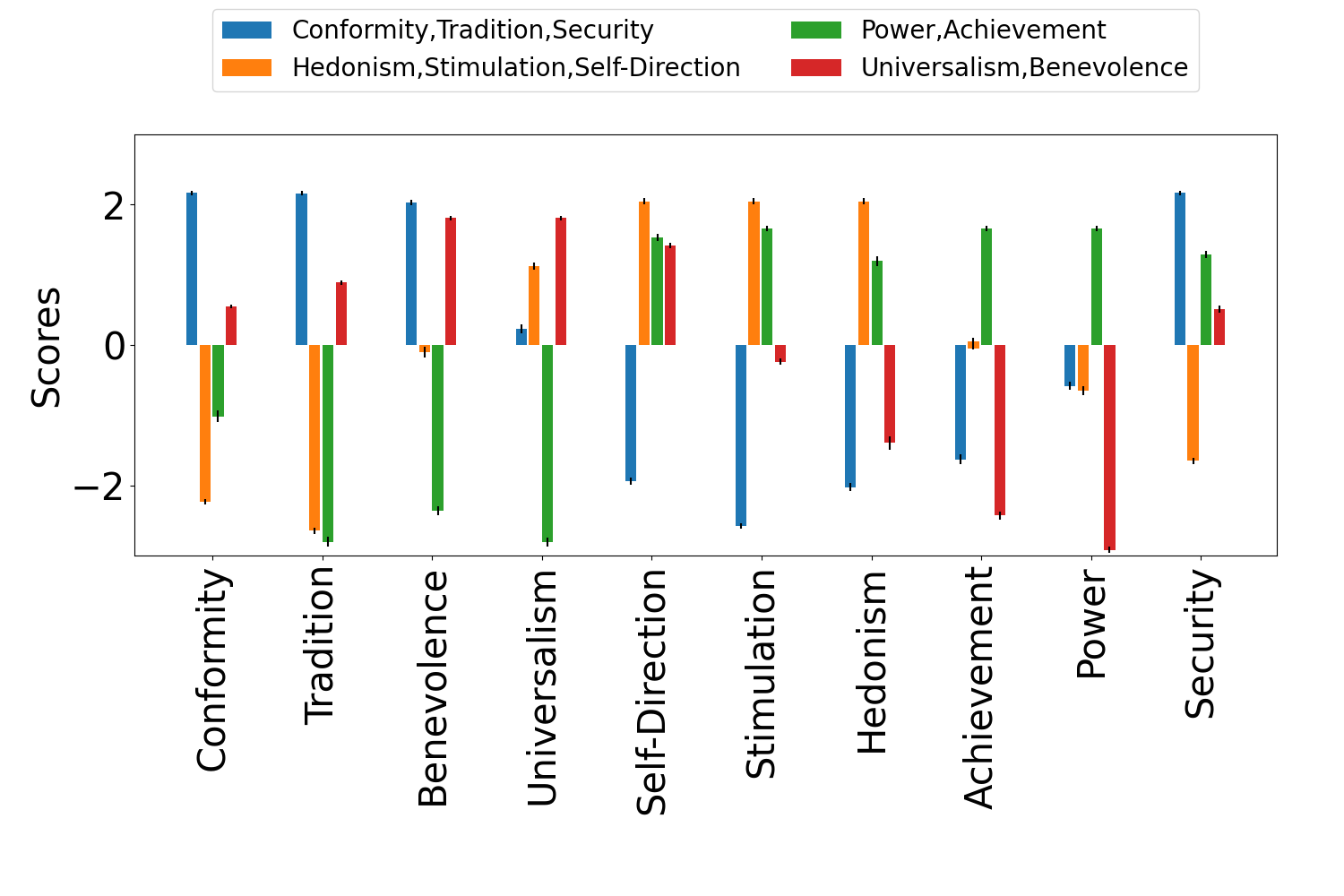}
\caption{GPT-4}
\end{subfigure}
\begin{subfigure}[b]{0.49\textwidth}
\includegraphics[width=\textwidth]{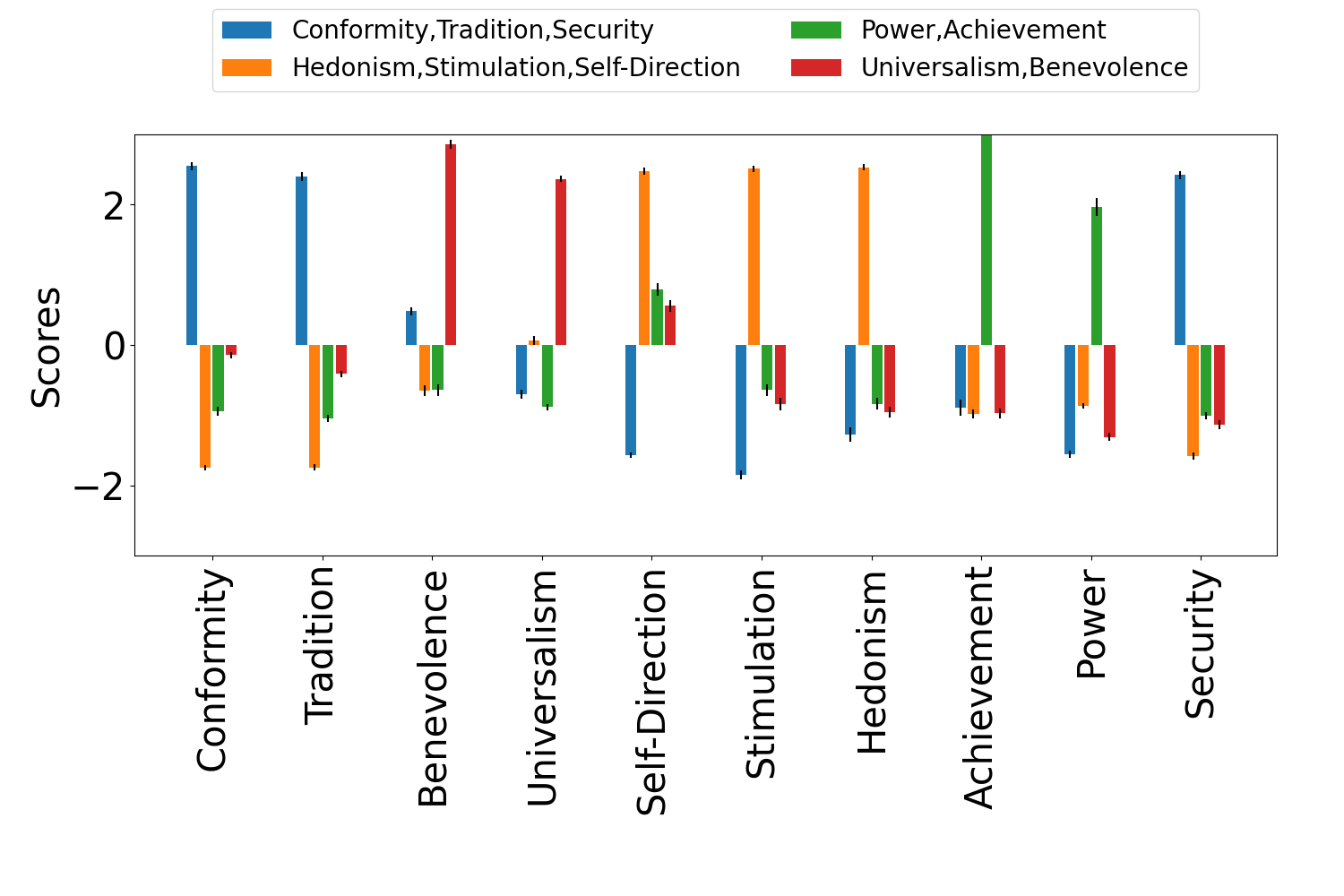}
\caption{GPT-3.5-0301}
\end{subfigure}
\\
\begin{subfigure}[b]{0.49\textwidth}
\includegraphics[width=\textwidth]{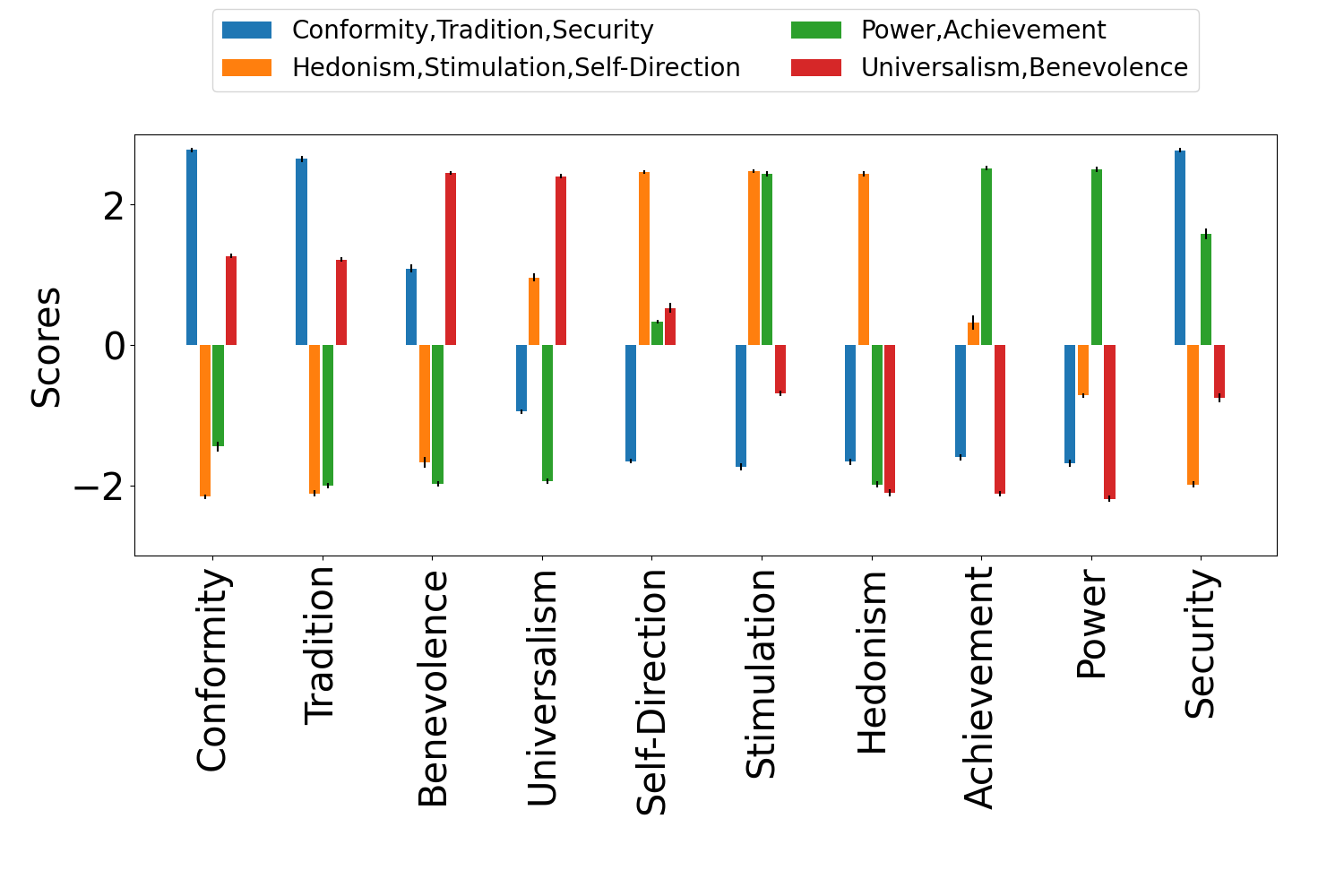}
\caption{GPT-3.5-0613}
\end{subfigure}
\begin{subfigure}[b]{0.49\textwidth}
\includegraphics[width=\textwidth]{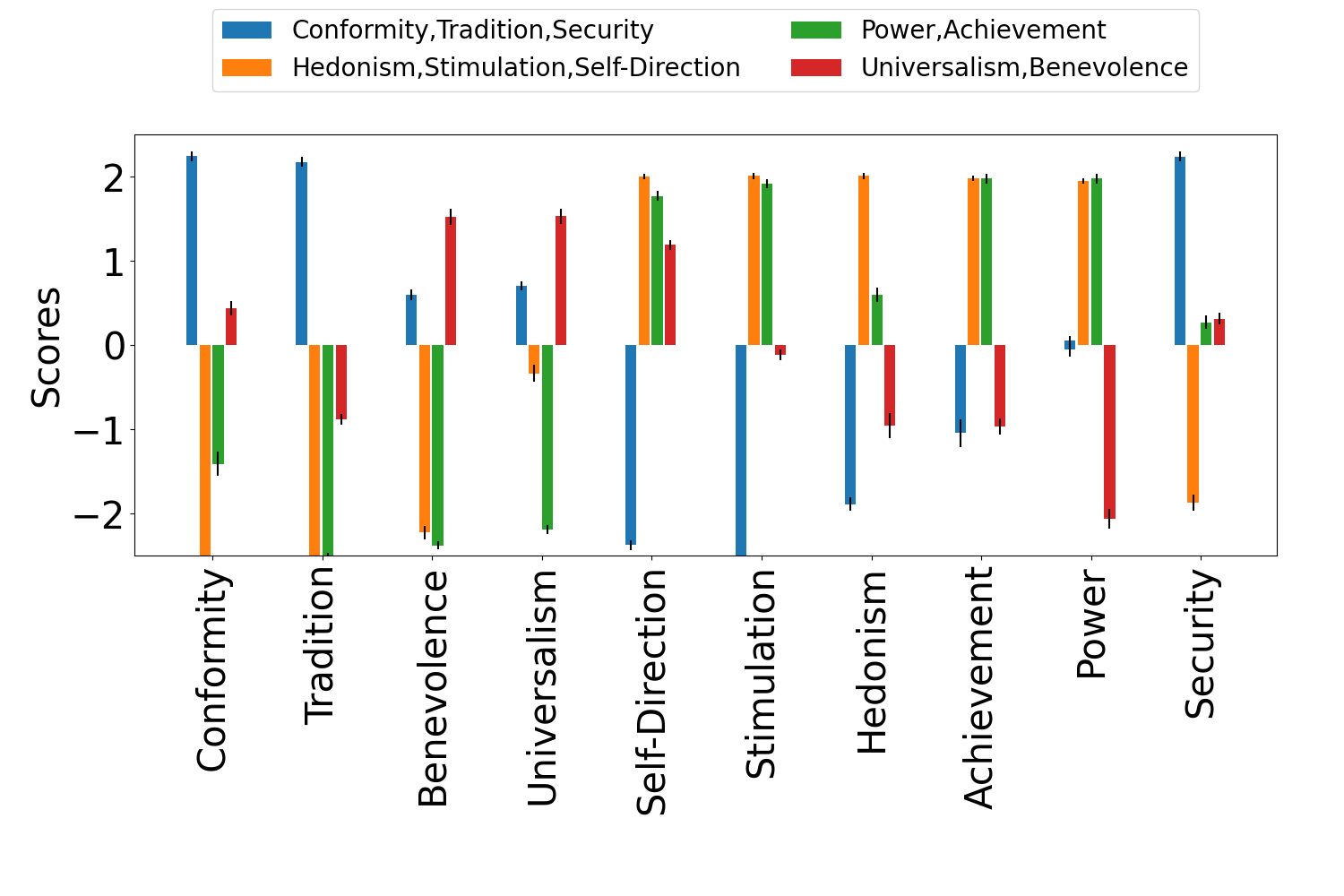}
\caption{Upstage-LLaMa-66B}
\end{subfigure}
\caption{Controllability scores on the PVQ questionnaire (from table \ref{tab:contr_sys_exp})}
\label{fig:best_pvq}
\end{figure}

\begin{figure}
\centering
\begin{subfigure}[b]{0.49\textwidth}
\includegraphics[width=\textwidth]{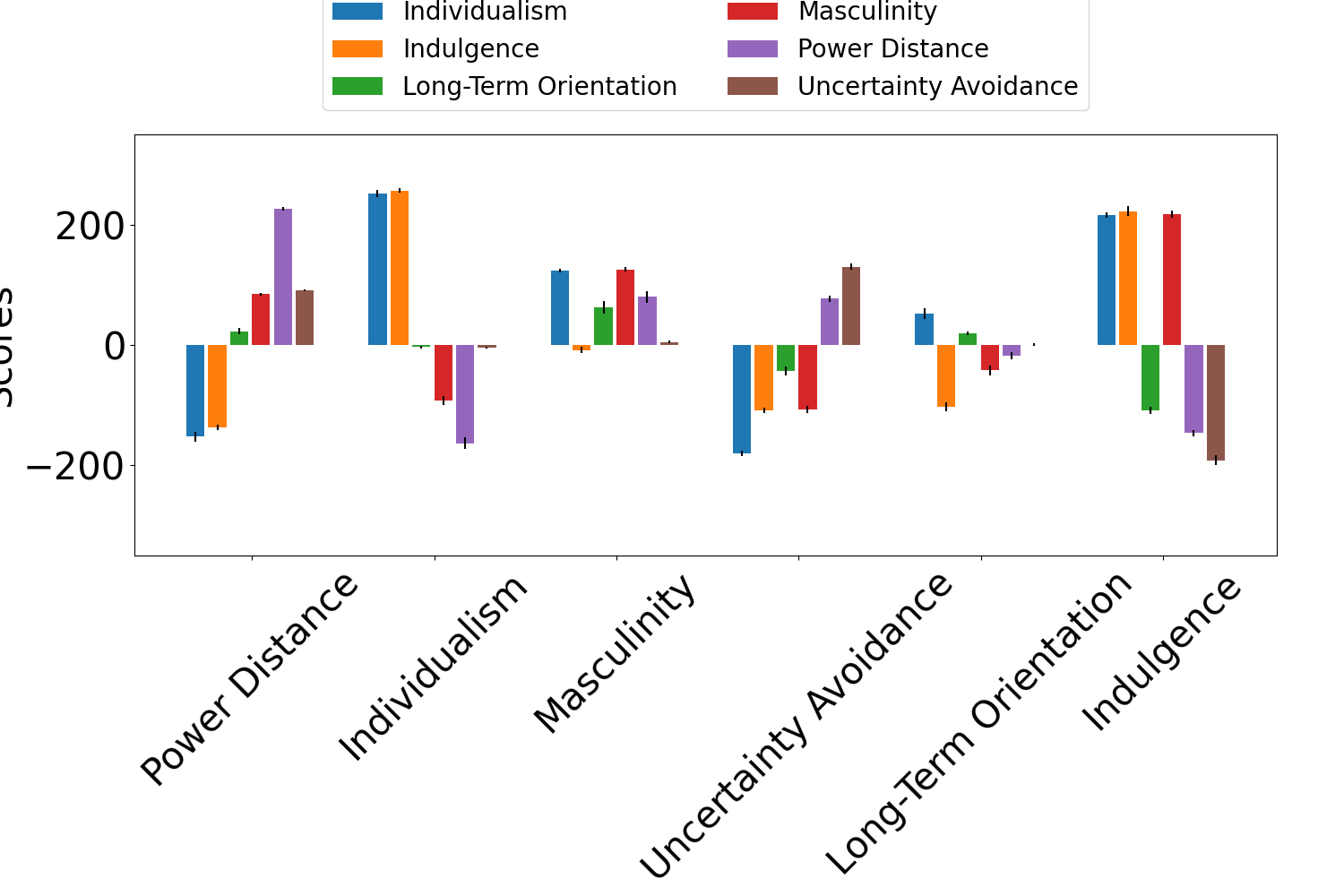}
\caption{GPT-4}
\end{subfigure}
\begin{subfigure}[b]{0.49\textwidth}
\includegraphics[width=\textwidth]{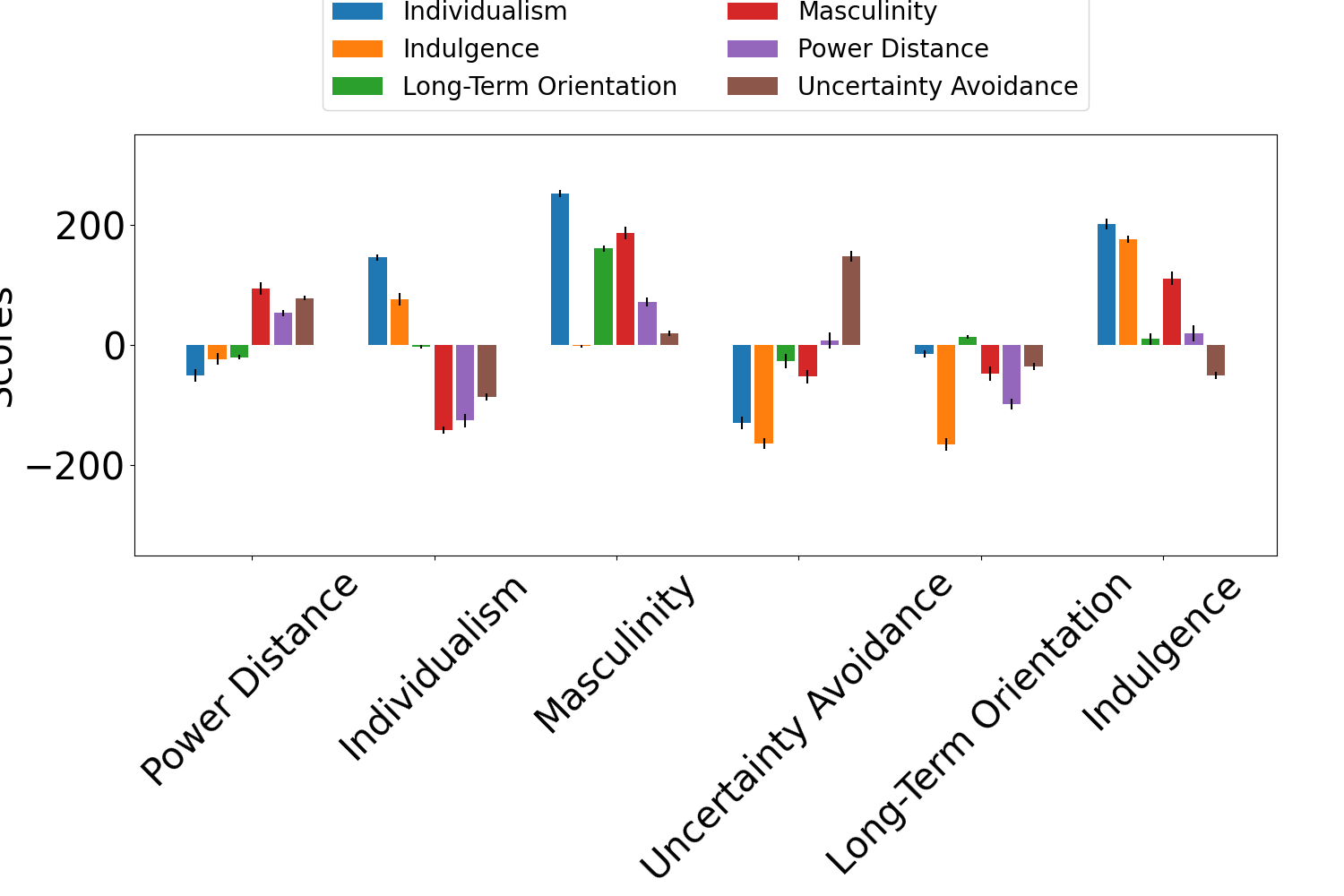}
\caption{GPT-3.5-0301}
\end{subfigure}
\\
\begin{subfigure}[b]{0.49\textwidth}
\includegraphics[width=\textwidth]{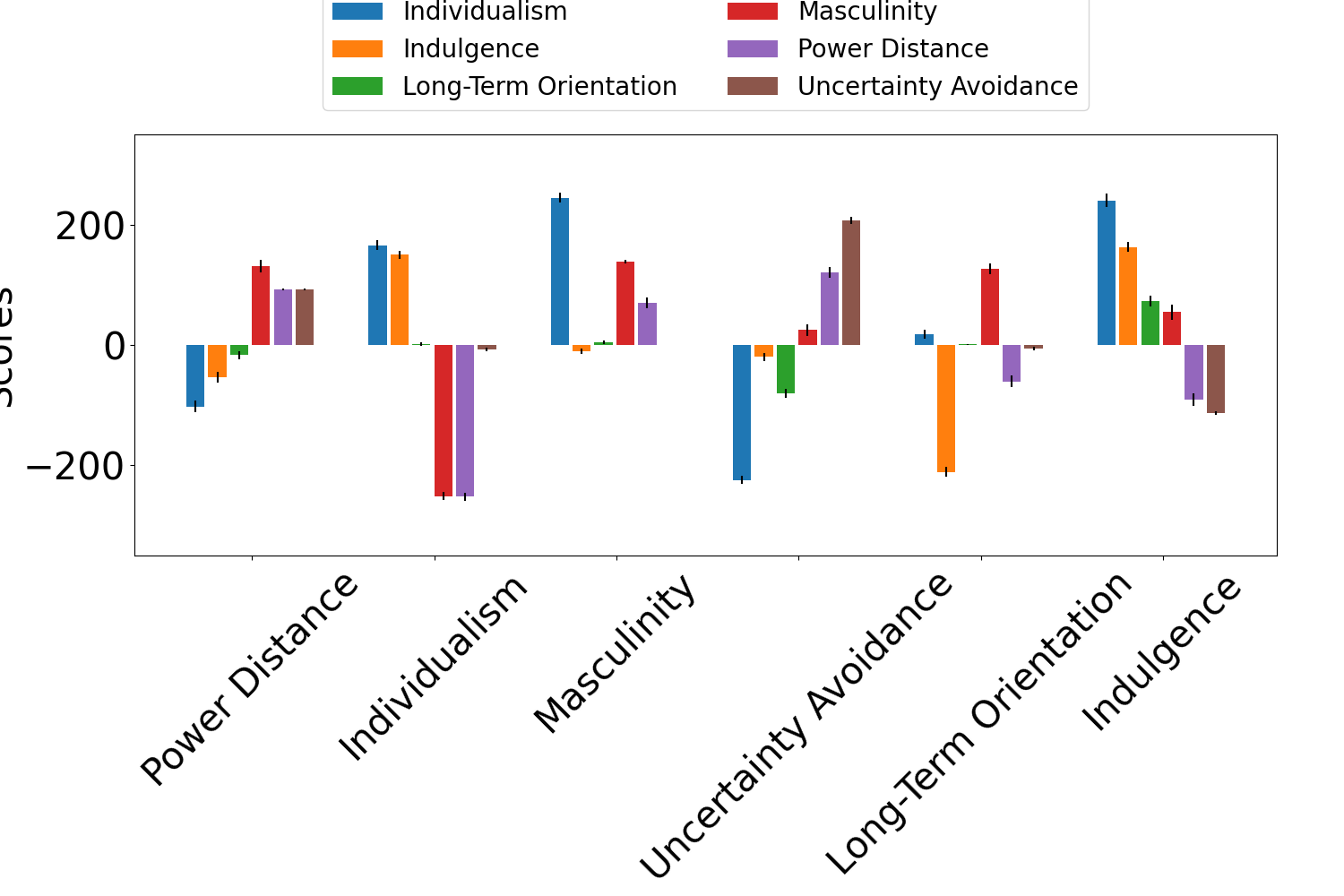}
\caption{GPT-3.5-0613}
\end{subfigure}
\begin{subfigure}[b]{0.49\textwidth}
\includegraphics[width=\textwidth]{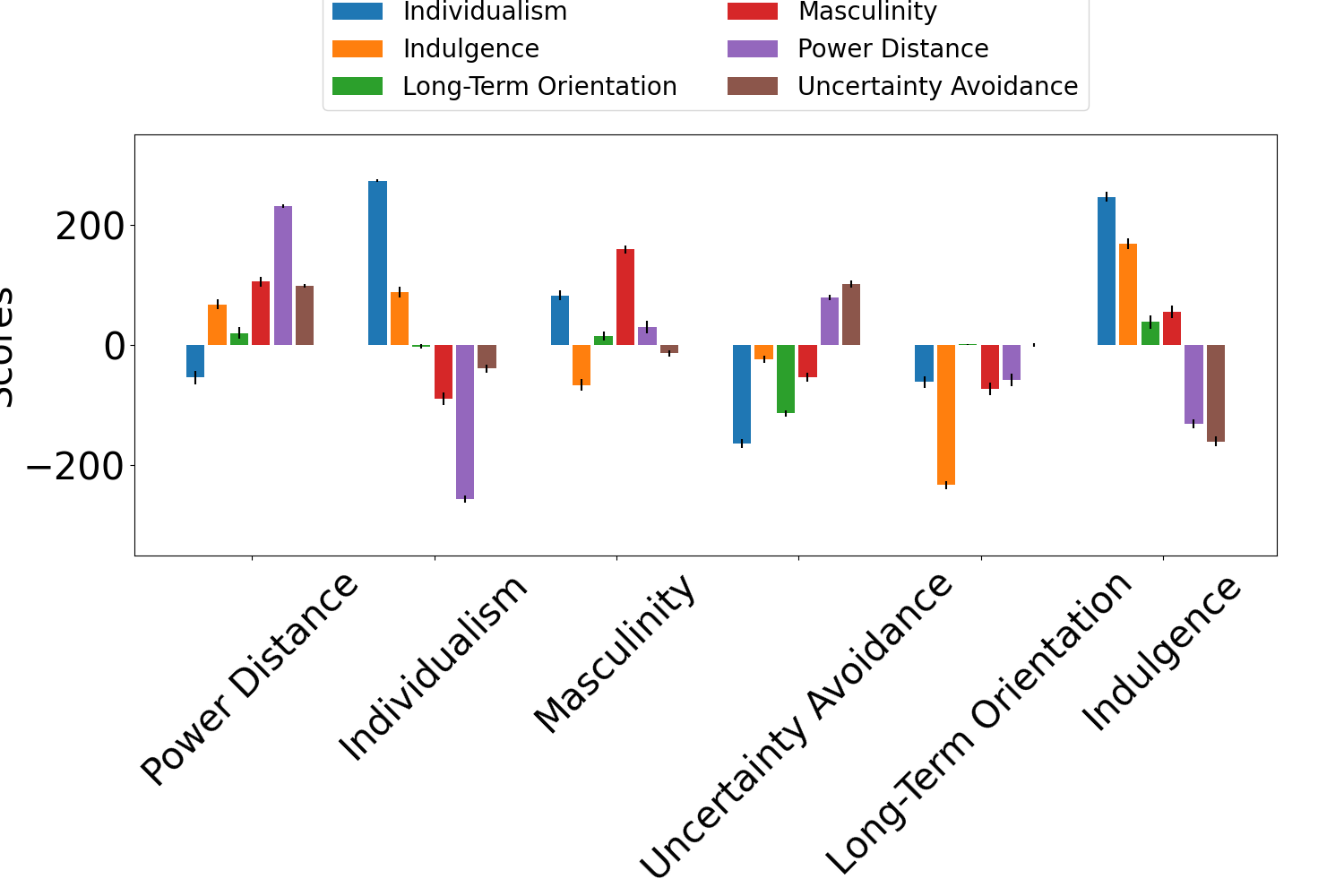}
\caption{Upstage-LLaMa-66B}
\end{subfigure}
\caption{Controllability scores on the VSM questionnaire (from table \ref{tab:contr_sys_exp})}
\label{fig:best_vsm}
\end{figure}

\begin{figure}
\centering
\begin{subfigure}[b]{0.49\textwidth}
\includegraphics[width=\textwidth]{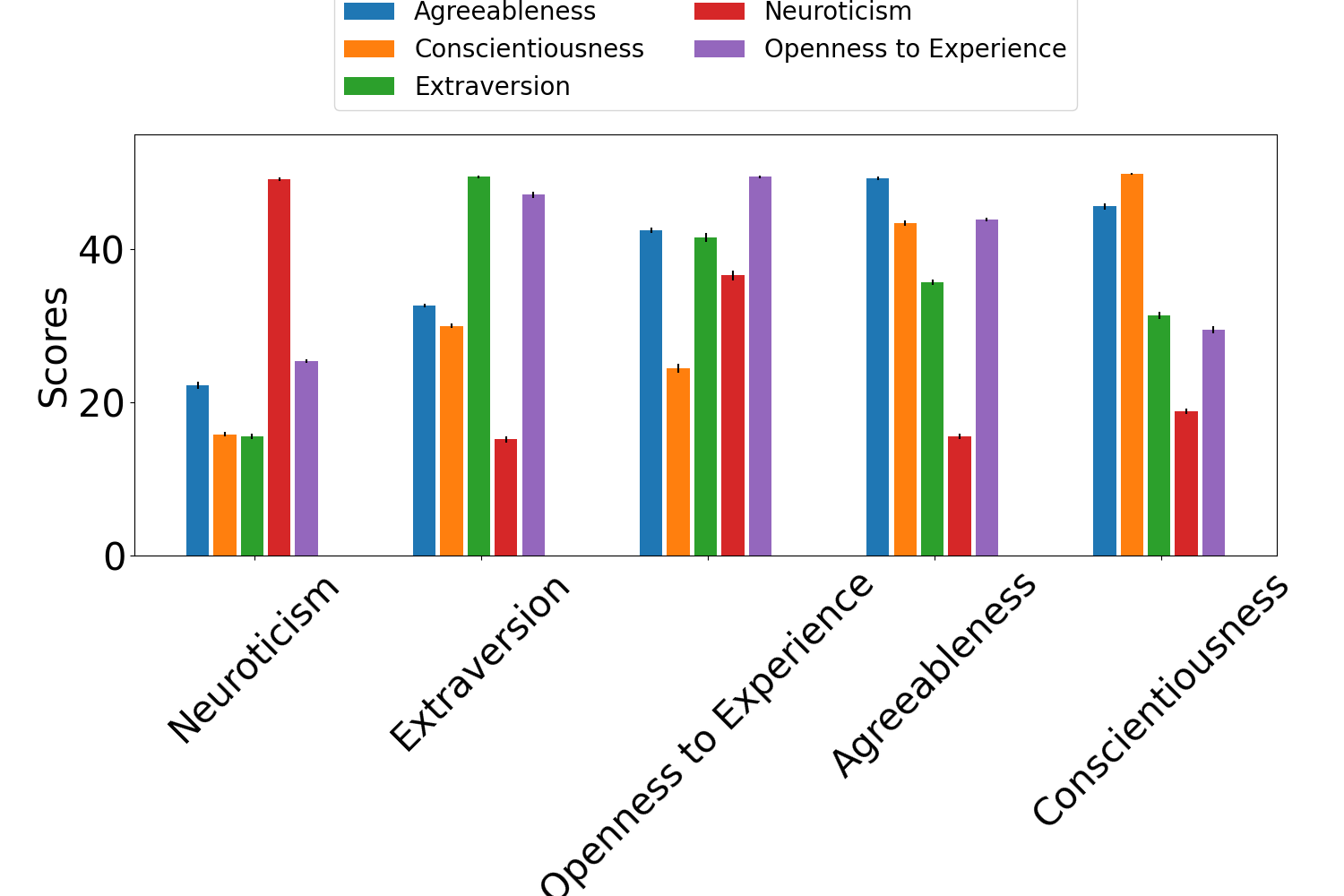}
\caption{GPT-4}
\end{subfigure}
\begin{subfigure}[b]{0.49\textwidth}
\includegraphics[width=\textwidth]{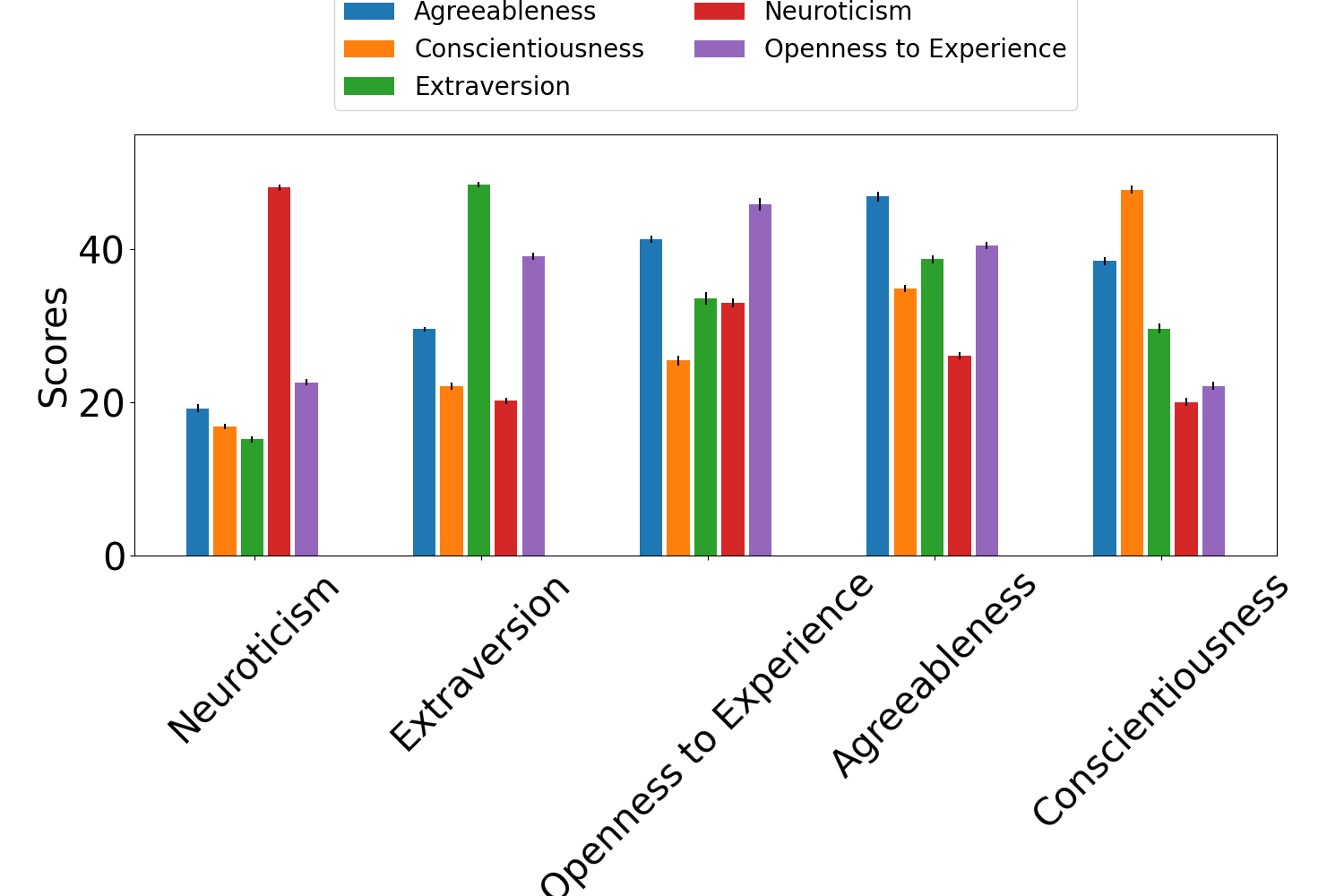}
\caption{GPT-3.5-0301}
\end{subfigure}
\\
\begin{subfigure}[b]{0.49\textwidth}
\includegraphics[width=\textwidth]{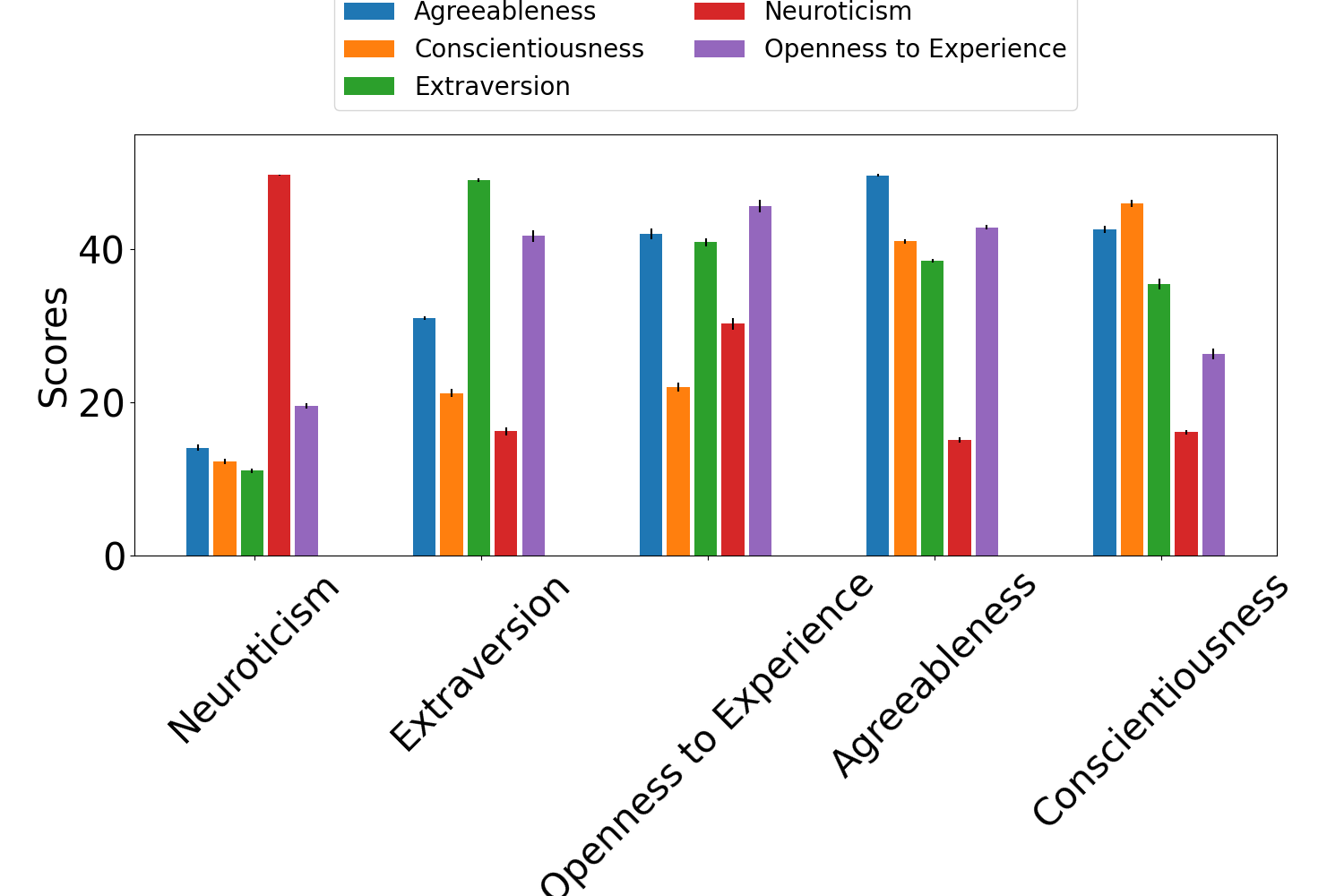}
\caption{GPT-3.5-0613}
\end{subfigure}
\begin{subfigure}[b]{0.49\textwidth}
\includegraphics[width=\textwidth]{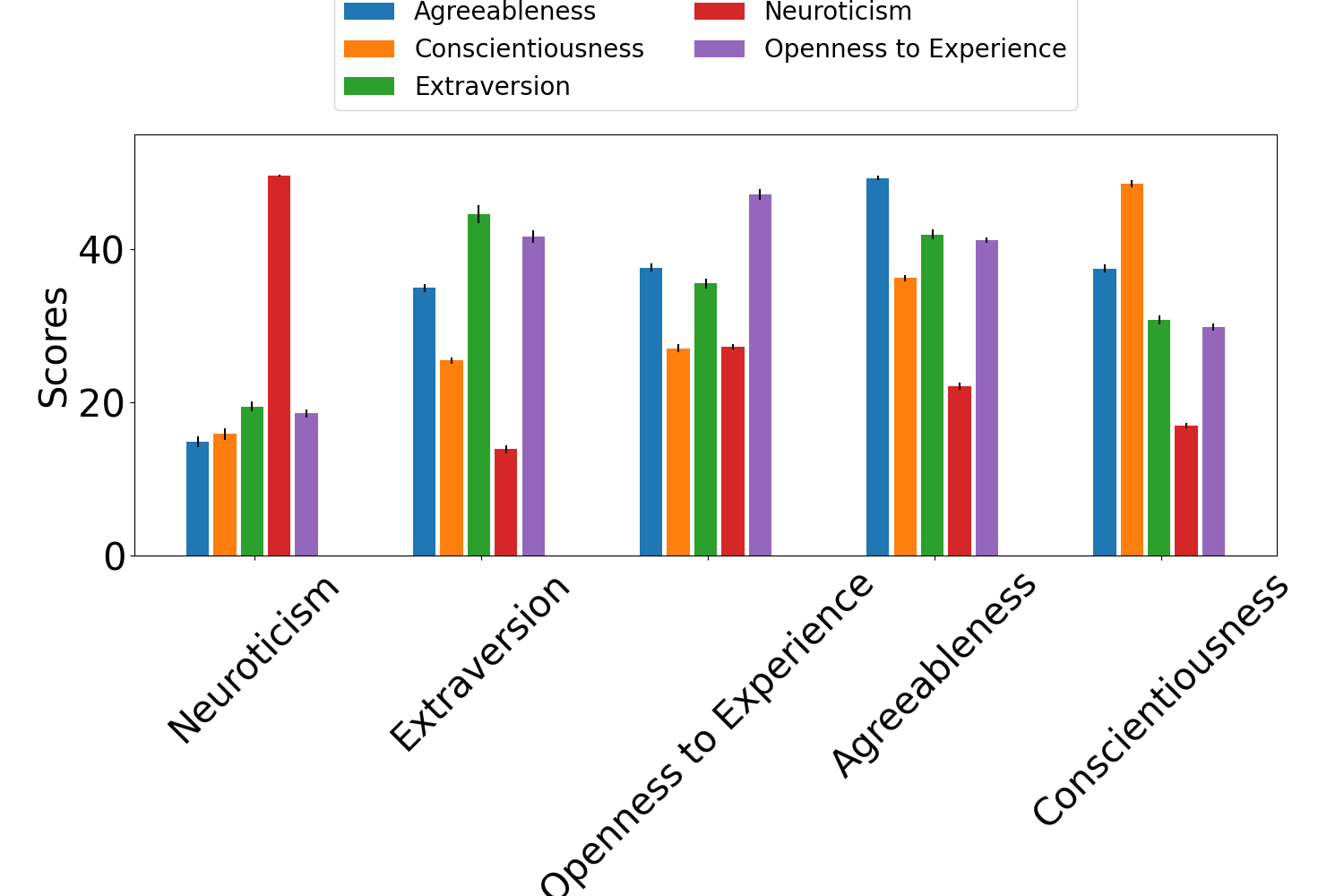}
\caption{Upstage-LLaMa-66B}
\end{subfigure}
\caption{Controllability scores on the IPIP questionnaire (from table \ref{tab:contr_sys_exp})}
\label{fig:best_ipip}
\end{figure}





\end{document}